\pdfoutput=1
\documentclass{article}
\usepackage{preprint,times}

\usepackage[utf8]{inputenc}
\usepackage[T1]{fontenc}
\usepackage{graphicx}
\usepackage{amsmath,amssymb}
\usepackage{booktabs}
\usepackage{pifont}
\usepackage{colortbl}
\usepackage{array}
\usepackage{caption,subcaption}
\usepackage{float}
\usepackage{xcolor}
\usepackage[pagebackref=false,breaklinks=true,colorlinks,bookmarks=false,citecolor=citecolor,linkcolor=linkcolor]{hyperref}
\usepackage{url}
\usepackage[capitalize,noabbrev]{cleveref}

\crefname{figure}{Fig.}{Figs.}
\Crefname{figure}{Figure}{Figures}

\crefname{table}{Table}{Tables}
\Crefname{table}{Table}{Tables}

\crefname{section}{Sec.}{Secs.}
\Crefname{section}{Section}{Sections}

\crefname{equation}{Eq.}{Eqs.}
\Crefname{equation}{Equation}{Equations}

\crefname{appendix}{Appendix}{Appendices}
\Crefname{appendix}{Appendix}{Appendices}

\definecolor{okGreen}{HTML}{009E73}
\definecolor{okVermillion}{HTML}{D55E00}
\definecolor{okRed}{HTML}{D62728}
\definecolor{okOrange}{HTML}{E69F00}
\definecolor{citecolor}{HTML}{0071BC}
\definecolor{linkcolor}{HTML}{ED1C24}

\newcommand{\method}{4Director}

\newcommand{\nparticipants}{20}

\newcommand{\cmark}{{\color{okGreen}\ding{51}}}
\newcommand{\pmark}{{\color{okOrange}\raisebox{0.1ex}{\scalebox{0.9}{\ding{51}}}\kern-0.15em\textsuperscript{\tiny p}}}
\newcommand{\xmark}{{\color{okRed}\ding{55}}}
\newcommand{\nmark}{{\color{black!35}--}}

\title{\method: Controlling Video World Models with Rigid 3D Geometry}

\author{%
\hspace*{-\tabcolsep}\parbox[t]{\textwidth}{\centering\bf
Wei Cao$^{1,2}$\hspace{0.8em}Hao Zhang$^{2}$\hspace{0.8em}Vikram Voleti$^{1}$\hspace{0.8em}Yuqun Wu$^{1,2}$\\[0.3ex]
Mallikarjun B R$^{1}$\hspace{0.8em}Shimon Vainer$^{1}$\hspace{0.8em}Mark Boss$^{1}$\hspace{0.8em}Yaoyao Liu$^{2}$\\[1.2ex]
{\normalfont $^{1}$Stability AI\qquad $^{2}$University of Illinois Urbana-Champaign}}\hspace*{-\tabcolsep}
}

\begin{document}

\maketitle
\lhead{Preprint}

\vspace{-1.2em}
\begin{figure}[H]
    \centering
    \includegraphics[width=0.94\linewidth]{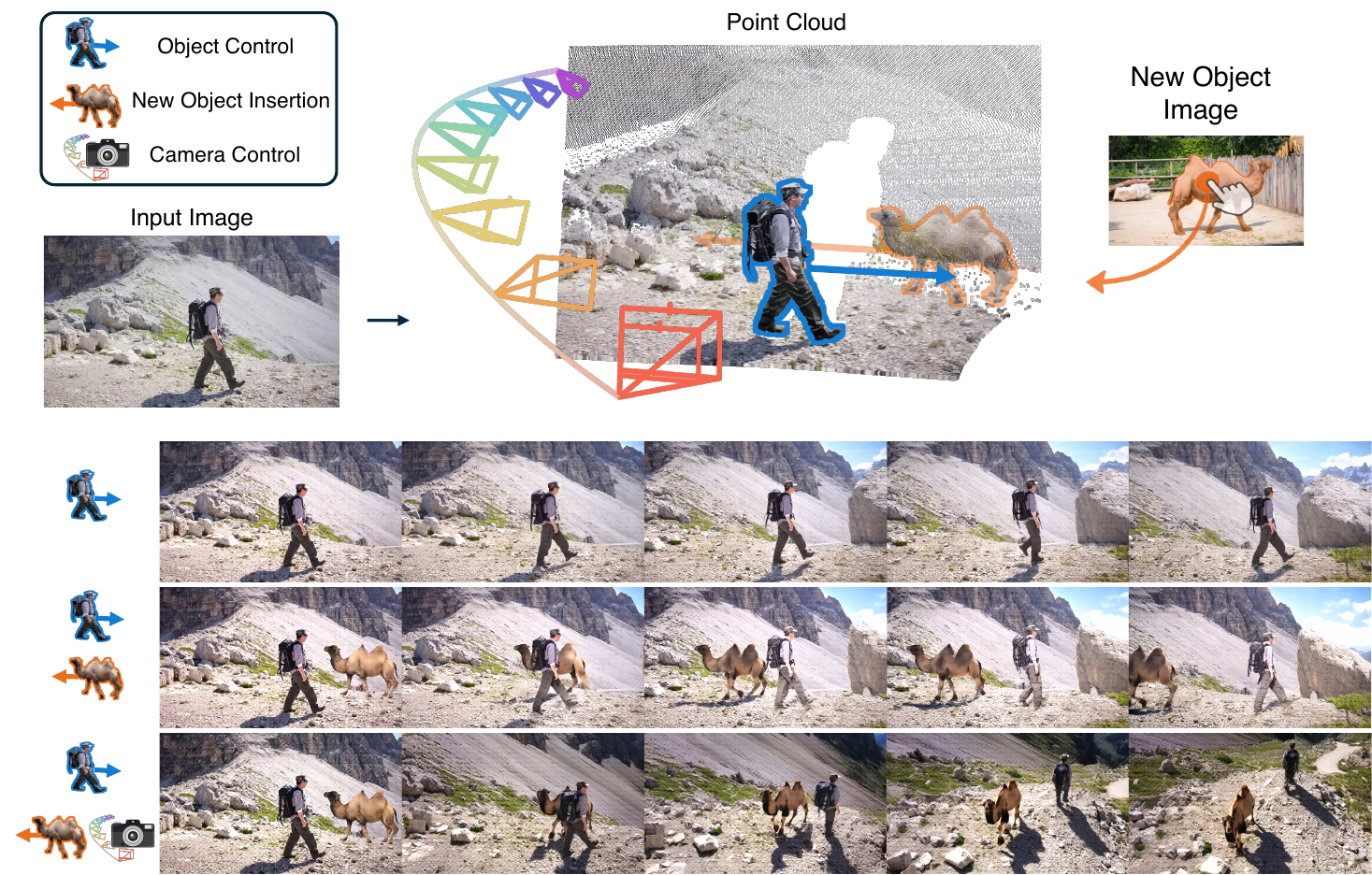}
    \caption{\textbf{Camera and object control from a single image.}
    \emph{Top:} the image is lifted into a background point cloud and
    complete rigid 3D geometry for each object, in which the user draws the
    object and camera trajectories and, optionally, inserts a new object from
    a reference image. \emph{Bottom:} \method{} generates videos that follow
    both trajectories, with plausible object motion, consistent appearance and
    illumination, and the background revealed by the camera filled in.}
    \label{fig:teaser}
\end{figure}
\vspace{-0.4em}

\begin{abstract}
Precise control over camera and object motion is essential for professional video production. 
Existing methods control objects only coarsely, through image-plane cues that are ambiguous in depth and rotation or through 3D tracks and blobs that lack complete geometry and lose consistency across viewpoint changes. 
We introduce \textbf{\method{}}, a video world model conditioned on an explicit 4D scene representation: each object is reconstructed once from the input image as a canonical mesh and moved by one prescribed rigid transformation per frame. 
This representation provides an intuitive 3D control interface and prevents unobserved geometry from being regenerated independently in every frame. We render the controlled scene as a depth video and introduce a \emph{Motion Adapter} that transforms this geometric scaffold into video while synthesizing view-consistent appearance, illumination, and non-rigid dynamics. 
For training, we construct \emph{\mbox{RealCOD-Rigid}}, a new dataset of 20{,}774 clips annotated with rigid 3D scenes by our automatic pipeline.
We further introduce \emph{Identity-Gated IoU} (IG-IoU), which jointly evaluates adherence to prescribed object motion and preservation of object identity. Experiments demonstrate that \method{} consistently outperforms prior methods in visual quality and in camera and object control.
Project page: \url{https://stability-ai.github.io/4director/}.
\end{abstract}

\section{Introduction}
\label{sec:intro}

A director stages a shot by designing the trajectories of the camera and
every actor, and leaves how an actor walks or how the light falls to the cast
and the crew. We develop \textbf{\method{}}, a video world
model~\citep{bruce2024genie,oasis,diamond} with the same principle: the
user prescribes the trajectory of each object and the camera in 3D
(\cref{fig:teaser}), and the generator supplies the non-rigid dynamics,
appearance and illumination along them.

\begin{figure}[t]
    \centering
    \includegraphics[width=0.96\linewidth]{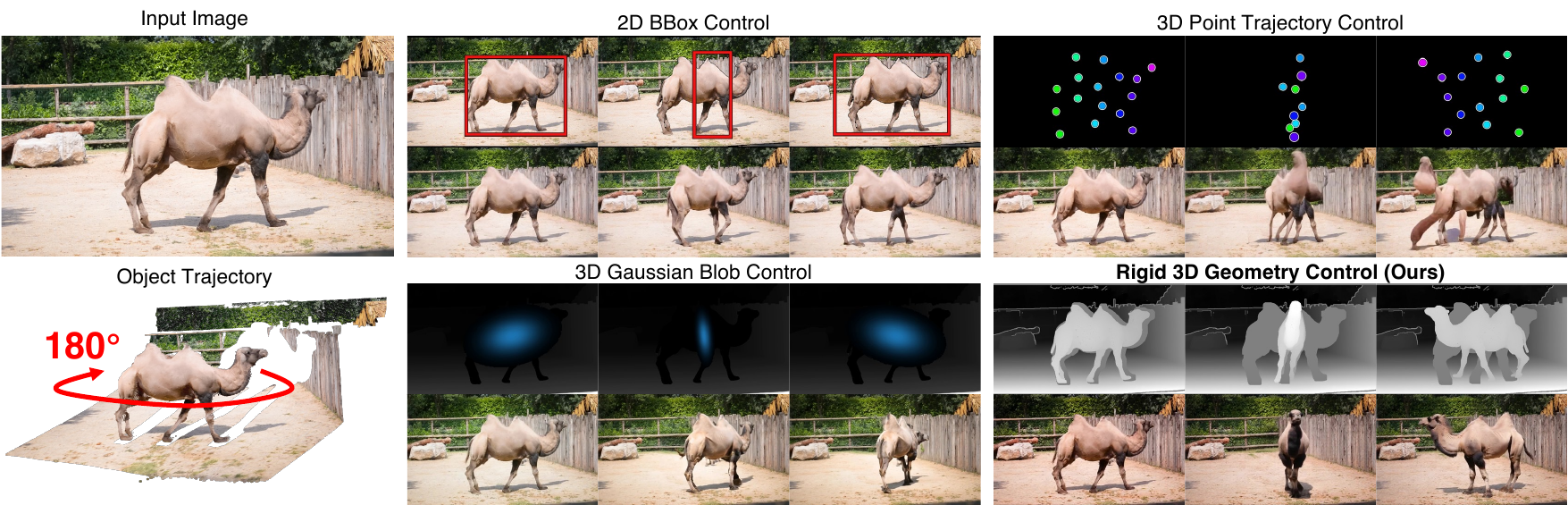}\\[3pt]
    \small\setlength{\tabcolsep}{5pt}
    \resizebox{\linewidth}{!}{%
    \begin{tabular}{lccccc}
        \toprule
        Method & \shortstack{Camera\\trajectory} & \shortstack{3D object\\position} & \shortstack{Object\\orientation} & \shortstack{Complete\\geometry} & \shortstack{New object\\insertion} \\
        \midrule
        MotionCtrl~\citep{motionctrl2024}            & \cmark & \pmark & \xmark & \xmark & \nmark \\
        Perception-as-Control~\citep{perceptioncontrol2025} & \cmark & \cmark & \xmark & \xmark & \nmark \\
        SymphoMotion~\citep{zhang2026symphomotion}   & \cmark & \cmark & \pmark & \xmark & \nmark \\
        VerseCrafter~\citep{zheng2026versecrafter}   & \cmark & \cmark & \pmark & \xmark & \nmark \\
        \midrule
        \rowcolor{okVermillion!10}
        \textbf{\method{} (ours)}                    & \cmark & \cmark & \cmark & \cmark & \cmark \\
        \bottomrule
    \end{tabular}}
    \caption{\textbf{Control representations.} \emph{Top:} given an input
    image and an object trajectory (a $180^\circ$ turn in place), we compare
    four control signals: a 2D bounding box, a 3D Gaussian blob, 3D point
    trajectories, and our rigid 3D geometry.
    \emph{Bottom:} capabilities of the compared representations
    (\cmark~supported, \pmark~partial, \xmark~not supported, \nmark~not stated
    in the paper).}
    \label{fig:control_representations}
\end{figure}

Existing controls fall short of such trajectories in one of two ways:
image-plane cues are ambiguous in depth and rotation, and 3D controls carry
incomplete object geometry. Image-plane cues such as dragged paths, bounding boxes, masks, or point
tracks~\citep{yin2023dragnuwa,wang2024boximator,zhang2025tora,zhou2025trackgo,geng2025motion,motionctrl2024}
state where an object appears in each frame, but one projection is
consistent with many 3D motions: a box does not determine the orientation of the object, a shrinking box does not distinguish an object moving away from one
becoming smaller, and a 2D path conflates camera parallax with object motion
(\cref{fig:control_representations}). 3D-aware controls avoid this
ambiguity with rigid trajectories or 3D
boxes~\citep{fu20243dtrajmaster,shuai2025free,wang2025cinemaster}, or with proxies lifted from the input, such as tracked 3D
points~\citep{zhang2026symphomotion}, spheres on tracked
parts~\citep{perceptioncontrol2025} and one 3D Gaussian blob per
object~\citep{zheng2026versecrafter}. Their geometry, however, is incomplete: a trajectory
or box carries no surface, and a lifted proxy captures only the visible shell
of the object and has no geometry for regions that a turn or a new viewpoint
brings into view (\cref{fig:control_representations}, Gaussian and point
panels); point-based proxies are, moreover, driven by per-point trajectories,
which are cumbersome to author one by one.

To address both shortcomings, we propose \textbf{\method{}}, a video world
model driven by an explicit 4D scene representation. Every object is a
complete canonical mesh, reconstructed once and moved by one rigid
transformation per frame. The meshes, a background point cloud and the camera share one coordinate frame. Explicit 3D space removes the ambiguity of image-plane
cues and gives the user a direct interface: a turn is a rotation, motion in depth is a displacement rather than a
change of size. Complete geometry
removes the limitation of lifted proxies: the mesh has a surface on every side, so
the geometry that a turn or new viewpoint reveals is rendered rather than regenerated (\cref{fig:control_representations}). And one rigid transformation moves the whole object without per-point trajectories.

We render the scene to a depth video in which every object moves as a single
rigid body. We design a \emph{Motion Adapter}, a trainable branch that
injects this rigid rendering into a pretrained video
generator~\citep{wan2025}. To train it, we build an automatic annotation
pipeline that recovers the rigid 3D scene of a monocular clip, and run it on
\mbox{RealCOD-25K}~\citep{zhang2026symphomotion} to construct
\emph{\mbox{RealCOD-Rigid}}, a dataset of 20{,}774 annotated clips. The adapter
is trained to generate each clip from its rigid rendering, and so learns to
supply what the rendering lacks: appearance, illumination, non-rigid dynamics
and the background that a camera move reveals.

Since \method{} is capable of generating videos with explicit control on the objects and camera, we evaluate \method{} on visual quality, object control, and camera control. For visual quality, we employ standard metrics as well as a user study. Camera control is evaluated with the standard trajectory error. Object control lacks a comparable metric: mask IoU alone rewards correct placement regardless of identity. We therefore introduce
\emph{Identity-Gated IoU} (IG-IoU), which accumulates mask IoU only over
frames in which the object's identity is preserved. From our evaluations, we find that \method{} outperforms prior methods on all three criteria.
Our contributions are as follows:

\begin{itemize}
    \setlength{\itemsep}{0pt}
    \item We introduce an \textit{\textbf{explicit 4D scene representation}}
    with complete object geometry and per-frame rigid transformations that
    holds the camera and every object in one coordinate frame, resolving the
    ambiguity of image-plane cues and the incomplete geometry of lifted
    proxies.

    \item We present the \textit{\textbf{Motion Adapter}}, a trainable branch
    that drives a pretrained video generator with the rendered rigid scene, so
    that the geometry fixes the camera and object motion while the generator
    supplies appearance, illumination and non-rigid dynamics.

    \item We introduce \textit{\textbf{\mbox{RealCOD-Rigid}}}, a dataset of
    20{,}774 clips annotated with complete object meshes, per-frame rigid
    transformations and camera trajectories by an automatic pipeline, and
    \textit{\textbf{Identity-Gated IoU}}, a metric that scores object control
    only where identity is preserved.

    \item Experiments and a user study show that \method{}
    outperforms state-of-the-art methods in visual quality, camera control, and object control.
\end{itemize}

\section{Related Work}
\label{sec:related}
\noindent\textbf{Video world models.}
World models roll out future observations~\citep{ha2018world,lecun2022path,hafner2023mastering,cao2025pseudo};
diffusion and transformer backbones now produce
realistic rollouts conditioned on actions, text or a camera path~\citep{blattmann2023stable,bruce2024genie,parkerholder2024genie2,parker2025genie,oasis,diamond,agarwal2025cosmos,alhaija2025cosmos,hu2023gaia,che2024gamegen,yu2025gamefactory,he2025matrix,li2025hunyuan,ma2023generating,ji2026ac3s},
and extend the horizon with memory~\citep{streamingt2v,freenoise,po2025long,contextasmemory,worldmem,vmem,wu2025video,hydra,liveworld,worldplay,relic}.
A geometry-aware line, including DeepVerse~\citep{chen2025deepverse},
Voyager~\citep{huang2025voyager}, Yume~\citep{mao2025yume} and
Aether~\citep{zhu2025aether}, adds reconstructed 3D structure for consistent
exploration~\citep{sun2024dimensionx,team2025hunyuanworld,yang2025matrix}.
Their control, however, is expressed as text, actions or camera tokens,
through which the motion of individual objects cannot be prescribed. \method{} instead exposes the scene itself: an explicit 4D scene
representation edited before generation, with complete object meshes,
one rigid transformation per frame and the camera in one coordinate frame.
We therefore use the term \emph{world model}, as VerseCrafter~\citep{zheng2026versecrafter}
does, for a generator whose output follows an explicit, editable scene state
rather than an interactive rollout.

\suppressfloats[t]

\noindent\textbf{Camera control.}
Pose conditioning, as Pl\"ucker rays~\citep{he2024cameractrl,sitzmann2021light,zhou2025stable,feng2024i2vcontrolcamera,li2025realcam,wang2025epic}, epipolar
attention~\citep{xu2024camco}, conditioning at chosen layers~\citep{bahmani2025ac3d,bahmani2024vd3d} or
multi-view and reference-driven control~\citep{kuang2024collaborative,bai2024syncammaster,he2025cameractrl,luo2025camclonemaster,zheng2024cami2v}, says nothing about what
the new view should contain. A second line renders lifted geometry along the target path: a colored point
cloud~\citep{yu2024viewcrafter}, a cached 3D scene~\citep{ren2025gen3c}, warped
geometry with its holes repaired~\citep{hou2024training,yu2025trajectorycrafter,song2026taming,popov2025camctrl3d,hu2025ex}, a source video re-rendered along a
new path~\citep{van2024generative,zhang2024recapture,bai2025recammaster,cao2026freeorbit4d}, a point cloud with a human body~\citep{cao2025uni3c}, or depth maps for a world
model~\citep{alhaija2025cosmos}, injected through a branch such as ControlNet or
VACE~\citep{controlnet,vace2025}. Lifted geometry covers only what the reference view saw, so what a camera move reveals is hallucinated anew in each frame, although complete shapes are available from image or 4D
generation and reconstruction~\citep{li2026pixal3d,voleti2024sv3d,xie2025sv4d,yao2025sv4d,zhang2026stable,wu2025monopatchnerf,wu2022sparse,cao2024motion2vecsets,tang2026motion2vecsets}. In contrast, \method{} reconstructs every
object as a complete mesh, so the object geometry that a camera move reveals is
rendered rather than regenerated, and its Motion Adapter learns to supply appearance, illumination and
non-rigid dynamics on top of fixed geometry.

\noindent\textbf{Object control.}
Image-plane cues, from dragged paths to tracked points and boxes, state
where an object should appear~\citep{wang2023videocomposer,yin2023dragnuwa,ma2024trailblazer,zhang2025tora,zhou2025trackgo,wang2024boximator,geng2025motion,shi2024motion,niu2025mofa,jain2024peekaboo,qiu2024freetraj,li2025magicmotion,wu2025draganything,xing2025motioncanvas,burgert2025go,wang2025ati,chu2025wan}; several systems add a camera module~\citep{yang2024direct,feng2024i2vcontrol,lei2024animateanything,li2025image,zheng2025vidcraft3}, MotionCtrl among them, with camera extrinsics and 2D object
trajectories~\citep{motionctrl2024}. Moving the
control into 3D, as a depth-augmented trajectory~\citep{wang2025levitor,wang2024objctrl}, a 6-DoF
pose~\citep{fu20243dtrajmaster,shuai2025free,liang2025realismotion}, a 3D box~\citep{wang2025cinemaster} or a coarse reconstruction in a 3D engine~\citep{zhang2025i2v3d}, fixes
the 3D position, and all but the trajectory also fix orientation. The closest methods lift a proxy of the object from the source and render
it with the camera: Perception-as-Control places spheres on tracked
parts~\citep{perceptioncontrol2025}, SymphoMotion~\citep{zhang2026symphomotion} and
Diffusion as Shader~\citep{gu2025diffusion} render 3D point tracks over a point cloud,
and VerseCrafter moves one 3D Gaussian per object beside a background
cloud~\citep{zheng2026versecrafter}. Image-plane cues do not fix depth and rotation, 3D poses and boxes carry no surface,
lifted proxies carry only the visible shell, and point-based ones are driven by
per-point trajectories. In contrast, \method{} moves a complete canonical mesh by one rigid
transformation per frame, so that position, orientation and the revealed surface are
fixed before generation and one trajectory moves the whole object.

\section{Method}
\label{sec:method}

\begin{figure}[t]
    \centering
    \includegraphics[width=\linewidth]{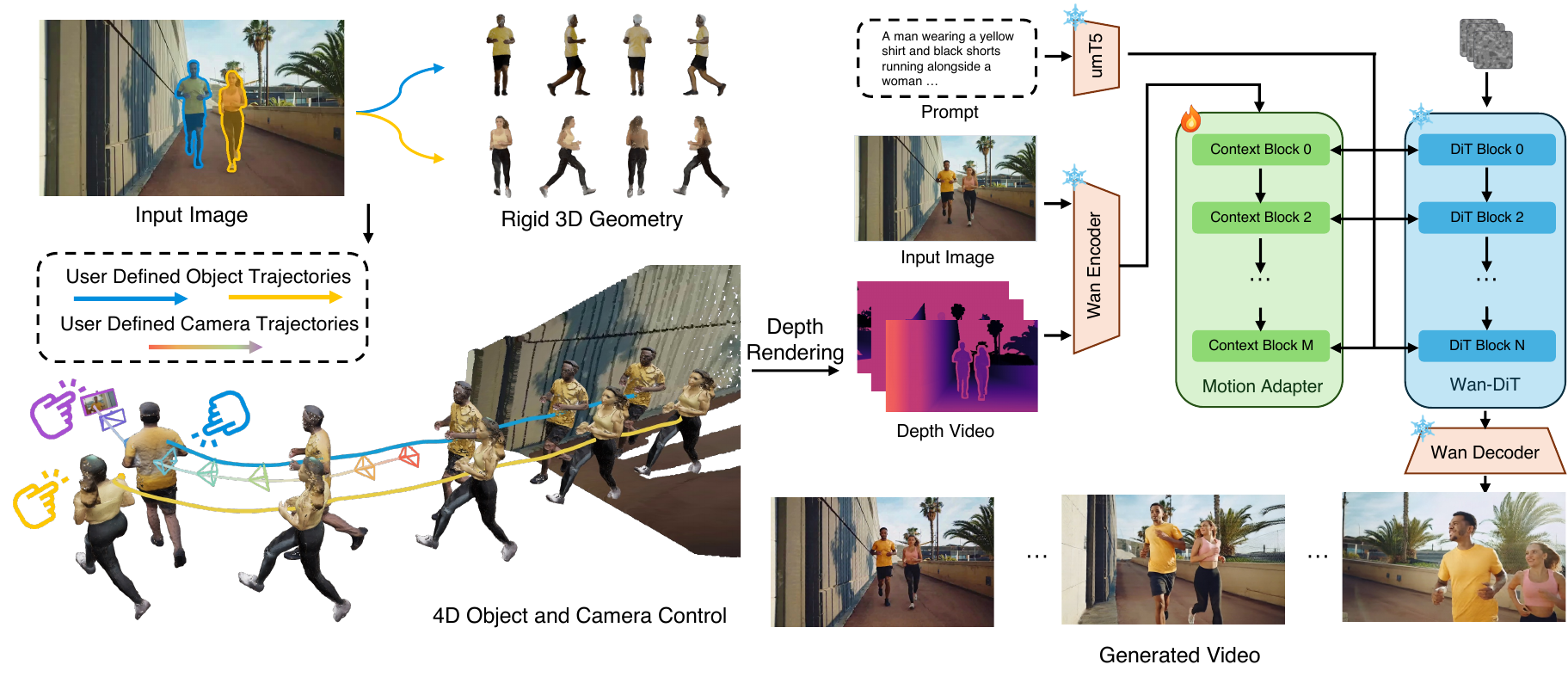}
    \caption{\textbf{Overview of \method.} \emph{Left:} each object is
    reconstructed once as a complete canonical mesh and moved by one rigid
    transformation per frame; the meshes, a background point cloud and the
    camera share one coordinate frame, in which the user prescribes the
    trajectory of each object and the camera, and the scene is rendered to a
    depth video (Sec.~\ref{sec:scene_representation}). \emph{Right:} the
    \emph{Motion Adapter}, a trainable branch, injects this rigid rendering
    into a pretrained video generator, which supplies appearance,
    illumination and non-rigid dynamics (Sec.~\ref{sec:video_training}); it
    is trained on \mbox{RealCOD-Rigid} (Sec.~\ref{sec:shared_geometry}).}
    \label{fig:method_pipeline}
\end{figure}

We study camera and object control from a single image: given an image, a text prompt, a camera trajectory
$\{\mathbf E^t\}_{t=1}^{F}$, and a rigid trajectory
$\{\mathbf T_o^t\}_{t=1}^{F}$ for each object $o$ that the user marks in the
image, our goal
is to synthesize a video of $F$ frames that starts from the image, follows
the camera trajectory and moves each object along its trajectory, with its
identity preserved and plausible non-rigid dynamics. Here $\mathbf E^t$ is the world-to-camera matrix of frame
$t$ and
$\mathbf T_o^t=(\mathbf R_o^t,\mathbf t_o^t)\in\mathrm{SE}(3)$ the rigid
transformation of object $o$ from its placement in the image to frame $t$
(Fig.~\ref{fig:method_pipeline}).

\suppressfloats[t]

\subsection{Rigid 3D Geometry Control}
\label{sec:scene_representation}

\noindent\textbf{Rigid 3D geometry.} We first lift the input image into 3D.
MoGe-2~\citep{moge22025} estimates its depth and intrinsic matrix
$\mathbf K$, and SAM~2~\citep{ravi2024sam2} turns the user's clicks on each
object into a mask. The camera of the image defines the reference frame ($\mathbf E^1$ is the identity), in which two kinds of geometry are built. The
background, the image outside the masks, is back-projected into a static
point cloud $\mathcal P$. Each marked object is reconstructed by
Pixal3D~\citep{li2026pixal3d} into a complete canonical mesh $\mathcal M_o$,
with the sides the image does not show, and aligned to its mask and depth.
Background, camera, and objects thus share one coordinate frame.

\noindent\textbf{Camera and object trajectories.}
\label{sec:directing}
The user prescribes the trajectories in this frame by keyframes in a 3D
viewer. Object $o$ moves as a whole by $\mathbf T_o^t$, with $\mathbf T_o^1$ the identity, so that at frame $t$ it is the placed mesh
\begin{equation}
    \mathcal M_o^t=\{\mathbf R_o^t\,\mathbf v+\mathbf t_o^t\mid\mathbf v\in\mathcal M_o\},
    \label{eq:mesh_motion}
\end{equation}
and the camera moves by $\mathbf E^t$. The scene thus carries only rigid
motion. A new object can be
inserted as a mesh reconstructed from a separate image and given a
trajectory in the same way; its textured mesh is rendered into the input
image, which then serves as the first frame. Altogether, the rigid 3D scene is the tuple
\begin{equation}
    \mathcal S=\big(\mathcal P,\ \{\mathcal M_o\},\ \{\mathbf T_o^t\},\ \mathbf K,\ \{\mathbf E^t\}\big),
    \label{eq:scene_state}
\end{equation}
in which $\mathcal P$, the meshes and $\mathbf K$ are constant and only
$\{\mathbf T_o^t\}$ and $\{\mathbf E^t\}$ vary in time.

\noindent\textbf{Depth rendering.}
\label{sec:control_rendering}
For each frame, $\mathcal P$ and the placed meshes $\{\mathcal M_o^t\}$ are
projected with $(\mathbf K,\mathbf E^t)$ into a depth map, in which each
pixel takes the depth of the nearest surface and pixels without geometry are
empty. The resulting depth video $\mathbf C$ is the control. It carries only
the viewpoint, the rigid transformation of every object and their occlusion;
appearance, illumination, non-rigid dynamics such as the movement of a
walker's limbs, and the background beyond what the image shows are absent,
and supplying them is the task of the Motion Adapter.

\subsection{Motion Adapter}
\label{sec:video_training}

\noindent\textbf{Architecture.} We build on
Wan2.1-VACE-14B~\citep{wan2025,vace2025}, a latent video diffusion
transformer whose VAE, umT5~\citep{chung2023unimax} text encoder and DiT
serve as a generic video prior and are left unchanged. The \emph{Motion
Adapter} is a DiT-style branch that follows the context-branch design of
VACE: as shown in Fig.~\ref{fig:method_pipeline} (right), the VAE encodes the
depth video $\mathbf C$ and the input image into a context stream, which
eight context blocks, one for every fifth DiT block, propagate with
cross-attention to the prompt and inject into the DiT as linearly projected
hints.

\noindent\textbf{Training.} Each training pair is built from one original
clip (Sec.~\ref{sec:shared_geometry}): the control video rendered from the
clip's recovered rigid 3D scene is the condition, and the clip is the
target to be generated; its first frame is the input image and its text
description is the prompt. The control video contains the rigid part of each object's motion but not
the non-rigid part. A surface point $i$ of object $o$ is observed at
\begin{equation}
    \mathbf X_i^t=\mathbf R_o^t\,\mathbf Y_i+\mathbf t_o^t+\mathbf r_i^t,
    \label{eq:rigid_residual}
\end{equation}
where $\mathbf Y_i$ is its position on the canonical mesh, the first two
terms are the rigid motion of the object, and the residual $\mathbf r_i^t$
is its non-rigid motion, such as a swinging limb; only the first two terms
enter the control video. To generate the clip, the adapter must
therefore follow the rigid geometry and let the generator supply what the
control omits: the non-rigid motion, with appearance and
illumination.

The adapter is trained with the standard flow-matching objective. Let
$\mathbf z$ be the latent of the clip and
$\mathbf z_k=(1-\sigma_k)\,\mathbf z+\sigma_k\,\boldsymbol\epsilon$ its
noised version at timestep $k$, with Gaussian noise $\boldsymbol\epsilon$.
The model predicts the velocity $\boldsymbol\nu(\mathbf z_k,k;\mathbf C)$
from $\mathbf z_k$, $\mathbf C$, the image and the prompt, and the adapter
minimizes
\begin{equation}
    \mathbb E_{\mathbf z,k,\boldsymbol\epsilon}
    \Big[
        \lambda_k\,
        \big\|\boldsymbol\nu(\mathbf z_k,k;\mathbf C)-
        (\boldsymbol\epsilon-\mathbf z)\big\|_2^2
    \Big],
    \label{eq:flow_loss}
\end{equation}
where $\lambda_k$ weights the timesteps.

\noindent\textbf{Inference.} At inference, the scene built from the input image and directed by the user
is rendered to its control video $\mathbf C$, and the adapter generates the
video from $\mathbf C$, the image and the prompt. Since the adapter sees the
scene only as a depth video, a directed scene is treated exactly like a
recovered one: new trajectories and inserted objects require no additional
training (Fig.~\ref{fig:teaser},
Appendix~\ref{sec:authored_section}).

\section{The \mbox{RealCOD-Rigid} Dataset}
\label{sec:shared_geometry}

\begin{figure}[t]
    \centering
    \includegraphics[width=\linewidth]{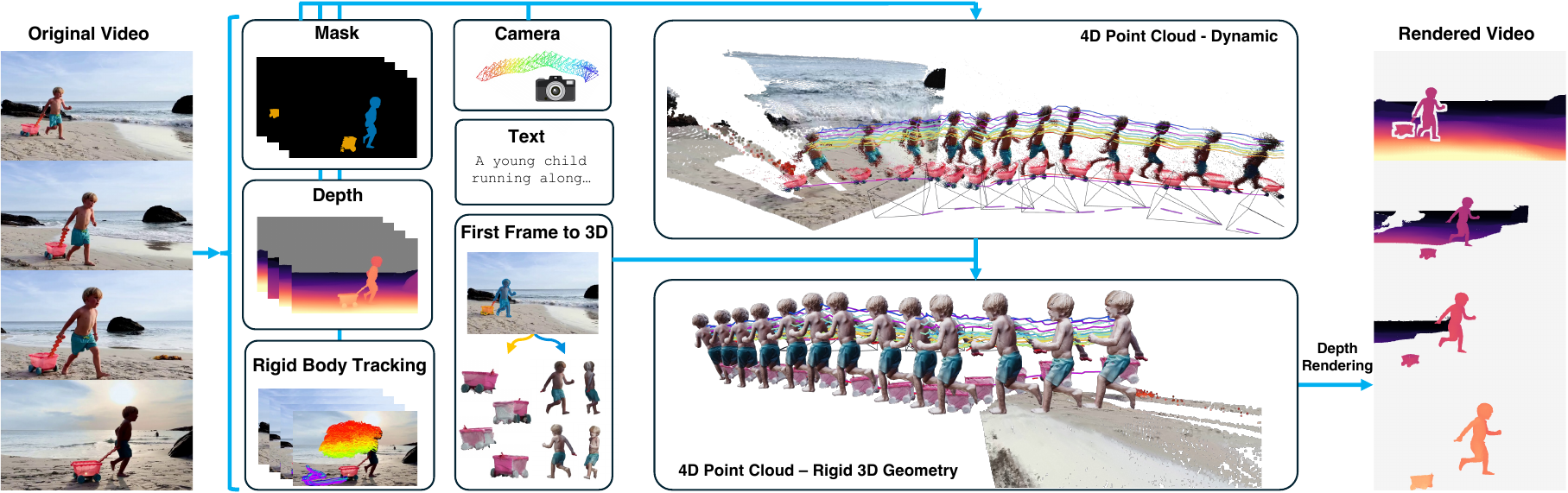}
    \caption{\textbf{\mbox{RealCOD-Rigid} annotation pipeline.} From each clip
    we estimate object masks, depth, and the camera trajectory, which together
    give a dynamic 4D point cloud of the scene (\emph{4D Point Cloud -- Dynamic});
    we reconstruct
    a canonical mesh for every masked object from the first frame
    (\emph{First Frame to 3D}) and track 3D points on it as a rigid body
    (\emph{Rigid Body Tracking}), which yields one rigid transformation per
    frame and aligns the canonical mesh to every frame (\emph{4D Point Cloud
    -- Rigid 3D Geometry}), giving the rigid 3D scene. Rendering it to a depth video yields the
    control; the original clip is the training target, with its text description as
    the prompt.}
    \label{fig:data_pipeline}
\end{figure}

Training the Motion Adapter requires clips paired with their rigid 3D
scenes, which no dataset provides and which cannot be annotated by hand at
scale. We therefore recover the rigid 3D scene of each clip automatically.
Starting from the clips of \mbox{RealCOD-25K}~\citep{zhang2026symphomotion},
each with a text prompt and SAM3 masks for one or two objects, the
pipeline lifts the first frame into 3D as in
Sec.~\ref{sec:scene_representation} and then tracks the camera and every object through the clip; the 20{,}774 clips that pass every stage and are not
used for evaluation form \mbox{RealCOD-Rigid}. It has four stages (Fig.~\ref{fig:data_pipeline}).

\noindent\textbf{(1) Camera and depth estimation.} MegaSaM~\citep{megasam2025}, with
MoGe-2~\citep{moge22025} as depth prior and UniDepthV2~\citep{unidepthv22025}
as metric branch, estimates per-frame depth, the intrinsic matrix $\mathbf K$
and the camera trajectory $\{\mathbf E^t\}$; its coordinate frame, of
arbitrary scale, is the scene frame of the clip. The background point cloud
$\mathcal P$ is the first frame back-projected into this frame with the
object masks removed.

\noindent\textbf{(2) First frame to 3D.}
\label{sec:canonical_alignment}
Pixal3D~\citep{li2026pixal3d} reconstructs the canonical mesh $\mathcal M_o$
from the object's crop in the first frame. Pixal3D is pixel-aligned, so every
visible mesh point comes with the first-frame pixel it was reconstructed from, and
the depth of stage (1) places that pixel in the scene. A robust similarity
fit (rotation, translation and scale) on these pairs moves the mesh onto the
object at $t=1$, which defines $\mathbf T_o^1$ as the identity.

\noindent\textbf{(3) Rigid body tracking.}
\label{sec:dominant_motion}
TAPIP3D~\citep{zhang2026tapip3d} tracks 3D points on each object through the
clip in the scene frame, seeded inside the object's mask; these tracks are
the observations $\mathbf X_i^t$ of Eq.~\eqref{eq:rigid_residual}. Every frame needs the rigid transformation that moves the object from its
first-frame placement to where the tracks observe it. Not every track moves
rigidly: a limb swings and some tracks drift, and a plain fit would follow
them. We therefore find the object's \emph{stable core} $\Omega_o$, the tracks
that move together rigidly across the clip, by alternating a robust fit with
the selection of its inliers, and fit each frame on the core alone,
\begin{equation}
    (\mathbf R_o^t,\mathbf t_o^t)=\arg\min_{\mathbf R,\,\mathbf t}\
    \sum_{i\in\Omega_o}\rho\big(\|\mathbf R\,\mathbf Y_i+\mathbf t-\mathbf X_i^t\|\big),
    \label{eq:core_fit}
\end{equation}
with $\mathbf Y_i$ the position of track $i$ at $t=1$ and a robust loss
$\rho$.

\noindent\textbf{(4) Depth rendering.} Placing each mesh by its rigid
transformation completes the rigid 3D scene of the clip. Where the per-frame
depth of stage (1) still shows a limb swinging, this scene moves each object
as one rigid body. Rendering it gives the control video
(Sec.~\ref{sec:video_training}).

\section{Experiments}
\label{sec:experiments}

\begin{table}[t]
    \centering
    \caption{\textbf{Quantitative comparison on joint camera and object
    control.} (a) Visual quality and text alignment (FID, FVD, CLIP-SIM),
    camera control (RotErr, TransErr) and object control (recognition rate,
    IG-IoU). Best \textbf{bold}, second best \underline{underlined}. (b)
    Identity-Gated IoU on one frame: a vision--language judge decides whether
    the generated object is still the same object; a valid frame contributes
    its mask IoU and an invalid frame zero (Eq.~\ref{eq:igiou}).}
    \label{tab:main_comparison}
    \vspace{0.2em}
    \begin{subtable}[b]{0.583\linewidth}
        \centering
        \setlength{\tabcolsep}{2.8pt}        \renewcommand{\arraystretch}{1.3}
        \resizebox{\linewidth}{!}{%
        \begin{tabular}{lccccccc}
            \toprule
            & \multicolumn{3}{c}{Visual quality / text}
            & \multicolumn{2}{c}{Camera control}
            & \multicolumn{2}{c}{Object control} \\
            \cmidrule(lr){2-4}\cmidrule(lr){5-6}\cmidrule(lr){7-8}
            Method & FID & FVD & CLIP & Rot. & Trans. & Rec. & IG-IoU \\
            & $\downarrow$ & $\downarrow$ & $\uparrow$ & ($^\circ$) $\downarrow$
            & $\downarrow$ & $\uparrow$ & $\uparrow$ \\
            \midrule
            MotionCtrl
            & 115.6 & 1653.3 & 27.5 & 4.74 & 0.231 & 32.6 & 18.8 \\
            PaC
            & 55.7 & 1454.6 & 30.2 & 7.00 & 0.228 & 80.5 & 40.8 \\
            VerseCrafter
            & 55.0 & 484.7 & \underline{31.7} & 6.16 & 0.174 & \underline{94.8} & \underline{54.8} \\
            SymphoMotion
            & \underline{45.8} & \underline{405.4} & 30.9 & \underline{3.89} & \underline{0.132} & 90.2 & 52.9 \\
            \midrule
            \textbf{\method}
            & \textbf{44.1} & \textbf{370.4} & \textbf{31.8} & \textbf{3.65} & \textbf{0.122} & \textbf{94.9} & \textbf{60.4} \\
            \bottomrule
        \end{tabular}}
        \subcaption{Quantitative comparison}
        \label{tab:main_numbers}
    \end{subtable}\hfill    \begin{subtable}[b]{0.325\linewidth}
        \centering
        \includegraphics[width=\linewidth]{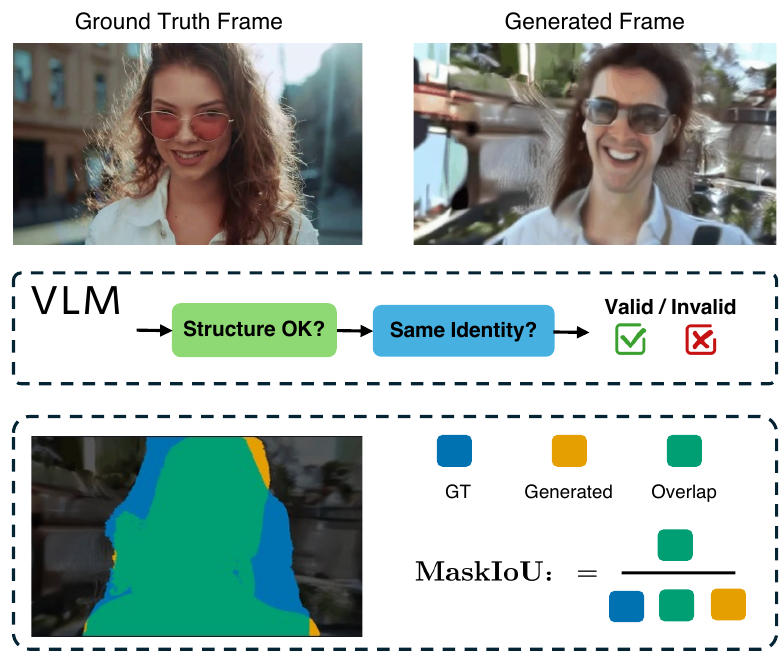}
        \subcaption{Identity-Gated IoU}
        \label{tab:igiou_panel}
    \end{subtable}
\end{table}

\begin{table}[t]
    \centering
    \caption{\textbf{VBench-I2V and user study.} VBench-I2V scores on the eight applicable dimensions (percent) and mean user ratings of visual quality, control accuracy and
    consistency (1 to 5, Sec.~\ref{sec:user_study}). \method{} leads on every dimension; VBench margins over the next
    best stay within 2 points, whereas the user study separates them clearly. Best \textbf{bold}, second best \underline{underlined}.}
    \label{tab:vbench}
    \vspace{0.15em}
    \setlength{\tabcolsep}{4pt}
    \renewcommand{\arraystretch}{1.0}
    \resizebox{\linewidth}{!}{%
    \begin{tabular}{lcccccccccccc}
        \toprule
        & \multicolumn{8}{c}{VBench-I2V}
        & \multicolumn{3}{c}{User study (1--5)} \\
        \cmidrule(lr){2-9}\cmidrule(lr){10-12}
        Method & Subject & Background & Motion & Dynamic & Aesthetic
        & Imaging & I2V & I2V & Quality & Control & Consistency \\
        & consistency & consistency & smoothness & degree & quality
        & quality & subject & background & $\uparrow$ & $\uparrow$ & $\uparrow$ \\
        \midrule
        MotionCtrl
        & 70.8 & 84.3 & 92.9 & \underline{94.0} & 43.8 & 45.7 & 81.6 & 84.2
        & 1.49 & 1.44 & 1.79 \\
        PaC
        & \underline{90.0} & 91.6 & 95.1 & 93.0 & 55.0 & 61.6 & 92.4 & 92.5
        & 1.98 & 2.01 & 2.41 \\
        VerseCrafter
        & 88.9 & \underline{91.9} & \underline{98.7} & 91.0 & \underline{56.7} & \underline{64.2} & \underline{96.0} & \underline{96.8}
        & \underline{3.69} & \underline{2.39} & \underline{3.88} \\
        SymphoMotion
        & 89.5 & 91.4 & 98.5 & 91.0 & 54.8 & 59.1 & 95.4 & 96.7
        & 3.35 & 2.33 & 3.69 \\
        \midrule
        \textbf{\method}
        & \textbf{90.3} & \textbf{92.0} & \textbf{98.8} & \textbf{96.0} & \textbf{57.0} & \textbf{64.6} & \textbf{96.3} & \textbf{97.0}
        & \textbf{4.62} & \textbf{4.82} & \textbf{4.69} \\
        \bottomrule
    \end{tabular}}
\end{table}

\subsection{Setup}
\label{sec:experimental_setup}

\noindent\textbf{Implementation details.} We initialize the Motion Adapter (3.0\,B parameters) from the released
VACE branch, perturb its attention and feed-forward matrices once, and train
all of it on the 20{,}774 pairs of RealCOD-Rigid at
$832\times480$ and 81 frames for three epochs on 24 GPUs with a global batch of
24, using AdamW with a peak learning rate of $5\times10^{-5}$ and 25 warmup
steps. At inference we use 20 sampling steps and a classifier-free guidance
scale of 5.0 for every case.

\noindent\textbf{Baselines.} We compare against four public methods for
joint camera and object control: MotionCtrl~\citep{motionctrl2024},
Perception-as-Control (PaC)~\citep{perceptioncontrol2025},
SymphoMotion~\citep{zhang2026symphomotion} and
VerseCrafter~\citep{zheng2026versecrafter}, each with its released weights,
resolution and clip length. To compare fairly, we give every method the same motion to follow: the
camera trajectory and object motion recovered from each evaluation clip are
converted into whatever control each baseline expects, from 2D trajectories
to 3D Gaussians.

\noindent\textbf{Metrics.} All methods are evaluated on the same 100 clips of
\mbox{RealCOD-25K} that are excluded from training, with joint camera and object motion, using four groups
of metrics. (1) \emph{Visual quality}: FID~\citep{heusel2017ttur},
FVD~\citep{unterthiner2018fvd} and the eight dimensions of
VBench-I2V~\citep{huang2024vbench} that apply to our videos.
(2) \emph{Text alignment}: CLIP-SIM~\citep{radford2021clip}.
(3) \emph{Camera control}: following CameraCtrl~\citep{he2024cameractrl},
the rotation error (RotErr) and relative translation error (TransErr) between the target trajectory and the one re-estimated from the generated
video with the pipeline of Sec.~\ref{sec:shared_geometry}. (4) \emph{Object
control}: Identity-Gated IoU, defined next.

\subsection{Identity-Gated IoU}
\label{sec:igiou}

Mask IoU against the source clip's object masks measures whether the
generated object is placed where the trajectory prescribes, but it also
rewards an object that is correctly placed and no longer the same object.
Identity-Gated IoU (IG-IoU) credits placement only where identity is
preserved. Each video is scored on the same $N=16$ frames, spaced uniformly over the
frames in which the object is visible in the source clip. Let $\mathrm{MaskIoU}_f$ denote the
intersection over union of the generated and the reference mask in frame
$f$. The identity gate $v_f\in\{0,1\}$ is set by a vision--language judge that compares the generated crop with the reference crop~\citep{wu2026scenediff}: $v_f=1$ if the
object is intact and still the same object, and $v_f=0$ otherwise. We
report
\begin{equation}
    \text{Recognition rate}=\frac{1}{N}\sum_{f=1}^{N} v_f,
    \qquad
    \mathrm{IG\text{-}IoU}=\frac{1}{N}\sum_{f=1}^{N} v_f\,\mathrm{MaskIoU}_f,
    \label{eq:igiou}
\end{equation}
averaged over the evaluation clips: a frame that fails the gate contributes zero but stays in the denominator. Masks come from SAM3~\citep{carion2025sam3} and the judge is Qwen3-VL~\citep{bai2025qwen3vl} (Table~\ref{tab:igiou_panel}; protocol in the appendix).
Because the reference mask is the object's silhouette at its true pose,
MaskIoU also penalizes an orientation error that changes the silhouette; a
$180^\circ$ turn of a front--back symmetric object is not penalized and is
shown qualitatively (\cref{fig:control_representations,fig:more_rotation}).

\subsection{Joint Camera and Object Control}
\label{sec:main_comparison}

\begin{figure}[t]
    \centering
    \includegraphics[width=\linewidth]{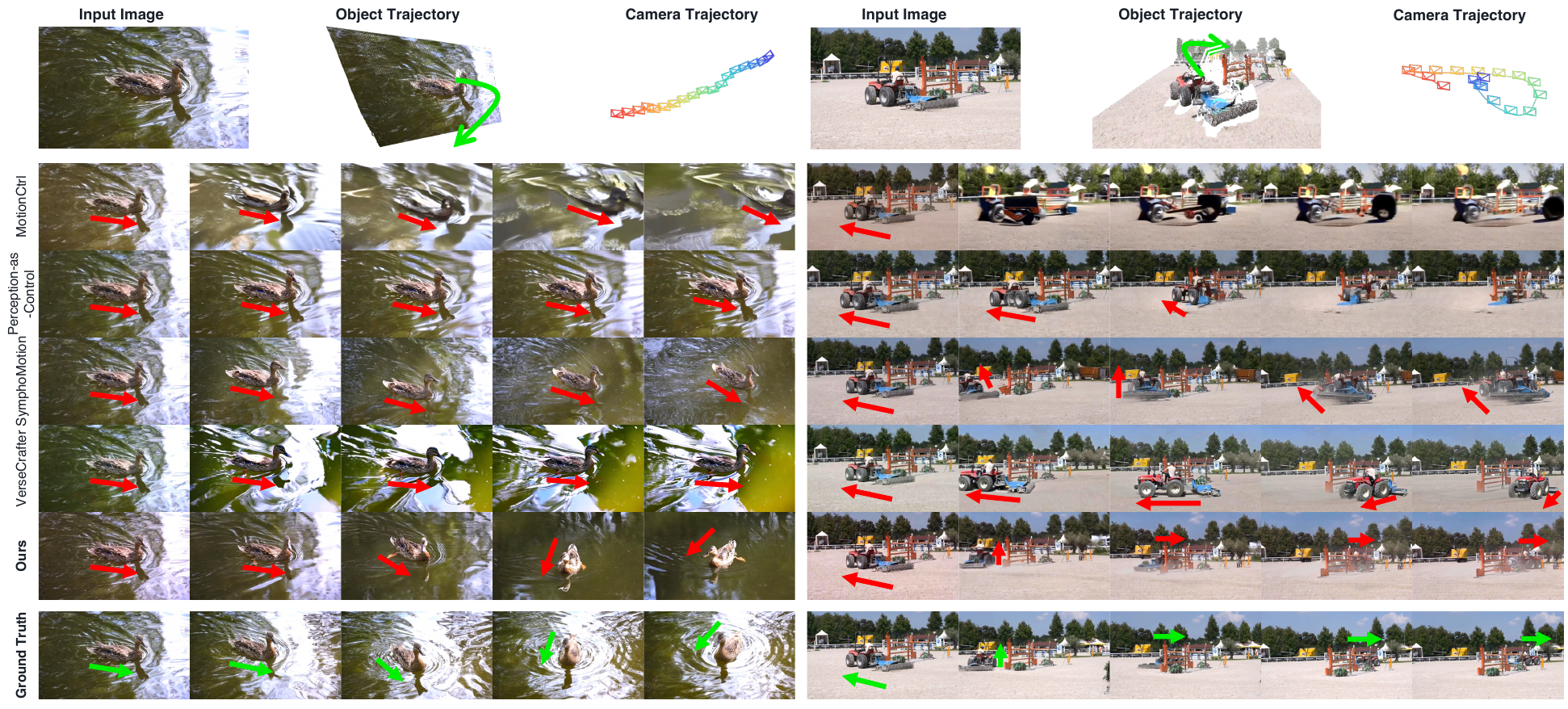}
    \caption{\textbf{Qualitative comparison on joint camera and object
    control.} For each clip, the top row shows the input image and the
    prescribed trajectories, and the rows below show temporally aligned
    frames from the four baselines, \method{} and the source video.
    Arrows mark the object heading, green in the source and red in the
    generated videos. Only \method{} completes the turn of the source video;
    the baselines keep the original heading, turn only partway, blur or lose
    the subject.}
    \label{fig:qualitative_comparison}
\end{figure}

\noindent\textbf{Quantitative comparison.} As shown in
Table~\ref{tab:main_numbers}, \method{} achieves the best score on every
metric. (1) \emph{Visual quality and text alignment}: it lowers FID by
3.7\% and FVD by 8.6\% relative to SymphoMotion, the next best on both, and matches VerseCrafter on CLIP-SIM. (2) \emph{Camera control}: it reduces the
rotation error from $3.89^\circ$ for SymphoMotion, the next best, to
$3.65^\circ$ and the translation error by 7.6\%, with the other three
baselines clearly behind. (3) \emph{Object control}, where the margins are
widest: IG-IoU rises from 54.8 for VerseCrafter, the next best, to 60.4, a
10.2\% relative gain at the same recognition rate (94.9 against 94.8).
\method{} thus places the object more accurately rather than keeping it
recognizable more often, which we attribute to the complete mesh fixing its
extent and orientation, as a Gaussian blob or tracked points
do not (\cref{fig:control_representations}); MotionCtrl, driven by 2D
trajectories, keeps the object recognizable in only 32.6\% of frames, so
its IG-IoU drops to 18.8.

\noindent\textbf{Qualitative comparison.}
Figure~\ref{fig:qualitative_comparison} shows two further clips in which the
object turns while the camera moves; the arrows mark its heading. Only
\method{} turns the object as far as the source video does. On the duck,
Perception-as-Control and VerseCrafter keep the original heading, and
MotionCtrl and SymphoMotion turn it only slightly; on the tractor,
SymphoMotion and VerseCrafter turn it only halfway, and Perception-as-Control
and MotionCtrl lose and blur it, respectively. The tractor also passes behind
the fence in the source video; \method{} reproduces this occlusion, whereas
SymphoMotion and VerseCrafter keep the tractor in front (see also
Appendix~\ref{sec:occlusion}).

\noindent\textbf{User study.}
\label{sec:user_study}
To complement the automatic metrics, \nparticipants{} participants rated
the five methods on nine cases from 1 to 5
on \emph{visual quality}, \emph{control accuracy} and \emph{consistency} (protocol in Appendix~\ref{sec:user_study_details}). As
Table~\ref{tab:vbench} shows, \method{} receives the highest rating on all
three, 4.6 to 4.8 against at most 3.9 for the next best, with
the largest margin on control accuracy (4.82 against 2.39).

\subsection{Objects Leaving and Re-entering the View}
\label{sec:out_of_view}

\begin{figure}[t]
    \centering
    \includegraphics[width=\linewidth]{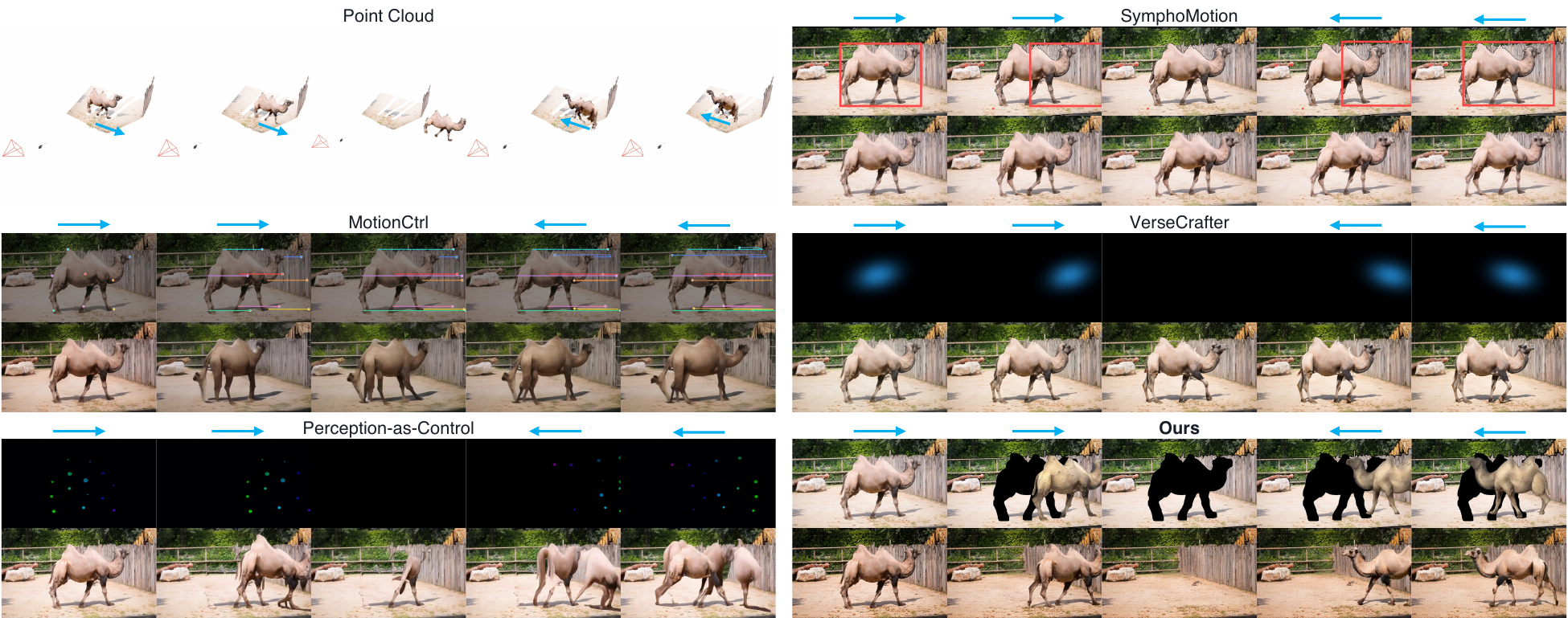}
    \caption{\textbf{Objects leaving and re-entering the view.} \emph{Top left:}
    the partial point cloud lifted from the input image; the camel walks out of
    the camera's view and comes back (blue arrows) while the camera stays
    fixed. For each method, the upper row is the control it receives and
    the lower row the video generated from it. Only \method{} shows the empty
    frame while the camel is away and then brings the same camel back.}
    \label{fig:out_of_view}
\end{figure}

\begin{figure}[t]
    \centering
    \includegraphics[width=\linewidth]{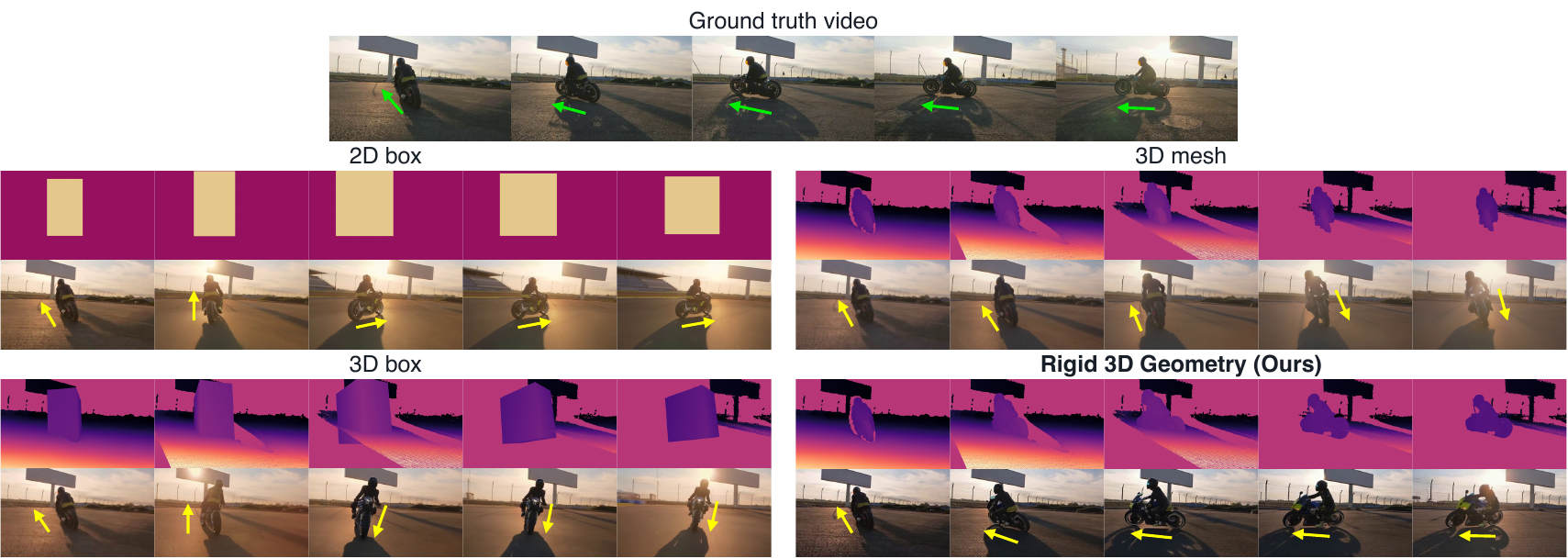}\\[3pt]
    \small\setlength{\tabcolsep}{3.5pt}
    \renewcommand{\arraystretch}{1.0}
    \resizebox{\linewidth}{!}{%
    \begin{tabular}{lccccccccccc}
        \toprule
        & \multicolumn{3}{c}{Control encodes}
        & \multicolumn{3}{c}{Visual quality / text}
        & \multicolumn{2}{c}{Camera control}
        & \multicolumn{2}{c}{Object control} \\
        \cmidrule(lr){2-4}\cmidrule(lr){5-7}\cmidrule(lr){8-9}\cmidrule(lr){10-11}
        Control & Depth & Rotation & Complete geometry
        & FID $\downarrow$ & FVD $\downarrow$ & CLIP $\uparrow$
        & Rot.\ ($^\circ$) $\downarrow$ & Trans.\ $\downarrow$
        & Rec.\ $\uparrow$ & IG-IoU $\uparrow$ \\
        \midrule
        2D box                &        &        &        & 58.3 & 608.8 & 31.2 & 9.61 & 0.233 & \textbf{95.0} & 48.5 \\
        3D box                & \cmark & \cmark &        & 54.7 & 556.0 & 31.1 & 5.00 & \underline{0.142} & 93.3 & 50.3 \\
        3D mesh               & \cmark &        & \cmark & \underline{53.0} & \underline{485.6} & \underline{31.2} & \underline{4.84} & 0.147 & 91.9 & \underline{53.0} \\
        \midrule
        \textbf{Rigid 3D geometry (ours)} & \cmark & \cmark & \cmark
            & \textbf{44.1} & \textbf{370.4} & \textbf{31.8} & \textbf{3.65} & \textbf{0.122} & \underline{94.9} & \textbf{60.4} \\
        \bottomrule
    \end{tabular}}
    \caption{\textbf{Ablation on the control representation.} \emph{Top:} the source video and three
    alternative controls, each shown as the control video above the video
    generated from it; arrows mark the motorcycle heading, and only the rigid
    3D geometry follows the prescribed turn. \emph{Bottom:} what each control
    encodes and its scores; the three arms are trained with the recipe of
    \method{}, and the last row is \method{} as in
    Table~\ref{tab:main_numbers}. Best \textbf{bold}, second best
    \underline{underlined}.}
    \label{fig:ablation_control}
\end{figure}

We further evaluate a case in which an object moves out of the camera view
and later returns (Fig.~\ref{fig:out_of_view}). The model must then keep the
identity of the object while it is not visible. In \method{}, each object is
a mesh with a trajectory defined at every frame, so the object is simply not
rendered while it is outside the view and is rendered again from the same
mesh when it returns. As shown in Fig.~\ref{fig:out_of_view}, \method{}
removes the camel from the frame while it is away, fills in the background it
occupied, and brings back the same camel at the specified position. In
contrast, MotionCtrl, SymphoMotion and VerseCrafter keep the camel in the
frame throughout, and Perception-as-Control fails to preserve its shape when
it returns.

\subsection{Ablation Study}
\label{sec:ablation_control}

Figure~\ref{fig:ablation_control} replaces the rigid 3D geometry by three
alternative controls. The 3D box and the 3D mesh each drop one of their two
ingredients: the 3D box keeps depth and orientation but has no surface, and
the 3D mesh keeps the complete surface but does not rotate. Dropping either
lowers IG-IoU from 60.4 to 50.3 without the surface and to 53.0 without
rotation, while the recognition rate stays between 91.9 and 95.0 for every
control; the loss is thus mainly in placement, not identity. In the top
panel, only the rigid 3D geometry follows the prescribed turn: with the mesh
that does not rotate, the motorcycle ends up facing the camera instead of
turning sideways, an orientation error that also changes its silhouette.
The 2D box, which encodes neither depth nor orientation, has the lowest
IG-IoU (48.5), and the four controls rank in the same order on FID, FVD, and
IG-IoU.

\section{Conclusion and Future Work}
\label{sec:conclusion}

We present \method{}, a video world model controlled with rigid 3D
geometry: an explicit 4D scene representation in which every object is a
complete canonical mesh moved by one rigid transformation per frame, sharing
one coordinate frame with the background and the camera. The Motion Adapter injects a depth rendering of this scene, which carries the camera and the rigid motion of every object, together with the input image into a pretrained video generator that supplies appearance, illumination and non-rigid dynamics. To train the adapter, we built \mbox{RealCOD-Rigid}, a dataset of 20{,}774 real video clips whose rigid 3D scenes, with complete object meshes, per-frame rigid transformations and camera trajectories, are recovered by an automatic annotation pipeline. To evaluate object control, we introduced Identity-Gated IoU, which credits the mask IoU of a generated object only in frames where the object keeps its identity.
Experiments, ablations, and a user study show higher visual quality and more
accurate camera and object control than four state-of-the-art baselines. The representation, however, controls motion only at the level of a rigid
body: an articulated or deforming object moves as one whole, so finer motion
such as running, jumping, or the movement of the limbs cannot be prescribed
and is decided by the generator, which may even keep a mostly articulated
object in its first-frame pose (Appendix~\ref{sec:failure}). Extending the rigid 3D scene with articulated parts, so that such motion can be prescribed, is a natural next step.

\bibliography{main}

\begin{thebibliography}{128}
\providecommand{\natexlab}[1]{#1}
\providecommand{\url}[1]{\texttt{#1}}
\expandafter\ifx\csname urlstyle\endcsname\relax
  \providecommand{\doi}[1]{doi: #1}\else
  \providecommand{\doi}{doi: \begingroup \urlstyle{rm}\Url}\fi

\bibitem[Agarwal et~al.(2025)Agarwal, Ali, Bala, Balaji, Barker, Cai,
  Chattopadhyay, Chen, Cui, Ding, et~al.]{agarwal2025cosmos}
Niket Agarwal, Arslan Ali, Maciej Bala, Yogesh Balaji, Erik Barker, Tiffany
  Cai, Prithvijit Chattopadhyay, Yongxin Chen, Yin Cui, Yifan Ding, et~al.
\newblock Cosmos world foundation model platform for physical ai.
\newblock \emph{arXiv preprint arXiv:2501.03575}, 2025.

\bibitem[Alhaija et~al.(2025)Alhaija, Alvarez, Bala, Cai, Cao, Cha, Chen, Chen,
  Ferroni, Fidler, et~al.]{alhaija2025cosmos}
Hassan~Abu Alhaija, Jose Alvarez, Maciej Bala, Tiffany Cai, Tianshi Cao, Liz
  Cha, Joshua Chen, Mike Chen, Francesco Ferroni, Sanja Fidler, et~al.
\newblock Cosmos-transfer1: Conditional world generation with adaptive
  multimodal control.
\newblock \emph{arXiv preprint arXiv:2503.14492}, 2025.

\bibitem[Alonso et~al.(2024)Alonso, Jelley, Micheli, Kanervisto, Storkey,
  Pearce, and Fleuret]{diamond}
Eloi Alonso, Adam Jelley, Vincent Micheli, Anssi Kanervisto, Amos Storkey, Tim
  Pearce, and Fran{\c{c}}ois Fleuret.
\newblock Diffusion for world modeling: Visual details matter in atari.
\newblock In \emph{Advances in Neural Information Processing Systems
  (NeurIPS)}, 2024.

\bibitem[Bahmani et~al.(2024)Bahmani, Skorokhodov, Siarohin, Menapace, Qian,
  Vasilkovsky, Lee, Wang, Zou, Tagliasacchi, et~al.]{bahmani2024vd3d}
Sherwin Bahmani, Ivan Skorokhodov, Aliaksandr Siarohin, Willi Menapace,
  Guocheng Qian, Michael Vasilkovsky, Hsin-Ying Lee, Chaoyang Wang, Jiaxu Zou,
  Andrea Tagliasacchi, et~al.
\newblock Vd3d: Taming large video diffusion transformers for 3d camera
  control.
\newblock \emph{arXiv preprint arXiv:2407.12781}, 2024.

\bibitem[Bahmani et~al.(2025)Bahmani, Skorokhodov, Qian, Siarohin, Menapace,
  Tagliasacchi, Lindell, and Tulyakov]{bahmani2025ac3d}
Sherwin Bahmani, Ivan Skorokhodov, Guocheng Qian, Aliaksandr Siarohin, Willi
  Menapace, Andrea Tagliasacchi, David~B. Lindell, and Sergey Tulyakov.
\newblock {AC3D}: Analyzing and improving {3D} camera control in video
  diffusion transformers.
\newblock In \emph{IEEE/CVF Conference on Computer Vision and Pattern
  Recognition (CVPR)}, pp.\  22875--22889, 2025.

\bibitem[Bai et~al.(2024)Bai, Xia, Wang, Yuan, Fu, Liu, Hu, Wan, and
  Zhang]{bai2024syncammaster}
Jianhong Bai, Menghan Xia, Xintao Wang, Ziyang Yuan, Xiao Fu, Zuozhu Liu, Haoji
  Hu, Pengfei Wan, and Di~Zhang.
\newblock Syncammaster: Synchronizing multi-camera video generation from
  diverse viewpoints.
\newblock \emph{arXiv preprint arXiv:2412.07760}, 2024.

\bibitem[Bai et~al.(2025{\natexlab{a}})Bai, Xia, Fu, Wang, Mu, Cao, Liu, Hu,
  Bai, Wan, et~al.]{bai2025recammaster}
Jianhong Bai, Menghan Xia, Xiao Fu, Xintao Wang, Lianrui Mu, Jinwen Cao, Zuozhu
  Liu, Haoji Hu, Xiang Bai, Pengfei Wan, et~al.
\newblock Recammaster: Camera-controlled generative rendering from a single
  video.
\newblock In \emph{IEEE/CVF International Conference on Computer Vision
  (ICCV)}, 2025{\natexlab{a}}.

\bibitem[Bai et~al.(2025{\natexlab{b}})Bai, Cai, Chen, Chen,
  et~al.]{bai2025qwen3vl}
Shuai Bai, Yuxuan Cai, Ruizhe Chen, Keqin Chen, et~al.
\newblock {Qwen3-VL} technical report.
\newblock \emph{arXiv preprint arXiv:2511.21631}, 2025{\natexlab{b}}.

\bibitem[Blattmann et~al.(2023)Blattmann, Dockhorn, Kulal, Mendelevitch,
  Kilian, Lorenz, Levi, English, Voleti, Letts, et~al.]{blattmann2023stable}
Andreas Blattmann, Tim Dockhorn, Sumith Kulal, Daniel Mendelevitch, Maciej
  Kilian, Dominik Lorenz, Yam Levi, Zion English, Vikram Voleti, Adam Letts,
  et~al.
\newblock Stable video diffusion: Scaling latent video diffusion models to
  large datasets.
\newblock \emph{arXiv preprint arXiv:2311.15127}, 2023.

\bibitem[Bruce et~al.(2024)Bruce, Dennis, Edwards, Parker-Holder, Shi, Hughes,
  Lai, Mavalankar, Steigerwald, Apps, et~al.]{bruce2024genie}
Jake Bruce, Michael Dennis, Ashley Edwards, Jack Parker-Holder, Yuge Shi,
  Edward Hughes, Matthew Lai, Aditi Mavalankar, Richie Steigerwald, Chris Apps,
  et~al.
\newblock Genie: Generative interactive environments.
\newblock In \emph{International Conference on Machine Learning (ICML)}, 2024.

\bibitem[Burgert et~al.(2025)Burgert, Xu, Xian, Pilarski, Clausen, He, Ma,
  Deng, Li, Mousavi, et~al.]{burgert2025go}
Ryan Burgert, Yuancheng Xu, Wenqi Xian, Oliver Pilarski, Pascal Clausen,
  Mingming He, Li~Ma, Yitong Deng, Lingxiao Li, Mohsen Mousavi, et~al.
\newblock Go-with-the-flow: Motion-controllable video diffusion models using
  real-time warped noise.
\newblock In \emph{IEEE/CVF Conference on Computer Vision and Pattern
  Recognition (CVPR)}, pp.\  13--23, 2025.

\bibitem[Cao et~al.(2025{\natexlab{a}})Cao, Zhou, Li, Liang, Yu, Wang, Xue, and
  Fu]{cao2025uni3c}
Chenjie Cao, Jingkai Zhou, Shikai Li, Jingyun Liang, Chaohui Yu, Fan Wang,
  Xiangyang Xue, and Yanwei Fu.
\newblock {Uni3C}: Unifying precisely {3D}-enhanced camera and human motion
  controls for video generation.
\newblock In \emph{ACM SIGGRAPH Asia 2025 Conference Papers}, pp.\
  21:1--21:12, 2025{\natexlab{a}}.

\bibitem[Cao et~al.(2024)Cao, Luo, Zhang, Nie{\ss}ner, and
  Tang]{cao2024motion2vecsets}
Wei Cao, Chang Luo, Biao Zhang, Matthias Nie{\ss}ner, and Jiapeng Tang.
\newblock {Motion2VecSets}: {4D} latent vector set diffusion for non-rigid
  shape reconstruction and tracking.
\newblock In \emph{IEEE/CVF Conference on Computer Vision and Pattern
  Recognition (CVPR)}, pp.\  20496--20506, 2024.

\bibitem[Cao et~al.(2025{\natexlab{b}})Cao, Hallgarten, Li, Dauner, Gu, Wang,
  Miron, Aiello, Li, Gilitschenski, et~al.]{cao2025pseudo}
Wei Cao, Marcel Hallgarten, Tianyu Li, Daniel Dauner, Xunjiang Gu, Caojun Wang,
  Yakov Miron, Marco Aiello, Hongyang Li, Igor Gilitschenski, et~al.
\newblock Pseudo-simulation for autonomous driving.
\newblock \emph{arXiv preprint arXiv:2506.04218}, 2025{\natexlab{b}}.

\bibitem[Cao et~al.(2026)Cao, Zhang, Tian, Wu, Li, Wang, Yu, and
  Liu]{cao2026freeorbit4d}
Wei Cao, Hao Zhang, Fengrui Tian, Yulun Wu, Yingying Li, Shenlong Wang, Ning
  Yu, and Yaoyao Liu.
\newblock {FreeOrbit4D}: Training-free arbitrary camera redirection for
  monocular videos via foreground-complete {4D} reconstruction.
\newblock In \emph{ACM SIGGRAPH 2026 Conference Papers}, pp.\  65:1--65:12,
  2026.

\bibitem[Carion et~al.(2025)Carion, Gustafson, Hu, Debnath,
  et~al.]{carion2025sam3}
Nicolas Carion, Laura Gustafson, Yuan-Ting Hu, Shoubhik Debnath, et~al.
\newblock {SAM 3}: Segment anything with concepts.
\newblock \emph{arXiv preprint arXiv:2511.16719}, 2025.

\bibitem[Che et~al.(2025)Che, He, Liu, Jin, and Chen]{che2024gamegen}
Haoxuan Che, Xuanhua He, Quande Liu, Cheng Jin, and Hao Chen.
\newblock Gamegen-x: Interactive open-world game video generation.
\newblock In \emph{International Conference on Learning Representations
  (ICLR)}, 2025.

\bibitem[Chen et~al.(2025{\natexlab{a}})Chen, Zhu, He, Wang, Zhou, Chang, Zhou,
  Li, Fu, Pang, et~al.]{chen2025deepverse}
Junyi Chen, Haoyi Zhu, Xianglong He, Yifan Wang, Jianjun Zhou, Wenzheng Chang,
  Yang Zhou, Zizun Li, Zhoujie Fu, Jiangmiao Pang, et~al.
\newblock Deepverse: 4d autoregressive video generation as a world model.
\newblock \emph{arXiv preprint arXiv:2506.01103}, 2025{\natexlab{a}}.

\bibitem[Chen et~al.(2026)Chen, Liang, Zhou, Ding, Liu, Wan, and Bai]{hydra}
Kaijin Chen, Dingkang Liang, Xin Zhou, Yikang Ding, Xiaoqiang Liu, Pengfei Wan,
  and Xiang Bai.
\newblock Out of sight but not out of mind: Hybrid memory for dynamic video
  world models.
\newblock \emph{arXiv preprint arXiv:2603.25716}, 2026.

\bibitem[Chen et~al.(2025{\natexlab{b}})Chen, Men, Yao, Cui, and
  Bo]{perceptioncontrol2025}
Yingjie Chen, Yifang Men, Yuan Yao, Miaomiao Cui, and Liefeng Bo.
\newblock Perception-as-control: Fine-grained controllable image animation with
  3d-aware motion representation.
\newblock In \emph{IEEE/CVF International Conference on Computer Vision
  (ICCV)}, 2025{\natexlab{b}}.

\bibitem[Chu et~al.(2025)Chu, He, Chen, Zhang, Xu, Xia, Wang, Yi, Liu, Zhao,
  et~al.]{chu2025wan}
Ruihang Chu, Yefei He, Zhekai Chen, Shiwei Zhang, Xiaogang Xu, Bin Xia,
  Dingdong Wang, Hongwei Yi, Xihui Liu, Hengshuang Zhao, et~al.
\newblock Wan-move: Motion-controllable video generation via latent trajectory
  guidance.
\newblock \emph{arXiv preprint arXiv:2512.08765}, 2025.

\bibitem[Chung et~al.(2023)Chung, Constant, Garcia, Roberts, Tay, Narang, and
  Firat]{chung2023unimax}
Hyung~Won Chung, Noah Constant, Xavier Garcia, Adam Roberts, Yi~Tay, Sharan
  Narang, and Orhan Firat.
\newblock {UniMax}: Fairer and more effective language sampling for large-scale
  multilingual pretraining.
\newblock In \emph{International Conference on Learning Representations
  (ICLR)}, 2023.

\bibitem[{Decart} et~al.(2024){Decart}, Quevedo, McIntyre, Campbell, Chen, and
  Wachen]{oasis}
{Decart}, Julian Quevedo, Quinn McIntyre, Spruce Campbell, Xinlei Chen, and
  Robert Wachen.
\newblock Oasis: A universe in a transformer.
\newblock Project page, 2024.
\newblock URL \url{https://oasis-model.github.io/}.

\bibitem[Duan et~al.(2026)Duan, Xia, Zhang, Zhang, Zhou, Gou, He, Chen, Zhang,
  and Liu]{liveworld}
Zicheng Duan, Jiatong Xia, Zeyu Zhang, Wenbo Zhang, Gengze Zhou, Chenhui Gou,
  Yefei He, Feng Chen, Xinyu Zhang, and Lingqiao Liu.
\newblock {LiveWorld}: Simulating out-of-sight dynamics in generative video
  world models.
\newblock \emph{arXiv preprint arXiv:2603.07145}, 2026.

\bibitem[Feng et~al.(2024{\natexlab{a}})Feng, Liu, Tu, Qi, Sun, Ma, Zhao, Zhou,
  and He]{feng2024i2vcontrolcamera}
Wanquan Feng, Jiawei Liu, Pengqi Tu, Tianhao Qi, Mingzhen Sun, Tianxiang Ma,
  Songtao Zhao, Siyu Zhou, and Qian He.
\newblock I2vcontrol-camera: Precise video camera control with adjustable
  motion strength.
\newblock \emph{arXiv preprint arXiv:2411.06525}, 2024{\natexlab{a}}.

\bibitem[Feng et~al.(2024{\natexlab{b}})Feng, Qi, Liu, Sun, Tu, Ma, Dai, Zhao,
  Zhou, and He]{feng2024i2vcontrol}
Wanquan Feng, Tianhao Qi, Jiawei Liu, Mingzhen Sun, Pengqi Tu, Tianxiang Ma,
  Fei Dai, Songtao Zhao, Siyu Zhou, and Qian He.
\newblock I2vcontrol: Disentangled and unified video motion synthesis control.
\newblock \emph{arXiv preprint arXiv:2411.17765}, 2024{\natexlab{b}}.

\bibitem[Fu et~al.(2025)Fu, Liu, Wang, Peng, Xia, Shi, Yuan, Wan, Zhang, and
  Lin]{fu20243dtrajmaster}
Xiao Fu, Xian Liu, Xintao Wang, Sida Peng, Menghan Xia, Xiaoyu Shi, Ziyang
  Yuan, Pengfei Wan, Di~Zhang, and Dahua Lin.
\newblock {3DTrajMaster}: Mastering {3D} trajectory for multi-entity motion in
  video generation.
\newblock In \emph{International Conference on Learning Representations
  (ICLR)}, 2025.

\bibitem[Geng et~al.(2025)Geng, Herrmann, Hur, Cole, Zhang, Pfaff,
  Lopez-Guevara, Aytar, Rubinstein, Sun, et~al.]{geng2025motion}
Daniel Geng, Charles Herrmann, Junhwa Hur, Forrester Cole, Serena Zhang, Tobias
  Pfaff, Tatiana Lopez-Guevara, Yusuf Aytar, Michael Rubinstein, Chen Sun,
  et~al.
\newblock Motion prompting: Controlling video generation with motion
  trajectories.
\newblock In \emph{IEEE/CVF Conference on Computer Vision and Pattern
  Recognition (CVPR)}, 2025.

\bibitem[Gu et~al.(2025)Gu, Yan, Lu, Li, Dou, Si, Dong, Liu, Lin, Liu,
  et~al.]{gu2025diffusion}
Zekai Gu, Rui Yan, Jiahao Lu, Peng Li, Zhiyang Dou, Chenyang Si, Zhen Dong,
  Qifeng Liu, Cheng Lin, Ziwei Liu, et~al.
\newblock Diffusion as shader: 3d-aware video diffusion for versatile video
  generation control.
\newblock In \emph{ACM SIGGRAPH 2025 Conference Papers}, pp.\  1--12, 2025.

\bibitem[Ha \& Schmidhuber(2018)Ha and Schmidhuber]{ha2018world}
David Ha and J{\"u}rgen Schmidhuber.
\newblock World models.
\newblock \emph{arXiv preprint arXiv:1803.10122}, 2\penalty0 (3), 2018.

\bibitem[Hafner et~al.(2023)Hafner, Pasukonis, Ba, and
  Lillicrap]{hafner2023mastering}
Danijar Hafner, Jurgis Pasukonis, Jimmy Ba, and Timothy Lillicrap.
\newblock Mastering diverse domains through world models.
\newblock \emph{arXiv preprint arXiv:2301.04104}, 2023.

\bibitem[He et~al.(2024)He, Xu, Guo, Wetzstein, Dai, Li, and
  Yang]{he2024cameractrl}
Hao He, Yinghao Xu, Yuwei Guo, Gordon Wetzstein, Bo~Dai, Hongsheng Li, and
  Ceyuan Yang.
\newblock Cameractrl: Enabling camera control for text-to-video generation.
\newblock \emph{arXiv preprint arXiv:2404.02101}, 2024.

\bibitem[He et~al.(2025{\natexlab{a}})He, Yang, Lin, Xu, Wei, Gui, Zhao,
  Wetzstein, Jiang, and Li]{he2025cameractrl}
Hao He, Ceyuan Yang, Shanchuan Lin, Yinghao Xu, Meng Wei, Liangke Gui, Qi~Zhao,
  Gordon Wetzstein, Lu~Jiang, and Hongsheng Li.
\newblock Cameractrl ii: Dynamic scene exploration via camera-controlled video
  diffusion models.
\newblock \emph{arXiv preprint arXiv:2503.10592}, 2025{\natexlab{a}}.

\bibitem[He et~al.(2025{\natexlab{b}})He, Peng, Liu, Wang, Zhang, Cui, Kang,
  Jiang, An, Ren, et~al.]{he2025matrix}
Xianglong He, Chunli Peng, Zexiang Liu, Boyang Wang, Yifan Zhang, Qi~Cui, Fei
  Kang, Biao Jiang, Mengyin An, Yangyang Ren, et~al.
\newblock Matrix-game 2.0: An open-source, real-time, and streaming interactive
  world model.
\newblock \emph{arXiv preprint arXiv:2508.13009}, 2025{\natexlab{b}}.

\bibitem[Henschel et~al.(2025)Henschel, Khachatryan, Poghosyan, Hayrapetyan,
  Tadevosyan, Wang, Navasardyan, and Shi]{streamingt2v}
Roberto Henschel, Levon Khachatryan, Hayk Poghosyan, Daniil Hayrapetyan, Vahram
  Tadevosyan, Zhangyang Wang, Shant Navasardyan, and Humphrey Shi.
\newblock {StreamingT2V}: Consistent, dynamic, and extendable long video
  generation from text.
\newblock In \emph{IEEE/CVF Conference on Computer Vision and Pattern
  Recognition (CVPR)}, 2025.

\bibitem[Heusel et~al.(2017)Heusel, Ramsauer, Unterthiner, Nessler, and
  Hochreiter]{heusel2017ttur}
Martin Heusel, Hubert Ramsauer, Thomas Unterthiner, Bernhard Nessler, and Sepp
  Hochreiter.
\newblock {GANs} trained by a two time-scale update rule converge to a local
  nash equilibrium.
\newblock In \emph{Advances in Neural Information Processing Systems
  (NeurIPS)}, 2017.

\bibitem[Hong et~al.(2025)Hong, Mei, Ge, Xu, Zhou, Bi, Hold-Geoffroy, Roberts,
  Fisher, Shechtman, et~al.]{relic}
Yicong Hong, Yiqun Mei, Chongjian Ge, Yiran Xu, Yang Zhou, Sai Bi, Yannick
  Hold-Geoffroy, Mike Roberts, Matthew Fisher, Eli Shechtman, et~al.
\newblock Relic: Interactive video world model with long-horizon memory.
\newblock \emph{arXiv preprint arXiv:2512.04040}, 2025.

\bibitem[Hou et~al.(2024)Hou, Wei, Zeng, and Chen]{hou2024training}
Chen Hou, Guoqiang Wei, Yan Zeng, and Zhibo Chen.
\newblock Training-free camera control for video generation.
\newblock \emph{arXiv preprint arXiv:2406.10126}, 2024.

\bibitem[Hu et~al.(2023)Hu, Russell, Yeo, Murez, Fedoseev, Kendall, Shotton,
  and Corrado]{hu2023gaia}
Anthony Hu, Lloyd Russell, Hudson Yeo, Zak Murez, George Fedoseev, Alex
  Kendall, Jamie Shotton, and Gianluca Corrado.
\newblock Gaia-1: A generative world model for autonomous driving.
\newblock \emph{arXiv preprint arXiv:2309.17080}, 2023.

\bibitem[Hu et~al.(2025)Hu, Peng, Liu, and Ma]{hu2025ex}
Tao Hu, Haoyang Peng, Xiao Liu, and Yuewen Ma.
\newblock Ex-4d: Extreme viewpoint 4d video synthesis via depth watertight
  mesh.
\newblock \emph{arXiv preprint arXiv:2506.05554}, 2025.

\bibitem[Huang et~al.(2025)Huang, Zheng, Wang, Liu, Wang, Wu, Jiang, Li, Lau,
  Zuo, et~al.]{huang2025voyager}
Tianyu Huang, Wangguandong Zheng, Tengfei Wang, Yuhao Liu, Zhenwei Wang, Junta
  Wu, Jie Jiang, Hui Li, Rynson~WH Lau, Wangmeng Zuo, et~al.
\newblock Voyager: Long-range and world-consistent video diffusion for
  explorable 3d scene generation.
\newblock \emph{arXiv preprint arXiv:2506.04225}, 2025.

\bibitem[Huang et~al.(2024)Huang, He, Yu, Zhang, Si, Jiang, Zhang, Wu, Jin,
  Chanpaisit, Wang, Chen, Wang, Lin, Qiao, and Liu]{huang2024vbench}
Ziqi Huang, Yinan He, Jiashuo Yu, Fan Zhang, Chenyang Si, Yuming Jiang, Yuanhan
  Zhang, Tianxing Wu, Qingyang Jin, Nattapol Chanpaisit, Yaohui Wang, Xinyuan
  Chen, Limin Wang, Dahua Lin, Yu~Qiao, and Ziwei Liu.
\newblock {VBench}: Comprehensive benchmark suite for video generative models.
\newblock In \emph{IEEE/CVF Conference on Computer Vision and Pattern
  Recognition (CVPR)}, pp.\  21807--21818, 2024.

\bibitem[Jain et~al.(2024)Jain, Nasery, Vineet, and Behl]{jain2024peekaboo}
Yash Jain, Anshul Nasery, Vibhav Vineet, and Harkirat Behl.
\newblock Peekaboo: Interactive video generation via masked-diffusion.
\newblock In \emph{IEEE/CVF Conference on Computer Vision and Pattern
  Recognition (CVPR)}, pp.\  8079--8088, 2024.

\bibitem[Ji et~al.(2026)Ji, Hu, Ma, Jain, Li, Do, and Liu]{ji2026ac3s}
Eric Ji, Qiran Hu, Wufei Ma, Sarthak Jain, Yingying Li, Minh~N Do, and Yaoyao
  Liu.
\newblock {AC3S}: Adaptive conditioning for {3D}-aware synthetic data
  generation.
\newblock In \emph{European Conference on Computer Vision (ECCV)}, pp.\
  510--525, 2026.

\bibitem[Jiang et~al.(2025)Jiang, Han, Mao, Zhang, Pan, and Liu]{vace2025}
Zeyinzi Jiang, Zhen Han, Chaojie Mao, Jingfeng Zhang, Yulin Pan, and Yu~Liu.
\newblock {VACE}: All-in-one video creation and editing.
\newblock In \emph{IEEE/CVF International Conference on Computer Vision
  (ICCV)}, 2025.

\bibitem[Kuang et~al.(2024)Kuang, Cai, He, Xu, Li, Guibas, and
  Wetzstein]{kuang2024collaborative}
Zhengfei Kuang, Shengqu Cai, Hao He, Yinghao Xu, Hongsheng Li, Leonidas~J
  Guibas, and Gordon Wetzstein.
\newblock Collaborative video diffusion: Consistent multi-video generation with
  camera control.
\newblock \emph{Advances in Neural Information Processing Systems (NeurIPS)},
  37:\penalty0 16240--16271, 2024.

\bibitem[LeCun(2022)]{lecun2022path}
Yann LeCun.
\newblock A path towards autonomous machine intelligence.
\newblock OpenReview, 2022.
\newblock Version 0.9.2.

\bibitem[Lei et~al.(2024)Lei, Wang, Li, Zhang, Wang, and
  Xu]{lei2024animateanything}
Guojun Lei, Chi Wang, Hong Li, Rong Zhang, Yikai Wang, and Weiwei Xu.
\newblock Animateanything: Consistent and controllable animation for video
  generation.
\newblock \emph{arXiv preprint arXiv:2411.10836}, 2024.

\bibitem[Li et~al.(2026)Li, Zhao, Chen, Hu, Guo, Zhang, Shan, and
  Hu]{li2026pixal3d}
Dong-Yang Li, Wang Zhao, Yuxin Chen, Wenbo Hu, Meng-Hao Guo, Fang-Lue Zhang,
  Ying Shan, and Shi-Min Hu.
\newblock {Pixal3D}: Pixel-aligned {3D} generation from images.
\newblock In \emph{ACM SIGGRAPH 2026 Conference Papers}, pp.\  8:1--8:12, 2026.

\bibitem[Li et~al.(2025{\natexlab{a}})Li, Tang, Xu, Wu, Zhou, Shao, Yu, Cao,
  and Lu]{li2025hunyuan}
Jiaqi Li, Junshu Tang, Zhiyong Xu, Longhuang Wu, Yuan Zhou, Shuai Shao, Tianbao
  Yu, Zhiguo Cao, and Qinglin Lu.
\newblock Hunyuan-gamecraft: High-dynamic interactive game video generation
  with hybrid history condition.
\newblock \emph{arXiv preprint arXiv:2506.17201}, 2025{\natexlab{a}}.

\bibitem[Li et~al.(2025{\natexlab{b}})Li, Xing, Wang, Zhang, Dai, and
  Wu]{li2025magicmotion}
Quanhao Li, Zhen Xing, Rui Wang, Hui Zhang, Qi~Dai, and Zuxuan Wu.
\newblock Magicmotion: Controllable video generation with dense-to-sparse
  trajectory guidance.
\newblock In \emph{IEEE/CVF International Conference on Computer Vision
  (ICCV)}, 2025{\natexlab{b}}.

\bibitem[Li et~al.(2025{\natexlab{c}})Li, Torr, Vedaldi, and Jakab]{vmem}
Runjia Li, Philip Torr, Andrea Vedaldi, and Tomas Jakab.
\newblock {VMem}: Consistent interactive video scene generation with
  surfel-indexed view memory.
\newblock In \emph{IEEE/CVF International Conference on Computer Vision
  (ICCV)}, 2025{\natexlab{c}}.

\bibitem[Li et~al.(2025{\natexlab{d}})Li, Zheng, Jiang, Wu, Lu, Lin, Li,
  et~al.]{li2025realcam}
Teng Li, Guangcong Zheng, Rui Jiang, Tao Wu, Yehao Lu, Yining Lin, Xi~Li,
  et~al.
\newblock Realcam-i2v: Real-world image-to-video generation with interactive
  complex camera control.
\newblock \emph{arXiv preprint arXiv:2502.10059}, 2025{\natexlab{d}}.

\bibitem[Li et~al.(2025{\natexlab{e}})Li, Wang, Zhang, Wang, Yuan, Xie, Shan,
  and Zou]{li2025image}
Yaowei Li, Xintao Wang, Zhaoyang Zhang, Zhouxia Wang, Ziyang Yuan, Liangbin
  Xie, Ying Shan, and Yuexian Zou.
\newblock Image conductor: Precision control for interactive video synthesis.
\newblock In \emph{AAAI Conference on Artificial Intelligence (AAAI)},
  2025{\natexlab{e}}.

\bibitem[Li et~al.(2025{\natexlab{f}})Li, Tucker, Cole, Wang, Jin, Ye,
  Kanazawa, Holynski, and Snavely]{megasam2025}
Zhengqi Li, Richard Tucker, Forrester Cole, Qianqian Wang, Linyi Jin, Vickie
  Ye, Angjoo Kanazawa, Aleksander Holynski, and Noah Snavely.
\newblock {MegaSaM}: Accurate, fast and robust structure and motion from casual
  dynamic videos.
\newblock In \emph{IEEE/CVF Conference on Computer Vision and Pattern
  Recognition (CVPR)}, 2025{\natexlab{f}}.

\bibitem[Liang et~al.(2025)Liang, Zhou, Li, Cao, Sun, Qian, Chen, and
  Wang]{liang2025realismotion}
Jingyun Liang, Jingkai Zhou, Shikai Li, Chenjie Cao, Lei Sun, Yichen Qian,
  Weihua Chen, and Fan Wang.
\newblock Realismotion: Decomposed human motion control and video generation in
  the world space.
\newblock \emph{arXiv preprint arXiv:2508.08588}, 2025.

\bibitem[Liu et~al.(2021{\natexlab{a}})Liu, Schiele, and Sun]{liu2021adaptive}
Yaoyao Liu, Bernt Schiele, and Qianru Sun.
\newblock Adaptive aggregation networks for class-incremental learning.
\newblock In \emph{IEEE/CVF Conference on Computer Vision and Pattern
  Recognition (CVPR)}, pp.\  2544--2553, 2021{\natexlab{a}}.

\bibitem[Liu et~al.(2021{\natexlab{b}})Liu, Schiele, and Sun]{liu2021rmm}
Yaoyao Liu, Bernt Schiele, and Qianru Sun.
\newblock {RMM}: Reinforced memory management for class-incremental learning.
\newblock \emph{Advances in neural information processing systems (NeurIPS)},
  34:\penalty0 3478--3490, 2021{\natexlab{b}}.

\bibitem[Liu et~al.(2023{\natexlab{a}})Liu, Li, Schiele, and
  Sun]{liu2023online}
Yaoyao Liu, Yingying Li, Bernt Schiele, and Qianru Sun.
\newblock Online hyperparameter optimization for class-incremental learning.
\newblock In \emph{AAAI Conference on Artificial Intelligence (AAAI)},
  volume~37, pp.\  8906--8913, 2023{\natexlab{a}}.

\bibitem[Liu et~al.(2023{\natexlab{b}})Liu, Schiele, Vedaldi, and
  Rupprecht]{liu2023continual}
Yaoyao Liu, Bernt Schiele, Andrea Vedaldi, and Christian Rupprecht.
\newblock Continual detection transformer for incremental object detection.
\newblock In \emph{IEEE/CVF Conference on Computer Vision and Pattern
  Recognition (CVPR)}, pp.\  23799--23808, 2023{\natexlab{b}}.

\bibitem[Liu et~al.(2024)Liu, Li, Schiele, and Sun]{liu2024wakening}
Yaoyao Liu, Yingying Li, Bernt Schiele, and Qianru Sun.
\newblock Wakening past concepts without past data: Class-incremental learning
  from online placebos.
\newblock In \emph{IEEE/CVF Winter Conference on Applications of Computer
  Vision (WACV)}, pp.\  2215--2224, 2024.

\bibitem[Luo et~al.(2025)Luo, Bai, Shi, Xia, Wang, Wan, Zhang, Gai, and
  Xue]{luo2025camclonemaster}
Yawen Luo, Jianhong Bai, Xiaoyu Shi, Menghan Xia, Xintao Wang, Pengfei Wan,
  Di~Zhang, Kun Gai, and Tianfan Xue.
\newblock Camclonemaster: Enabling reference-based camera control for video
  generation.
\newblock In \emph{ACM SIGGRAPH Asia 2025 Conference Papers}, 2025.

\bibitem[Ma et~al.(2024{\natexlab{a}})Ma, Lewis, and Kleijn]{ma2024trailblazer}
Wan-Duo~Kurt Ma, John~P Lewis, and W~Bastiaan Kleijn.
\newblock Trailblazer: Trajectory control for diffusion-based video generation.
\newblock In \emph{SIGGRAPH Asia 2024 Conference Papers}, pp.\  1--11,
  2024{\natexlab{a}}.

\bibitem[Ma et~al.(2024{\natexlab{b}})Ma, Liu, Wang, Wang, Yuan, Zhang, Xiao,
  Zhang, Lu, Duan, Qi, Kortylewski, Liu, and Yuille]{ma2023generating}
Wufei Ma, Qihao Liu, Jiahao Wang, Angtian Wang, Xiaoding Yuan, Yi~Zhang, Zihao
  Xiao, Guofeng Zhang, Beijia Lu, Ruxiao Duan, Yongrui Qi, Adam Kortylewski,
  Yaoyao Liu, and Alan~L. Yuille.
\newblock Generating images with {3D} annotations using diffusion models.
\newblock In \emph{International Conference on Learning Representations
  (ICLR)}, 2024{\natexlab{b}}.

\bibitem[Mao et~al.(2025)Mao, Lin, Li, Li, Peng, He, Pang, Chi, Qiao, and
  Zhang]{mao2025yume}
Xiaofeng Mao, Shaoheng Lin, Zhen Li, Chuanhao Li, Wenshuo Peng, Tong He,
  Jiangmiao Pang, Mingmin Chi, Yu~Qiao, and Kaipeng Zhang.
\newblock Yume: An interactive world generation model.
\newblock \emph{arXiv preprint arXiv:2507.17744}, 2025.

\bibitem[Niu et~al.(2024)Niu, Cun, Wang, Zhang, Shan, and Zheng]{niu2025mofa}
Muyao Niu, Xiaodong Cun, Xintao Wang, Yong Zhang, Ying Shan, and Yinqiang
  Zheng.
\newblock Mofa-video: Controllable image animation via generative motion field
  adaptions in frozen image-to-video diffusion model.
\newblock In \emph{European Conference on Computer Vision (ECCV)}, pp.\
  111--128. Springer, 2024.

\bibitem[Parker-Holder et~al.(2024)Parker-Holder, Ball, Bruce, Dasagi,
  Holsheimer, Kaplanis, Moufarek, Scully, Shar, Shi, Spencer, Yung, Dennis,
  Kenjeyev, Long, Mnih, Chan, Gazeau, Li, Pardo, Wang, Zhang, Besse, Harley,
  Mitenkova, Wang, Clune, Hassabis, Hadsell, Bolton, Singh, and
  Rockt{\"a}schel]{parkerholder2024genie2}
Jack Parker-Holder, Philip Ball, Jake Bruce, Vibhavari Dasagi, Kristian
  Holsheimer, Christos Kaplanis, Alexandre Moufarek, Guy Scully, Jeremy Shar,
  Jimmy Shi, Stephen Spencer, Jessica Yung, Michael Dennis, Sultan Kenjeyev,
  Shangbang Long, Vlad Mnih, Harris Chan, Maxime Gazeau, Bonnie Li, Fabio
  Pardo, Luyu Wang, Lei Zhang, Frederic Besse, Tim Harley, Anna Mitenkova, Jane
  Wang, Jeff Clune, Demis Hassabis, Raia Hadsell, Adrian Bolton, Satinder
  Singh, and Tim Rockt{\"a}schel.
\newblock Genie 2: A large-scale foundation world model.
\newblock \emph{Google DeepMind technical report}, 2024.
\newblock URL
  \url{https://deepmind.google/discover/blog/genie-2-a-large-scale-foundation-world-model/}.

\bibitem[Parker-Holder et~al.(2025)Parker-Holder, Fruchter,
  et~al.]{parker2025genie}
Jack Parker-Holder, Shlomi Fruchter, et~al.
\newblock Genie 3: A new frontier for world models.
\newblock \emph{Google DeepMind Blog}, 4, 2025.

\bibitem[Piccinelli et~al.(2025)Piccinelli, Sakaridis, Yang, Segu, Li,
  Abbeloos, and Van~Gool]{unidepthv22025}
Luigi Piccinelli, Christos Sakaridis, Yung-Hsu Yang, Mattia Segu, Siyuan Li,
  Wim Abbeloos, and Luc Van~Gool.
\newblock {UniDepthV2}: Universal monocular metric depth estimation made
  simpler.
\newblock \emph{IEEE Transactions on Pattern Analysis and Machine
  Intelligence}, 2025.

\bibitem[Po et~al.(2025)Po, Nitzan, Zhang, Chen, Dao, Shechtman, Wetzstein, and
  Huang]{po2025long}
Ryan Po, Yotam Nitzan, Richard Zhang, Berlin Chen, Tri Dao, Eli Shechtman,
  Gordon Wetzstein, and Xun Huang.
\newblock Long-context state-space video world models.
\newblock \emph{arXiv preprint arXiv:2505.20171}, 2025.

\bibitem[Popov et~al.(2025)Popov, Raj, Krainin, Li, Freeman, and
  Rubinstein]{popov2025camctrl3d}
Stefan Popov, Amit Raj, Michael Krainin, Yuanzhen Li, William~T Freeman, and
  Michael Rubinstein.
\newblock Camctrl3d: Single-image scene exploration with precise 3d camera
  control.
\newblock \emph{arXiv preprint arXiv:2501.06006}, 2025.

\bibitem[Qiu et~al.(2023)Qiu, Xia, Zhang, He, Wang, Shan, and Liu]{freenoise}
Haonan Qiu, Menghan Xia, Yong Zhang, Yingqing He, Xintao Wang, Ying Shan, and
  Ziwei Liu.
\newblock {FreeNoise}: Tuning-free longer video diffusion via noise
  rescheduling.
\newblock \emph{arXiv preprint arXiv:2310.15169}, 2023.

\bibitem[Qiu et~al.(2024)Qiu, Chen, Wang, He, Xia, and Liu]{qiu2024freetraj}
Haonan Qiu, Zhaoxi Chen, Zhouxia Wang, Yingqing He, Menghan Xia, and Ziwei Liu.
\newblock Freetraj: Tuning-free trajectory control in video diffusion models.
\newblock \emph{arXiv preprint arXiv:2406.16863}, 2024.

\bibitem[Radford et~al.(2021)Radford, Kim, Hallacy, Ramesh, Goh, Agarwal,
  Sastry, Askell, Mishkin, Clark, Krueger, and Sutskever]{radford2021clip}
Alec Radford, Jong~Wook Kim, Chris Hallacy, Aditya Ramesh, Gabriel Goh,
  Sandhini Agarwal, Girish Sastry, Amanda Askell, Pamela Mishkin, Jack Clark,
  Gretchen Krueger, and Ilya Sutskever.
\newblock Learning transferable visual models from natural language
  supervision.
\newblock In \emph{International Conference on Machine Learning (ICML)}, volume
  139, pp.\  8748--8763, 2021.

\bibitem[Ravi et~al.(2025)Ravi, Gabeur, Hu, Hu, Ryali, Ma, Khedr, R{\"a}dle,
  Rolland, Gustafson, Mintun, Pan, Alwala, Carion, Wu, Girshick, Doll{\'a}r,
  and Feichtenhofer]{ravi2024sam2}
Nikhila Ravi, Valentin Gabeur, Yuan-Ting Hu, Ronghang Hu, Chaitanya Ryali,
  Tengyu Ma, Haitham Khedr, Roman R{\"a}dle, Chloe Rolland, Laura Gustafson,
  Eric Mintun, Junting Pan, Kalyan~Vasudev Alwala, Nicolas Carion, Chao-Yuan
  Wu, Ross Girshick, Piotr Doll{\'a}r, and Christoph Feichtenhofer.
\newblock {SAM 2}: Segment anything in images and videos.
\newblock In \emph{International Conference on Learning Representations
  (ICLR)}, 2025.

\bibitem[Ren et~al.(2025)Ren, Shen, Huang, Ling, Lu, Nimier-David, M{\"u}ller,
  Keller, Fidler, and Gao]{ren2025gen3c}
Xuanchi Ren, Tianchang Shen, Jiahui Huang, Huan Ling, Yifan Lu, Merlin
  Nimier-David, Thomas M{\"u}ller, Alexander Keller, Sanja Fidler, and Jun Gao.
\newblock Gen3c: 3d-informed world-consistent video generation with precise
  camera control.
\newblock In \emph{IEEE/CVF Conference on Computer Vision and Pattern
  Recognition (CVPR)}, pp.\  6121--6132, 2025.

\bibitem[Shi et~al.(2024)Shi, Huang, Wang, Bian, Li, Zhang, Zhang, Cheung, See,
  Qin, et~al.]{shi2024motion}
Xiaoyu Shi, Zhaoyang Huang, Fu-Yun Wang, Weikang Bian, Dasong Li, Yi~Zhang,
  Manyuan Zhang, Ka~Chun Cheung, Simon See, Hongwei Qin, et~al.
\newblock Motion-i2v: Consistent and controllable image-to-video generation
  with explicit motion modeling.
\newblock In \emph{ACM SIGGRAPH 2024 Conference Papers}, pp.\  1--11, 2024.

\bibitem[Shuai et~al.(2025)Shuai, Ding, Qin, Luo, Ma, and Tao]{shuai2025free}
Xincheng Shuai, Henghui Ding, Zhenyuan Qin, Hao Luo, Xingjun Ma, and Dacheng
  Tao.
\newblock Free-form motion control: Controlling the {6D} poses of camera and
  objects in video generation.
\newblock In \emph{IEEE/CVF International Conference on Computer Vision
  (ICCV)}, 2025.

\bibitem[Sitzmann et~al.(2021)Sitzmann, Rezchikov, Freeman, Tenenbaum, and
  Durand]{sitzmann2021light}
Vincent Sitzmann, Semon Rezchikov, Bill Freeman, Josh Tenenbaum, and Fredo
  Durand.
\newblock Light field networks: Neural scene representations with
  single-evaluation rendering.
\newblock \emph{Advances in Neural Information Processing Systems (NeurIPS)},
  2021.

\bibitem[Song et~al.(2026)Song, Yang, Zhao, Li, and Zhang]{song2026taming}
Chenxi Song, Yanming Yang, Tong Zhao, Ruibo Li, and Chi Zhang.
\newblock Taming video models for 3d and 4d generation via zero-shot camera
  control.
\newblock In \emph{IEEE/CVF Conference on Computer Vision and Pattern
  Recognition (CVPR)}, pp.\  40352--40363, 2026.

\bibitem[Sun et~al.(2024)Sun, Chen, Liu, Chen, Duan, Zhang, and
  Wang]{sun2024dimensionx}
Wenqiang Sun, Shuo Chen, Fangfu Liu, Zilong Chen, Yueqi Duan, Jun Zhang, and
  Yikai Wang.
\newblock Dimensionx: Create any 3d and 4d scenes from a single image with
  controllable video diffusion.
\newblock \emph{arXiv preprint arXiv:2411.04928}, 2024.

\bibitem[Sun et~al.(2025)Sun, Zhang, Wang, Wu, Wang, Wang, Wang, Zhang, Wang,
  and Guo]{worldplay}
Wenqiang Sun, Haiyu Zhang, Haoyuan Wang, Junta Wu, Zehan Wang, Zhenwei Wang,
  Yunhong Wang, Jun Zhang, Tengfei Wang, and Chunchao Guo.
\newblock Worldplay: Towards long-term geometric consistency for real-time
  interactive world modeling.
\newblock \emph{arXiv preprint arXiv:2512.14614}, 2025.

\bibitem[Tang et~al.(2026)Tang, Cao, Zhang, Luo, Liu, and
  Nie{\ss}ner]{tang2026motion2vecsets}
Jiapeng Tang, Wei Cao, Biao Zhang, Chang Luo, Yaoyao Liu, and Matthias
  Nie{\ss}ner.
\newblock {Motion2VecSets}: Non-rigid shape reconstruction and tracking with
  {4D} latent set diffusion.
\newblock \emph{IEEE Transactions on Pattern Analysis and Machine
  Intelligence}, 2026.

\bibitem[Team et~al.(2025)Team, Wang, Liu, Wu, Gu, Wang, Zuo, Huang, Li, Zhang,
  et~al.]{team2025hunyuanworld}
HunyuanWorld Team, Zhenwei Wang, Yuhao Liu, Junta Wu, Zixiao Gu, Haoyuan Wang,
  Xuhui Zuo, Tianyu Huang, Wenhuan Li, Sheng Zhang, et~al.
\newblock Hunyuanworld 1.0: Generating immersive, explorable, and interactive
  3d worlds from words or pixels.
\newblock \emph{arXiv preprint arXiv:2507.21809}, 2025.

\bibitem[Unterthiner et~al.(2018)Unterthiner, van Steenkiste, Kurach, Marinier,
  Michalski, and Gelly]{unterthiner2018fvd}
Thomas Unterthiner, Sjoerd van Steenkiste, Karol Kurach, Rapha{\"e}l Marinier,
  Marcin Michalski, and Sylvain Gelly.
\newblock Towards accurate generative models of video: A new metric \&
  challenges.
\newblock \emph{arXiv preprint arXiv:1812.01717}, 2018.

\bibitem[Van~Hoorick et~al.(2024)Van~Hoorick, Wu, Ozguroglu, Sargent, Liu,
  Tokmakov, Dave, Zheng, and Vondrick]{van2024generative}
Basile Van~Hoorick, Rundi Wu, Ege Ozguroglu, Kyle Sargent, Ruoshi Liu, Pavel
  Tokmakov, Achal Dave, Changxi Zheng, and Carl Vondrick.
\newblock Generative camera dolly: Extreme monocular dynamic novel view
  synthesis.
\newblock In \emph{European Conference on Computer Vision (ECCV)}, pp.\
  313--331. Springer, 2024.

\bibitem[Voleti et~al.(2024)Voleti, Yao, Boss, Letts, Pankratz, Tochilkin,
  Laforte, Rombach, and Jampani]{voleti2024sv3d}
Vikram Voleti, Chun-Han Yao, Mark Boss, Adam Letts, David Pankratz, Dmitry
  Tochilkin, Christian Laforte, Robin Rombach, and Varun Jampani.
\newblock Sv3d: Novel multi-view synthesis and 3d generation from a single
  image using latent video diffusion.
\newblock In \emph{European Conference on Computer Vision (ECCV)}, pp.\
  439--457. Springer, 2024.

\bibitem[Wang et~al.(2025{\natexlab{a}})Wang, Ai, Wen, Mao, Xie, Chen, Yu,
  Zhao, Yang, Zeng, et~al.]{wan2025}
Ang Wang, Baole Ai, Bin Wen, Chaojie Mao, Chen-Wei Xie, Di~Chen, Feiwu Yu,
  Haiming Zhao, Jianxiao Yang, Jianyuan Zeng, et~al.
\newblock {Wan}: Open and advanced large-scale video generative models.
\newblock \emph{arXiv preprint arXiv:2503.20314}, 2025{\natexlab{a}}.

\bibitem[Wang et~al.(2025{\natexlab{b}})Wang, Huang, Fang, Yang, and
  Ma]{wang2025ati}
Angtian Wang, Haibin Huang, Jacob~Zhiyuan Fang, Yiding Yang, and Chongyang Ma.
\newblock Ati: Any trajectory instruction for controllable video generation.
\newblock \emph{arXiv preprint arXiv:2505.22944}, 2025{\natexlab{b}}.

\bibitem[Wang et~al.(2025{\natexlab{c}})Wang, Ouyang, Wang, Wang, Cheng, Chen,
  Shen, and Wang]{wang2025levitor}
Hanlin Wang, Hao Ouyang, Qiuyu Wang, Wen Wang, Ka~Leong Cheng, Qifeng Chen,
  Yujun Shen, and Limin Wang.
\newblock {LeviTor}: {3D} trajectory oriented image-to-video synthesis.
\newblock In \emph{IEEE/CVF Conference on Computer Vision and Pattern
  Recognition (CVPR)}, 2025{\natexlab{c}}.

\bibitem[Wang et~al.(2024{\natexlab{a}})Wang, Karaev, Rupprecht, and
  Novotny]{wang2024vggsfm}
Jianyuan Wang, Nikita Karaev, Christian Rupprecht, and David Novotny.
\newblock {Vggsfm}: Visual geometry grounded deep structure from motion.
\newblock In \emph{2024 IEEE/CVF Conference on Computer Vision and Pattern
  Recognition (CVPR)}, pp.\  21686--21697. IEEE, 2024{\natexlab{a}}.

\bibitem[Wang et~al.(2024{\natexlab{b}})Wang, Zhang, Zou, Zeng, Wei, Yuan, and
  Li]{wang2024boximator}
Jiawei Wang, Yuchen Zhang, Jiaxin Zou, Yan Zeng, Guoqiang Wei, Liping Yuan, and
  Hang Li.
\newblock {Boximator}: Generating rich and controllable motions for video
  synthesis.
\newblock \emph{arXiv preprint arXiv:2402.01566}, 2024{\natexlab{b}}.

\bibitem[Wang et~al.(2025{\natexlab{d}})Wang, Luo, Shi, Jia, Lu, Xue, Wang,
  Wan, Zhang, and Gai]{wang2025cinemaster}
Qinghe Wang, Yawen Luo, Xiaoyu Shi, Xu~Jia, Huchuan Lu, Tianfan Xue, Xintao
  Wang, Pengfei Wan, Di~Zhang, and Kun Gai.
\newblock {CineMaster}: A {3D}-aware and controllable framework for cinematic
  text-to-video generation.
\newblock In \emph{ACM SIGGRAPH 2025 Conference Papers}, pp.\  60:1--60:10,
  2025{\natexlab{d}}.

\bibitem[Wang et~al.(2025{\natexlab{e}})Wang, Xu, Dong, Deng, Xiang, Lv, Sun,
  Tong, and Yang]{moge22025}
Ruicheng Wang, Sicheng Xu, Yue Dong, Yu~Deng, Jianfeng Xiang, Zelong Lv,
  Guangzhong Sun, Xin Tong, and Jiaolong Yang.
\newblock {MoGe-2}: Accurate monocular geometry with metric scale and sharp
  details.
\newblock \emph{arXiv preprint arXiv:2507.02546}, 2025{\natexlab{e}}.

\bibitem[Wang et~al.(2023)Wang, Yuan, Zhang, Chen, Wang, Zhang, Shen, Zhao, and
  Zhou]{wang2023videocomposer}
Xiang Wang, Hangjie Yuan, Shiwei Zhang, Dayou Chen, Jiuniu Wang, Yingya Zhang,
  Yujun Shen, Deli Zhao, and Jingren Zhou.
\newblock Videocomposer: Compositional video synthesis with motion
  controllability.
\newblock \emph{Advances in Neural Information Processing Systems (NeurIPS)},
  36:\penalty0 7594--7611, 2023.

\bibitem[Wang et~al.(2024{\natexlab{c}})Wang, Lan, Zhou, and
  Loy]{wang2024objctrl}
Zhouxia Wang, Yushi Lan, Shangchen Zhou, and Chen~Change Loy.
\newblock {ObjCtrl-2.5D}: Training-free object control with camera poses.
\newblock \emph{arXiv preprint arXiv:2412.07721}, 2024{\natexlab{c}}.

\bibitem[Wang et~al.(2024{\natexlab{d}})Wang, Yuan, Wang, Li, Chen, Xia, Luo,
  and Shan]{motionctrl2024}
Zhouxia Wang, Ziyang Yuan, Xintao Wang, Yaowei Li, Tianshui Chen, Menghan Xia,
  Ping Luo, and Ying Shan.
\newblock {MotionCtrl}: A unified and flexible motion controller for video
  generation.
\newblock In \emph{ACM SIGGRAPH 2024 Conference Papers}, 2024{\natexlab{d}}.

\bibitem[Wang et~al.(2025{\natexlab{f}})Wang, Cho, Li, Lin, Yoon, Zhang, and
  Bansal]{wang2025epic}
Zun Wang, Jaemin Cho, Jialu Li, Han Lin, Jaehong Yoon, Yue Zhang, and Mohit
  Bansal.
\newblock Epic: Efficient video camera control learning with precise
  anchor-video guidance.
\newblock \emph{arXiv preprint arXiv:2505.21876}, 2025{\natexlab{f}}.

\bibitem[Wu et~al.(2025{\natexlab{a}})Wu, Yang, Po, Xu, Liu, Lin, and
  Wetzstein]{wu2025video}
Tong Wu, Shuai Yang, Ryan Po, Yinghao Xu, Ziwei Liu, Dahua Lin, and Gordon
  Wetzstein.
\newblock Video world models with long-term spatial memory.
\newblock \emph{arXiv preprint arXiv:2506.05284}, 2025{\natexlab{a}}.

\bibitem[Wu et~al.(2024)Wu, Li, Gu, Zhao, He, Zhang, Shou, Li, Gao, and
  Zhang]{wu2025draganything}
Weijia Wu, Zhuang Li, Yuchao Gu, Rui Zhao, Yefei He, David~Junhao Zhang,
  Mike~Zheng Shou, Yan Li, Tingting Gao, and Di~Zhang.
\newblock Draganything: Motion control for anything using entity
  representation.
\newblock In \emph{European Conference on Computer Vision (ECCV)}, pp.\
  331--348. Springer, 2024.

\bibitem[Wu et~al.(2022)Wu, Lee, and Hoiem]{wu2022sparse}
Yuqun Wu, Jae~Yong Lee, and Derek Hoiem.
\newblock {Sparse SPN}: Depth completion from sparse keypoints.
\newblock \emph{arXiv preprint arXiv:2212.00987}, 2022.

\bibitem[Wu et~al.(2025{\natexlab{b}})Wu, Lee, Zou, Wang, and
  Hoiem]{wu2025monopatchnerf}
Yuqun Wu, Jae~Yong Lee, Chuhang Zou, Shenlong Wang, and Derek Hoiem.
\newblock {MonoPatchNeRF}: Improving neural radiance fields with patch-based
  monocular guidance.
\newblock In \emph{International Conference on 3D Vision (3DV)},
  2025{\natexlab{b}}.

\bibitem[Wu et~al.(2026)Wu, Lin, Che, Tiwari, Zou, Wang, and
  Hoiem]{wu2026scenediff}
Yuqun Wu, Chih-hao Lin, Henry Che, Aditi Tiwari, Chuhang Zou, Shenlong Wang,
  and Derek Hoiem.
\newblock {SceneDiff}: A benchmark and method for multiview object change
  detection.
\newblock In \emph{European Conference on Computer Vision (ECCV)}, 2026.

\bibitem[Xiao et~al.(2025)Xiao, Lan, Zhou, Ouyang, Yang, Zeng, and
  Pan]{worldmem}
Zeqi Xiao, Yushi Lan, Yifan Zhou, Wenqi Ouyang, Shuai Yang, Yanhong Zeng, and
  Xingang Pan.
\newblock {WORLDMEM}: Long-term consistent world simulation with memory.
\newblock In \emph{Advances in Neural Information Processing Systems
  (NeurIPS)}, 2025.

\bibitem[Xie et~al.(2025)Xie, Yao, Voleti, Jiang, and Jampani]{xie2025sv4d}
Yiming Xie, Chun-Han Yao, Vikram Voleti, Huaizu Jiang, and Varun Jampani.
\newblock Sv4d: Dynamic 3d content generation with multi-frame and multi-view
  consistency.
\newblock In \emph{International Conference on Learning Representations
  (ICLR)}, pp.\  33421--33441, 2025.

\bibitem[Xing et~al.(2025)Xing, Mai, Ham, Huang, Mahapatra, Fu, Wong, and
  Liu]{xing2025motioncanvas}
Jinbo Xing, Long Mai, Cusuh Ham, Jiahui Huang, Aniruddha Mahapatra, Chi-Wing
  Fu, Tien-Tsin Wong, and Feng Liu.
\newblock Motioncanvas: Cinematic shot design with controllable image-to-video
  generation.
\newblock In \emph{ACM SIGGRAPH 2025 Conference Papers}, pp.\  1--11, 2025.

\bibitem[Xu et~al.(2024)Xu, Nie, Liu, Liu, Kautz, Wang, and
  Vahdat]{xu2024camco}
Dejia Xu, Weili Nie, Chao Liu, Sifei Liu, Jan Kautz, Zhangyang Wang, and Arash
  Vahdat.
\newblock {CamCo}: Camera-controllable {3D}-consistent image-to-video
  generation.
\newblock \emph{arXiv preprint arXiv:2406.02509}, 2024.

\bibitem[Yang et~al.(2024)Yang, Hou, Huang, Ma, Wan, Zhang, Chen, and
  Liao]{yang2024direct}
Shiyuan Yang, Liang Hou, Haibin Huang, Chongyang Ma, Pengfei Wan, Di~Zhang,
  Xiaodong Chen, and Jing Liao.
\newblock Direct-a-video: Customized video generation with user-directed camera
  movement and object motion.
\newblock In \emph{ACM SIGGRAPH 2024 Conference Papers}, pp.\  1--12, 2024.

\bibitem[Yang et~al.(2025)Yang, Ge, Li, Chen, Li, An, Kang, Xue, Xu, Yin,
  et~al.]{yang2025matrix}
Zhongqi Yang, Wenhang Ge, Yuqi Li, Jiaqi Chen, Haoyuan Li, Mengyin An, Fei
  Kang, Hua Xue, Baixin Xu, Yuyang Yin, et~al.
\newblock Matrix-3d: Omnidirectional explorable 3d world generation.
\newblock \emph{arXiv preprint arXiv:2508.08086}, 2025.

\bibitem[Yao et~al.(2025)Yao, Xie, Voleti, Jiang, and Jampani]{yao2025sv4d}
Chun-Han Yao, Yiming Xie, Vikram Voleti, Huaizu Jiang, and Varun Jampani.
\newblock Sv4d 2.0: Enhancing spatio-temporal consistency in multi-view video
  diffusion for high-quality 4d generation.
\newblock In \emph{2025 IEEE/CVF International Conference on Computer Vision
  (ICCV)}, pp.\  13248--13258. IEEE, 2025.

\bibitem[Yin et~al.(2023)Yin, Wu, Liang, Shi, Li, Ming, and
  Duan]{yin2023dragnuwa}
Shengming Yin, Chenfei Wu, Jian Liang, Jie Shi, Houqiang Li, Gong Ming, and Nan
  Duan.
\newblock {DragNUWA}: Fine-grained control in video generation by integrating
  text, image, and trajectory.
\newblock \emph{arXiv preprint arXiv:2308.08089}, 2023.

\bibitem[Yu et~al.(2025{\natexlab{a}})Yu, Bai, Qin, Liu, Wang, Wan, Zhang, and
  Liu]{contextasmemory}
Jiwen Yu, Jianhong Bai, Yiran Qin, Quande Liu, Xintao Wang, Pengfei Wan,
  Di~Zhang, and Xihui Liu.
\newblock Context as memory: Scene-consistent interactive long video generation
  with memory retrieval.
\newblock \emph{arXiv preprint arXiv:2506.03141}, 2025{\natexlab{a}}.

\bibitem[Yu et~al.(2025{\natexlab{b}})Yu, Qin, Wang, Wan, Zhang, and
  Liu]{yu2025gamefactory}
Jiwen Yu, Yiran Qin, Xintao Wang, Pengfei Wan, Di~Zhang, and Xihui Liu.
\newblock Gamefactory: Creating new games with generative interactive videos.
\newblock \emph{arXiv preprint arXiv:2501.08325}, 2025{\natexlab{b}}.

\bibitem[Yu et~al.(2025{\natexlab{c}})Yu, Hu, Xing, and
  Shan]{yu2025trajectorycrafter}
Mark Yu, Wenbo Hu, Jinbo Xing, and Ying Shan.
\newblock Trajectorycrafter: Redirecting camera trajectory for monocular videos
  via diffusion models.
\newblock In \emph{IEEE/CVF International Conference on Computer Vision
  (ICCV)}, pp.\  100--111, 2025{\natexlab{c}}.

\bibitem[Yu et~al.(2024)Yu, Xing, Yuan, Hu, Li, Huang, Gao, Wong, Shan, and
  Tian]{yu2024viewcrafter}
Wangbo Yu, Jinbo Xing, Li~Yuan, Wenbo Hu, Xiaoyu Li, Zhipeng Huang, Xiangjun
  Gao, Tien-Tsin Wong, Ying Shan, and Yonghong Tian.
\newblock {ViewCrafter}: Taming video diffusion models for high-fidelity novel
  view synthesis.
\newblock \emph{arXiv preprint arXiv:2409.02048}, 2024.

\bibitem[Zhang et~al.(2025{\natexlab{a}})Zhang, Ke, Harley, and
  Fragkiadaki]{zhang2026tapip3d}
Bowei Zhang, Lei Ke, Adam Harley, and Katerina Fragkiadaki.
\newblock {Tapip3d}: Tracking any point in persistent {3D} geometry.
\newblock \emph{Advances in Neural Information Processing Systems (NeurIPS)},
  38:\penalty0 135284--135303, 2025{\natexlab{a}}.

\bibitem[Zhang et~al.(2024)Zhang, Paiss, Zada, Karnad, Jacobs, Pritch, Mosseri,
  Shou, Wadhwa, and Ruiz]{zhang2024recapture}
David~Junhao Zhang, Roni Paiss, Shiran Zada, Nikhil Karnad, David~E Jacobs,
  Yael Pritch, Inbar Mosseri, Mike~Zheng Shou, Neal Wadhwa, and Nataniel Ruiz.
\newblock Recapture: Generative video camera controls for user-provided videos
  using masked video fine-tuning.
\newblock \emph{arXiv preprint arXiv:2411.05003}, 2024.

\bibitem[Zhang et~al.(2026)Zhang, Chen, Xiang, Huang, Wang, and
  Jiang]{zhang2026symphomotion}
Guiyu Zhang, Yabo Chen, Xunzhi Xiang, Junchao Huang, Zhongyu Wang, and
  Li~Jiang.
\newblock {Symphomotion}: Joint control of camera motion and object dynamics
  for coherent video generation.
\newblock In \emph{IEEE/CVF Conference on Computer Vision and Pattern
  Recognition (CVPR)}, pp.\  11127--11137, 2026.

\bibitem[Zhang et~al.(2025{\natexlab{b}})Zhang, Yao, Donn{\'e}, Ahuja, and
  Jampani]{zhang2026stable}
Hao Zhang, Chun-Han Yao, Simon Donn{\'e}, Narendra Ahuja, and Varun Jampani.
\newblock Stable part diffusion 4d: Multi-view rgb and kinematic parts video
  generation.
\newblock \emph{Advances in Neural Information Processing Systems (NeurIPS)},
  38:\penalty0 114172--114195, 2025{\natexlab{b}}.

\bibitem[Zhang et~al.(2023)Zhang, Rao, and Agrawala]{controlnet}
Lvmin Zhang, Anyi Rao, and Maneesh Agrawala.
\newblock Adding conditional control to text-to-image diffusion models.
\newblock In \emph{IEEE/CVF International Conference on Computer Vision
  (ICCV)}, 2023.

\bibitem[Zhang et~al.(2025{\natexlab{c}})Zhang, Liao, Li, Dai, Qiu, Zhu, Qin,
  and Wang]{zhang2025tora}
Zhenghao Zhang, Junchao Liao, Menghao Li, Zuozhuo Dai, Bingxue Qiu, Siyu Zhu,
  Long Qin, and Weizhi Wang.
\newblock Tora: Trajectory-oriented diffusion transformer for video generation.
\newblock In \emph{IEEE/CVF Conference on Computer Vision and Pattern
  Recognition (CVPR)}, 2025{\natexlab{c}}.

\bibitem[Zhang et~al.(2025{\natexlab{d}})Zhang, Chen, and Liao]{zhang2025i2v3d}
Zhiyuan Zhang, Dongdong Chen, and Jing Liao.
\newblock I2v3d: Controllable image-to-video generation with 3d guidance.
\newblock \emph{arXiv preprint arXiv:2503.09733}, 2025{\natexlab{d}}.

\bibitem[Zheng et~al.(2024)Zheng, Li, Jiang, Lu, Wu, and Li]{zheng2024cami2v}
Guangcong Zheng, Teng Li, Rui Jiang, Yehao Lu, Tao Wu, and Xi~Li.
\newblock Cami2v: Camera-controlled image-to-video diffusion model.
\newblock \emph{arXiv preprint arXiv:2410.15957}, 2024.

\bibitem[Zheng et~al.(2025)Zheng, Peng, Zhou, Zhu, Xu, Huang, and
  Fu]{zheng2025vidcraft3}
Sixiao Zheng, Zimian Peng, Yanpeng Zhou, Yi~Zhu, Hang Xu, Xiangru Huang, and
  Yanwei Fu.
\newblock Vidcraft3: Camera, object, and lighting control for image-to-video
  generation.
\newblock \emph{arXiv preprint arXiv:2502.07531}, 2025.

\bibitem[Zheng et~al.(2026)Zheng, Yin, Hu, Li, Shan, and
  Fu]{zheng2026versecrafter}
Sixiao Zheng, Minghao Yin, Wenbo Hu, Xiaoyu Li, Ying Shan, and Yanwei Fu.
\newblock {VerseCrafter}: Dynamic realistic video world model with {4D}
  geometric control.
\newblock In \emph{IEEE/CVF Conference on Computer Vision and Pattern
  Recognition (CVPR)}, pp.\  40277--40290, 2026.

\bibitem[Zhou et~al.(2025{\natexlab{a}})Zhou, Wang, Nie, Liu, Yu, Yu, and
  Wang]{zhou2025trackgo}
Haitao Zhou, Chuang Wang, Rui Nie, Jinlin Liu, Dongdong Yu, Qian Yu, and
  Changhu Wang.
\newblock {TrackGo}: A flexible and efficient method for controllable video
  generation.
\newblock In \emph{AAAI Conference on Artificial Intelligence (AAAI)},
  volume~39, 2025{\natexlab{a}}.

\bibitem[Zhou et~al.(2025{\natexlab{b}})Zhou, Gao, Voleti, Vasishta, Yao, Boss,
  Torr, Rupprecht, and Jampani]{zhou2025stable}
Jensen Zhou, Hang Gao, Vikram Voleti, Aaryaman Vasishta, Chun-Han Yao, Mark
  Boss, Philip Torr, Christian Rupprecht, and Varun Jampani.
\newblock Stable virtual camera: Generative view synthesis with diffusion
  models.
\newblock In \emph{2025 IEEE/CVF International Conference on Computer Vision
  (ICCV)}, pp.\  12405--12414. IEEE, 2025{\natexlab{b}}.

\bibitem[Zhu et~al.(2025)Zhu, Wang, Zhou, Chang, Zhou, Li, Chen, Shen, Pang,
  and He]{zhu2025aether}
Haoyi Zhu, Yifan Wang, Jianjun Zhou, Wenzheng Chang, Yang Zhou, Zizun Li, Junyi
  Chen, Chunhua Shen, Jiangmiao Pang, and Tong He.
\newblock Aether: Geometric-aware unified world modeling.
\newblock In \emph{IEEE/CVF International Conference on Computer Vision
  (ICCV)}, pp.\  8535--8546, 2025.

\end{thebibliography}
\bibliographystyle{plainnat}

\clearpage
\appendix
\raggedbottom
\renewcommand{\topfraction}{0.9}
\renewcommand{\textfraction}{0.07}
\renewcommand{\floatpagefraction}{0.8}

\section{Implementation Details}
\label{sec:impl_details}

\subsection{Rigid 3D Scene Construction}
\label{sec:construction_details}
\label{sec:gauge_details}
\label{sec:track_details}
\label{sec:alignment_details}
\label{sec:motion_details}

\paragraph{Camera and depth.}
The scene coordinate frame is produced by MegaSaM~\citep{megasam2025} camera
tracking, bundle adjustment and consistent-video-depth refinement, with
MoGe-2~\citep{moge22025} as the monocular depth prior and
UniDepthV2~\citep{unidepthv22025} as the metric-depth branch. MegaSaM returns
one intrinsic matrix per clip and one world-to-camera matrix per frame.
Metric cues enter only through MegaSaM's per-frame alignment of the monocular
disparity, and no later stage rescales cameras, depth, or tracks; the result is
a shared coordinate system, not a metric reconstruction.

\paragraph{Object tracks.}
TAPIP3D~\citep{zhang2026tapip3d} is run with its released checkpoint (16-frame
window, stride 8) in a forward and a reversed pass so that tracks from later
anchors extend back to the first frame. Per object, eight anchor frames are
spread over the clip and 128 farthest-point queries are drawn inside the mask
at each anchor, giving 1{,}024 queries per object. An observation is valid if
the tracker's visibility score is at least 0.9, the point has positive
camera-axis depth, and it projects inside the object's mask dilated by 2\,px;
a track is retained if at least 20\% of its frames are valid.

\paragraph{Canonical mesh and alignment.}
Pixal3D~\citep{li2026pixal3d} receives the first frame cropped to a square of
side $1.1\times$ the longer side of the mask's bounding box, with the field of
view derived from the MegaSaM intrinsics, and returns a textured mesh (100k
faces). Rendering the mesh in its reconstruction camera pairs visible mesh
points with first-frame pixels; one correspondence is kept per pixel inside the
mask, the mask is eroded by 1\,px, and 20\% of the $8\times8$ pixel cells are
withheld from the fit. The similarity transform is a
closed-form Umeyama fit with a median${}+3\,\mathrm{MAD}$ inlier threshold. Residuals are normalized
by the object diameter $d$, the extent of the correspondence cloud. An object
passes the 3D gate if the residual on the withheld correspondences has median ${}\le0.15$, p90
${}\le0.30$ and p95 ${}\le0.50$, and the silhouette gate if the placed mesh
rasterized into the first-frame camera has IoU ${}\ge0.5$ with the mask. A
clip is complete only if every object passes both gates; otherwise it is
excluded from the training set.

\paragraph{Stable core.}
Per frame, Eq.~\eqref{eq:core_fit} is solved with the Tukey loss as $\rho$ and
with track weights that combine observation support, a body-center prior and
a spatial-balancing factor, so that a densely tracked region cannot dominate
the pose. Per frame, a seeded RANSAC (threshold
$0.035\,d$) followed by Tukey-reweighted iterations (Tukey constant
$0.05\,d$) updates the rotation by weighted Kabsch and the translation by a
robust geometric median. A track is stable if its weighted inlier fraction at
$0.05\,d$ is at least 0.70 over frames with a valid fit, and it is observed in
at least 30\% of them; the core is re-fit until the stable set repeats and
must contain at least 30 tracks. If the core is too small or degenerate,
tracks are clustered by the invariance of their pairwise distances over time
and the most rigid cluster seeds the core.

\subsection{Rendering and Encoding}
\label{sec:encoding_details}
\label{sec:visibility_details}

The background point cloud is the first-frame depth back-projected after
removing the union of all object masks dilated by 9\,px. The background is rasterized as a splatted point cloud and each
placed mesh with one face per pixel, both with the same intrinsics and
camera-axis depth $Z$. The front-most entity at each pixel defines the depth $Z_t$, and pixels
covered by no entity are invalid. The 1st and 99th
percentiles of $1/Z_t$ are computed per entity over all frames, and
$[\alpha,\beta]$ is their envelope over entities, so that the range of a small
near object is never clipped by the background. Because the range is fixed per
clip, a given depth maps to the same gray value in every frame. A valid pixel
is encoded as
\begin{equation}
    \mathbf C_t
    =
    \Big\lfloor
    255\,\operatorname{clip}\!\Big(
        \frac{1/Z_t-\alpha}{\beta-\alpha},\,0,\,1\Big)
    \Big\rceil,
    \label{eq:depth_encoding}
\end{equation}
and stored as 8-bit gray (near ${}=255$, far ${}=0$); invalid pixels receive
the constant 128.

\subsection{Motion Adapter}
\label{sec:vace_details}
\label{sec:training_details}
\label{sec:inference_details}

\paragraph{Context stream.}
The adapter follows the conditioning branch of
Wan2.1-VACE-14B~\citep{wan2025,vace2025} in architecture and is initialized
from its released weights. The VAE encoding of the control
video is paired with an all-ones mask that marks every pixel for synthesis,
and the encoding of the input image is prepended along time; the eight
context blocks share the architecture of the Wan-DiT blocks.

\paragraph{Training.}
Table~\ref{tab:training_variants} lists the hyper-parameters of the two
variants. Both train on the 20{,}774 clips of \mbox{RealCOD-Rigid} at
$81\times832\times480$ in bf16 with a learning rate that warms up linearly
and then stays constant. The timestep is sampled uniformly over the 1{,}000-step Wan
flow schedule (shift 5), and the timestep weight $\lambda_k$ of
Eq.~\eqref{eq:flow_loss} is the scheduler's default. Before training the full
adapter, its attention and feed-forward matrices are perturbed once with
uniform noise of magnitude $3\times10^{-4}$ times each tensor's standard
deviation, in the manner of NoisyTune; the LoRA variant starts from the
unperturbed released weights.

\begin{table}[H]
    \centering
    \caption{\textbf{Training hyper-parameters of the two variants.}}
    \label{tab:training_variants}
    \small
    \setlength{\tabcolsep}{3pt}
    \begin{tabular}{%
      >{\raggedright\arraybackslash}p{0.22\linewidth}%
      >{\raggedright\arraybackslash}p{0.33\linewidth}%
      >{\raggedright\arraybackslash}p{0.36\linewidth}}
        \toprule
        & \method{} (full) & \method-LoRA \\
        \midrule
        Trainable parameters & entire adapter, 3.049\,B & rank-16 LoRA on the context blocks' attention and feed-forward projections, ${\approx}15.3$\,M \\
        Initialization & official VACE + NoisyTune-style noise & official VACE \\
        GPUs / global batch & $3\times8$ / 24 & $3\times8$ / 24 \\
        Epochs & 3 & 3 \\
        Peak learning rate & $5\times10^{-5}$ & $5\times10^{-5}$ \\
        Warmup steps & 25 & 25 \\
        Weight decay & 0.01 & 0.01 \\
        Seed & 42 & 42 \\
        \bottomrule
    \end{tabular}
\end{table}

\paragraph{Inference.}
Both variants use 20 flow-matching sampling steps, classifier-free guidance
5.0 with the pipeline's default negative prompt, sigma shift 5.0, seed 42,
$832\times480$, 81 frames at 16\,fps, tiled VAE decoding and hint scale 1.0.
The same checkpoints and settings were used for every result in this paper.

\section{Dataset Details}
\label{sec:dataset_details}
\label{sec:data_contract}
\label{sec:audit_appendix}
\label{sec:construction_audit_details}

The annotation pipeline takes the 25{,}318 source clips of
RealCOD-25K~\citep{zhang2026symphomotion}, monocular clips of 81 frames at
$832\times480$ and 16\,fps, each with one text prompt and SAM3 instance masks
for one or two objects (29{,}338 objects in total). The clips are stock and
open-source web videos; we reuse the videos, text prompts and instance masks,
and derive every rigid 3D scene, trajectory, and control video of RealCOD-Rigid
with our pipeline. An
object is allowed to vanish in some frames so that genuine occlusion is
represented rather than suppressed. Clips that fail a stage, because of too few correspondences or missing
outputs, and clips that belong to the evaluation set of
Sec.~\ref{sec:experimental_setup} are excluded, leaving the 20{,}774 training
clips of RealCOD-Rigid.

\section{Evaluation Details}
\label{sec:eval_details}

\subsection{Baselines}
\label{sec:converter_details}

All four baselines run from their released checkpoints with their native
sampling settings, resolution, and clip length; MotionCtrl and
Perception-as-Control generate 16 frames, which are mapped to the evaluation
frames through each converter's frame table. Each converter reads the shared
camera, depth, and track estimates of Appendix~\ref{sec:construction_details}
and produces the baseline's native control at the capacity its
implementation accepts. \emph{MotionCtrl} receives the first frame through
SparseCtrl, the shared camera through its camera module and up to eight track
flows through its object module. \emph{Perception-as-Control} receives up to
16 tracked 3D points rendered as spheres plus its camera tensor, and does not
consume the text prompt. \emph{SymphoMotion} receives the camera, per-object
point sets built from the first-frame mask points and moved along the object
trajectory, the 2D boxes it projects from them, a reference point-cloud render and the prompt.
Since point-based controls natively take per-point trajectories, which are cumbersome to author one by one, moving the points rigidly along the
object trajectory gives SymphoMotion the same motion that \method{} receives.
\emph{VerseCrafter} receives background RGB and depth renders, a per-object
3D Gaussian fitted at the first frame and moved along the object trajectory,
and a merged mask.

\subsection{Metrics}
\label{sec:metric_details}
\label{sec:standard_metric_details}
\label{sec:camera_metric_details}
\label{sec:objfid_details}

\paragraph{Visual quality.}
FID~\citep{heusel2017ttur} uses InceptionV3 pool3 features on 16 uniformly
spaced frames per video, with the source clips of the same cases as the real
set. FVD~\citep{unterthiner2018fvd} is computed with the StyleGAN-V (I3D)
feature extractor over every native frame of each method. CLIP-SIM
is the mean frame--text cosine similarity of CLIP ViT-B/32~\citep{radford2021clip}
over 16 frames. VBench-I2V~\citep{huang2024vbench} is run on eight dimensions
(subject consistency, background consistency, motion smoothness, dynamic
degree, aesthetic quality, imaging quality, I2V subject, I2V background); its
camera-motion dimension requires one of seven fixed camera instructions and
does not apply to continuous trajectories.

\paragraph{Camera control.}
The camera trajectory of each generated video is estimated with the same
MegaSaM pipeline that produced the target trajectories and aligned to the
target by a $\mathrm{Sim}(3)$ transform, following the evaluation
protocol of VGGSfM~\citep{wang2024vggsfm}. A case is
successful if a trajectory was estimated and the alignment is valid (positive
scale, proper rotation). Rotation error is the mean geodesic distance between
aligned and target orientations in degrees; relative translation error is the
mean camera-center distance divided by the target trajectory's maximum
displacement from the first frame. Both errors are averaged over the cases in which a trajectory could be
estimated.

\paragraph{Identity-Gated IoU.}
A frame of the source clip counts as visible if the ground-truth mask
covers at least $10^{-4}$ of it, and the $N=16$ test frames of
Eq.~\ref{eq:igiou} are taken at a uniform spacing over the visible frames of
the first object of the case, mapped through each converter's frame table so
that all methods are scored on the same moments; when the object is visible in
fewer than 16 frames, $N$ is the number of visible frames.
Generated masks come from SAM3 prompted on the first generated frame with a
point and a box derived from the ground-truth first-frame mask and propagated
through the video. For each test frame, a
Qwen3-VL-2B-Instruct judge compares two RGB crops: the ground-truth frame and
the generated frame, cut with the same window, which is the bounding box of the
union of the two masks padded by 20\% of its longer side and at least 96 pixels
wide, and resized to a 448-pixel maximum edge. The judge never sees a mask
overlay, so the gate reads appearance rather than segmentation. A frame is valid only for a \texttt{PASS} verdict with
confidence at least 0.5; every other verdict, a low-confidence pass, and a
frame in which the segmenter finds no object are invalid, and a frame
without a generated mask is scored without calling the judge. The instruction
is:

\begin{quote}\small\ttfamily\raggedright
You are comparing the same target object at the same time in two videos.
Image 1 is the ground-truth video crop. Image 2 is the generated-video crop.
Both images use exactly the same spatial crop window. Judge only the main
target object shown in Image 1. Ignore background changes, small texture
differences, minor blur, and location errors.\\[2pt]
Choose exactly one verdict:\\
- PASS: Image 2 preserves the target object's identity and has no undeniable
major structural defect.\\
- IDENTITY\_FAIL: Image 2 shows a different semantic object or loses the
target object's defining identity.\\
- STRUCTURE\_FAIL: It is the same object, but it has an undeniable missing or
duplicated major part, melted/fused geometry, broken/floating major component,
or impossible attachment.\\
- BOTH\_FAIL: Both identity and major structure fail.\\
- UNCERTAIN: visibility or image quality is insufficient to decide.\\[2pt]
Do not penalize a part hidden by the same natural occlusion or frame boundary
in Image 1. Return one JSON object only, with exactly three keys: verdict (one
label above), confidence (a number from 0 to 1), and reason (one short
sentence). The verdict must agree with the reason.
\end{quote}

\section{LoRA Variant}
\label{sec:adapter_capacity}

We also train a much smaller adapter to test how much of the gain depends on
adapter capacity. \method-LoRA keeps the
released context blocks fixed and trains rank-16 LoRA updates of their
attention and feed-forward matrices (15.3\,M parameters) on the same data
and schedule as the full adapter (Table~\ref{tab:training_variants}).
Table~\ref{tab:lora_variant} reports its scores next to the full adapter on
the metrics of Tables~\ref{tab:main_numbers} and~\ref{tab:vbench}; it was
not included in the user study. \method-LoRA exceeds all four baselines on
FID, FVD, and IG-IoU, while the full adapter leads it on every metric, including rotation error ($4.02^\circ$ to $3.65^\circ$) and translation error
(0.131 to 0.122).

\begin{table}[H]
    \centering
    \caption{\textbf{LoRA variant against the full adapter.} The metrics of
    Table~\ref{tab:main_numbers} (left) and the VBench-I2V dimensions of
    Table~\ref{tab:vbench} (right).}
    \label{tab:lora_variant}
    \setlength{\tabcolsep}{3pt}
    \resizebox{\linewidth}{!}{%
    \begin{tabular}{lccccccc|cccccccc}
        \toprule
        & FID & FVD & CLIP & Rot.\ ($^\circ$) & Trans. & Rec. & IG-IoU
        & Subj.\ & Bkg.\ & Motion & Dyn.\ & Aesth.\ & Imag.\ & I2V & I2V \\
        Method & $\downarrow$ & $\downarrow$ & $\uparrow$ & $\downarrow$ & $\downarrow$ & $\uparrow$ & $\uparrow$
        & cons. & cons. & smooth. & degree & quality & quality & subj. & bkg. \\
        \midrule
        \method-LoRA & 44.4 & 393.2 & 31.4 & 4.02 & 0.131 & 94.1 & 57.3
        & 90.0 & 91.5 & 98.7 & 92.0 & 56.8 & 63.2 & 96.2 & 96.9 \\
        \method{} & \textbf{44.1} & \textbf{370.4} & \textbf{31.8} & \textbf{3.65} & \textbf{0.122} & \textbf{94.9} & \textbf{60.4}
        & \textbf{90.3} & \textbf{92.0} & \textbf{98.8} & \textbf{96.0} & \textbf{57.0} & \textbf{64.6} & \textbf{96.3} & \textbf{97.0} \\
        \bottomrule
    \end{tabular}}
\end{table}

\section{Occlusion}
\label{sec:occlusion}

\begin{figure}[tbp]
    \centering
    \includegraphics[width=\linewidth]{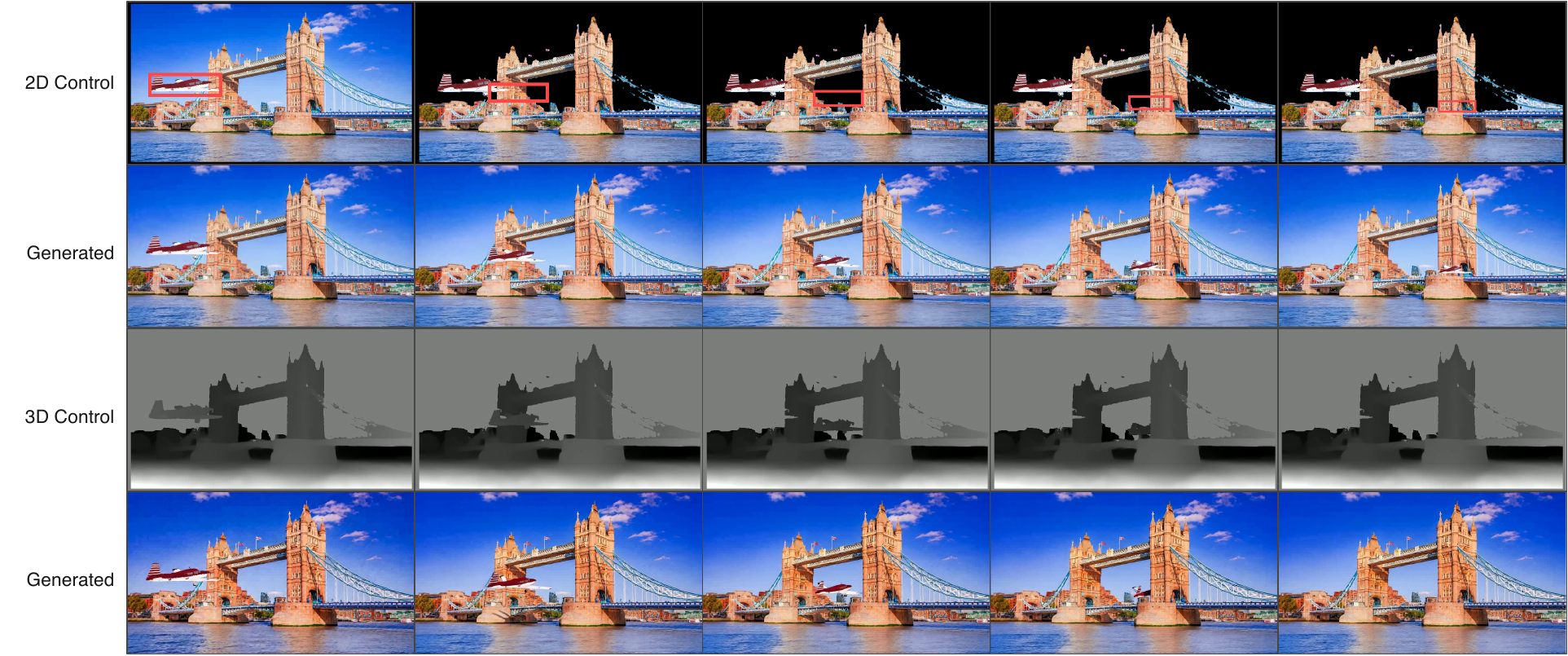}
    \caption{\textbf{Occlusion.} A plane is directed to fly under the bridge
    and behind its right tower. \emph{Rows 1--2:} the control of SymphoMotion, a 2D box
    projected from the 3D trajectory onto the rendered point cloud (red), and the video
    it generates. \emph{Rows 3--4:} the depth control
    rendered from the rigid 3D scene and the video \method{} generates from it. Only with the depth
    control does the tower hide the plane.}
    \label{fig:occlusion}
\end{figure}

A 2D control fixes where an object appears in the image, but not whether it
passes in front of or behind the rest of the scene. In the depth video of a
rigid 3D scene, each pixel takes the depth of the nearest surface
(Sec.~\ref{sec:control_rendering}), so this order is fixed before
generation. Fig.~\ref{fig:occlusion} directs a plane to fly under the bridge
and on behind its right tower. SymphoMotion marks the plane by a 2D box on the rendered point cloud and conditions on its 3D trajectory, but its control gives no per-pixel visibility, and the plane it generates passes in front of the tower; with the rigid 3D geometry, whose depth video states which surface is nearest at every pixel, \method{} flies it under the deck and hides it behind the tower.

\section{Failure Case}
\label{sec:failure}

\begin{figure}[H]
    \centering
    \includegraphics[width=\linewidth]{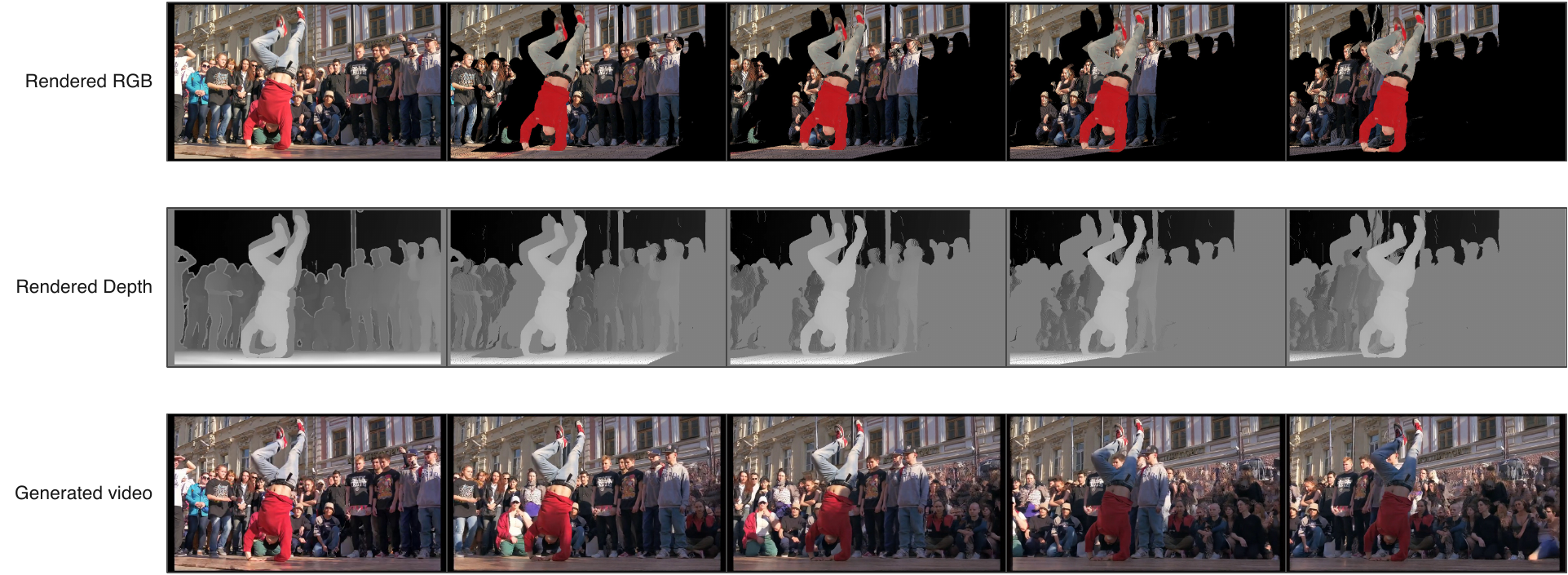}
    \caption{\textbf{Failure case.} A breakdancer in a handstand is moved to
    the right as one rigid body while the camera moves. \emph{Top:} the rigid 3D scene rendered to
    RGB, for visualization only, and to the depth video that conditions the
    adapter; pixels without geometry are black in the RGB render and uniform
    gray in the depth video. \emph{Bottom:} the generated dancer keeps the pose
    of the first frame throughout, following the rigid silhouette, instead of
    dancing.}
    \label{fig:failure}
\end{figure}

One rigid transformation per object cannot express motion that is mostly
articulated. Fig.~\ref{fig:failure} moves a breakdancer in a handstand to the
right as one rigid body while the camera moves. The depth control fixes the silhouette of the whole
body in every frame, and the generated dancer follows it: it keeps the
handstand pose of the first frame throughout instead of dancing. Prescribing
such motion requires articulated parts in the scene
(Sec.~\ref{sec:conclusion}).

\section{Interfaces}
\label{sec:interfaces}

\subsection{Authoring Interface}
\label{sec:authoring_ui}

One image is enough to build a rigid 3D scene, whose camera and object
trajectories are then authored in a 3D viewer; this is how the cases of
Appendix~\ref{sec:authored_controls} were produced. The interface has three
stages.

\begin{figure}[H]
    \centering
    \begin{subfigure}[b]{0.35\linewidth}
        \centering
        \includegraphics[height=4.9cm]{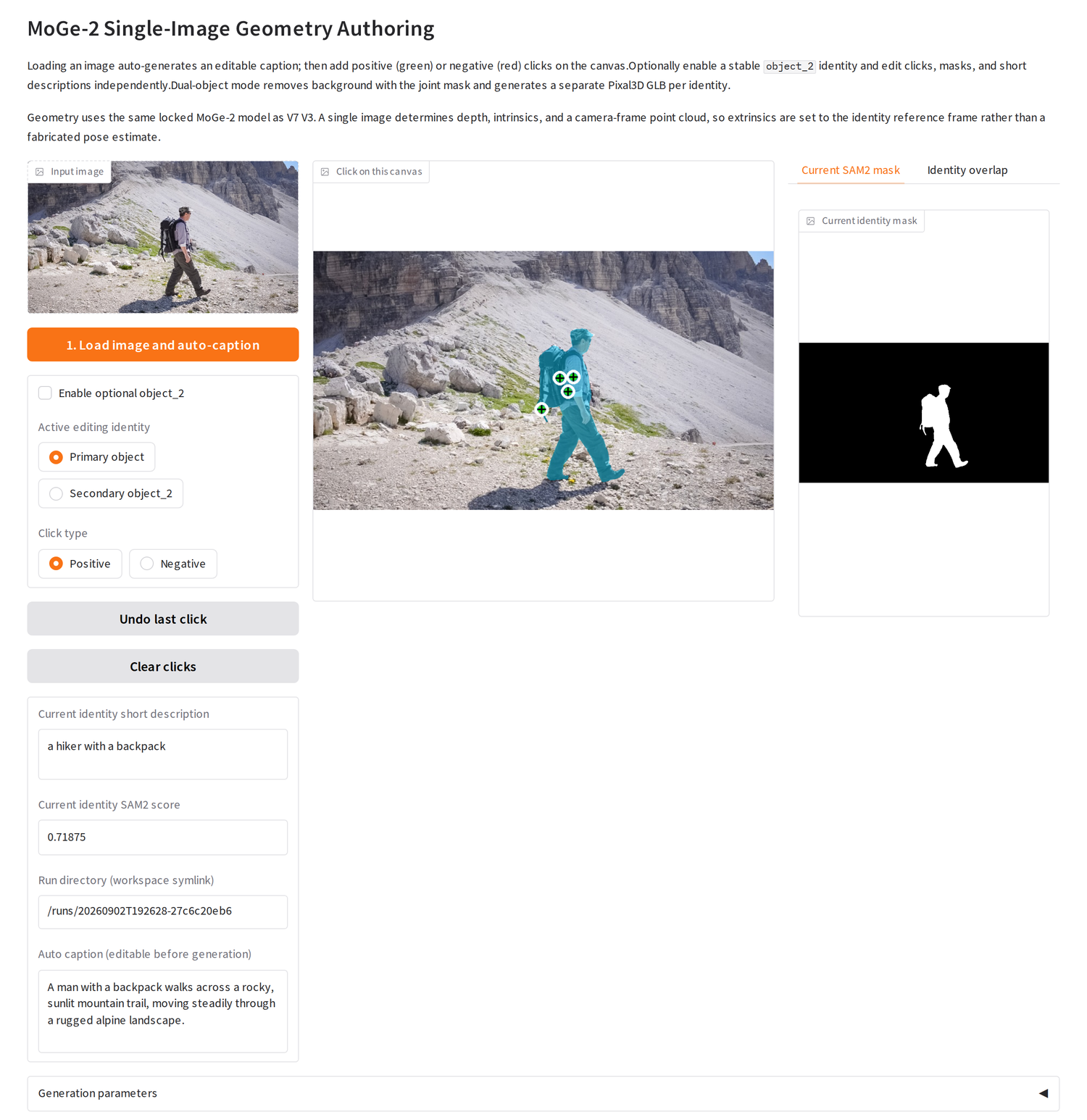}
        \subcaption{Identity}
        \label{fig:ui_identity}
    \end{subfigure}\hfill
    \begin{subfigure}[b]{0.22\linewidth}
        \centering
        \includegraphics[height=4.9cm]{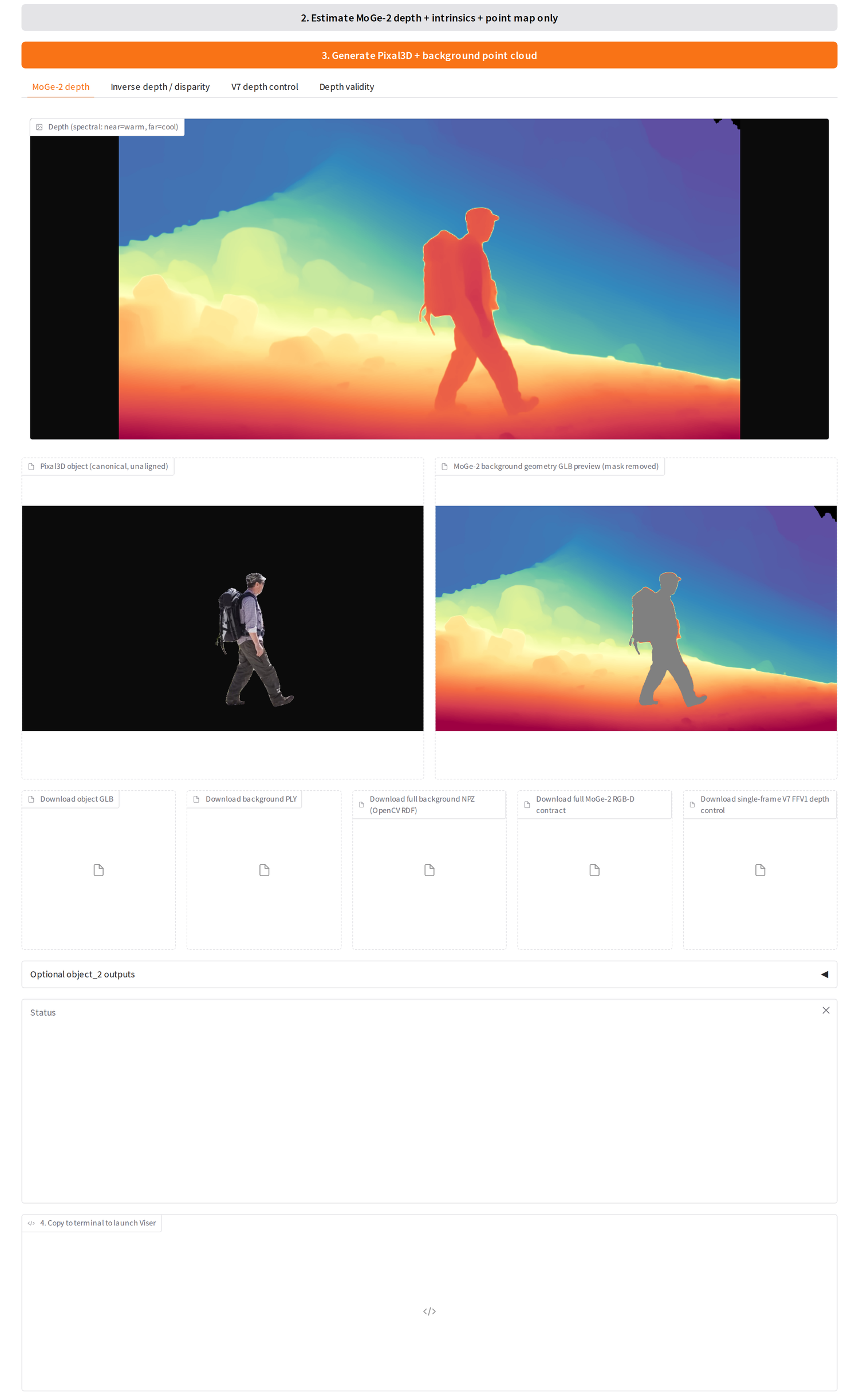}
        \subcaption{Geometry}
        \label{fig:ui_geometry}
    \end{subfigure}\hfill
    \begin{subfigure}[b]{0.38\linewidth}
        \centering
        \includegraphics[height=4.9cm]{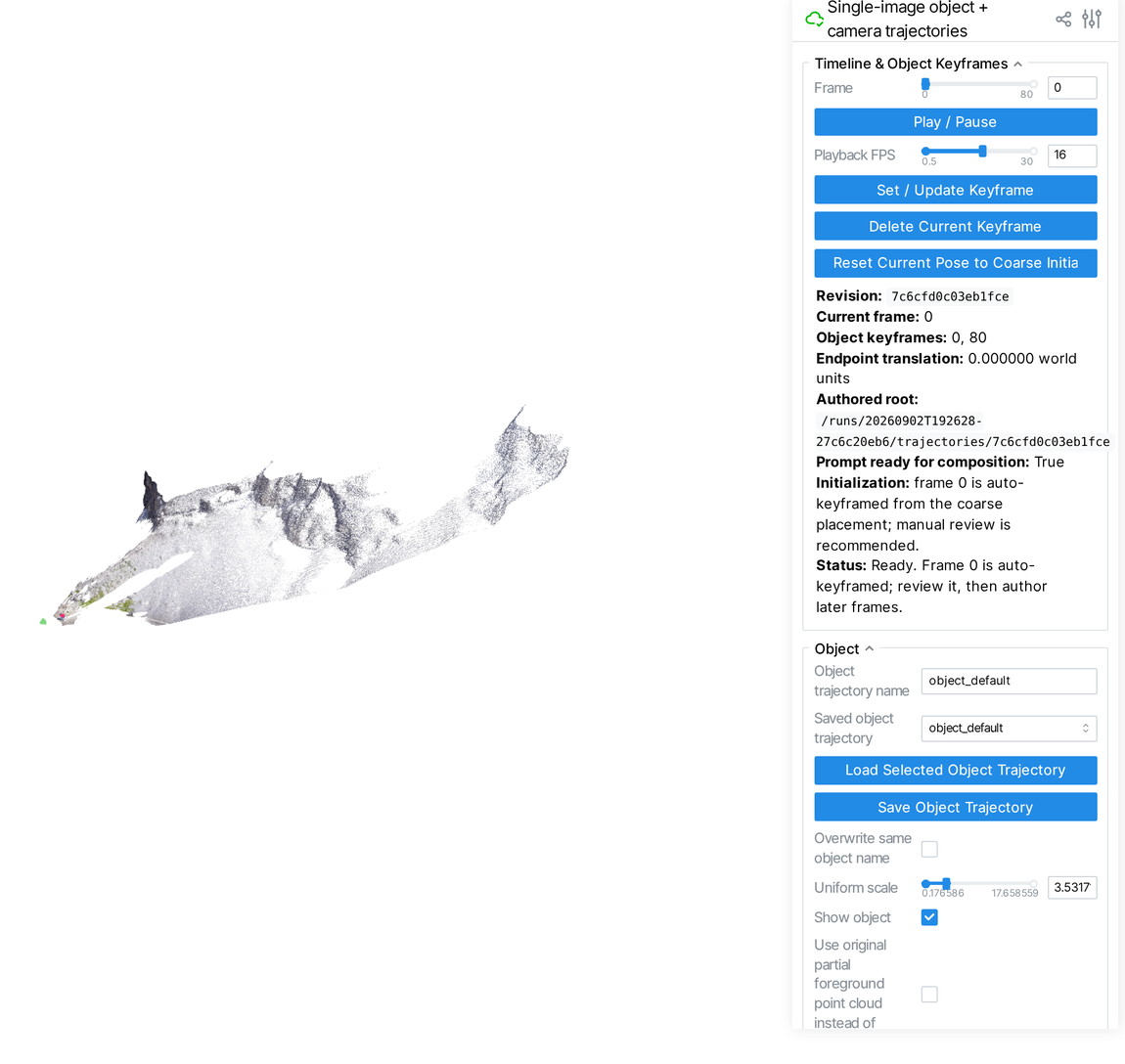}
        \subcaption{Trajectories}
        \label{fig:ui_viser}
    \end{subfigure}
    \caption{\textbf{Authoring interface.} The three stages of building and
    directing a rigid 3D scene from a single image; in (c) the trajectories
    are set by keyframes on a timeline.}
    \label{fig:ui_authoring}
\end{figure}

\noindent\textbf{(1) Identity.} The user loads an image, which is captioned
automatically, and marks each object with positive and negative clicks; the
segmentation and the resulting identity mask are shown beside the canvas
(Fig.~\ref{fig:ui_identity}).

\noindent\textbf{(2) Geometry.} Depth and intrinsics are estimated for the
frame, the canonical mesh of each identity is reconstructed, and the
background is lifted into a point cloud with the object masks removed. The
panel previews the result and exports the scene
(Fig.~\ref{fig:ui_geometry}).

\noindent\textbf{(3) Trajectories.} The scene opens in a viewer where the
camera trajectory and each object's rigid trajectory are set by keyframes on a
timeline (Fig.~\ref{fig:ui_viser}). The authored scene is rendered by the
same renderer as a recovered one (Sec.~\ref{sec:control_rendering}) and the
resulting depth video conditions the adapter.

\subsection{User Study Interface}
\label{sec:user_study_details}

\begin{figure}[H]
    \centering
    \includegraphics[width=\linewidth]{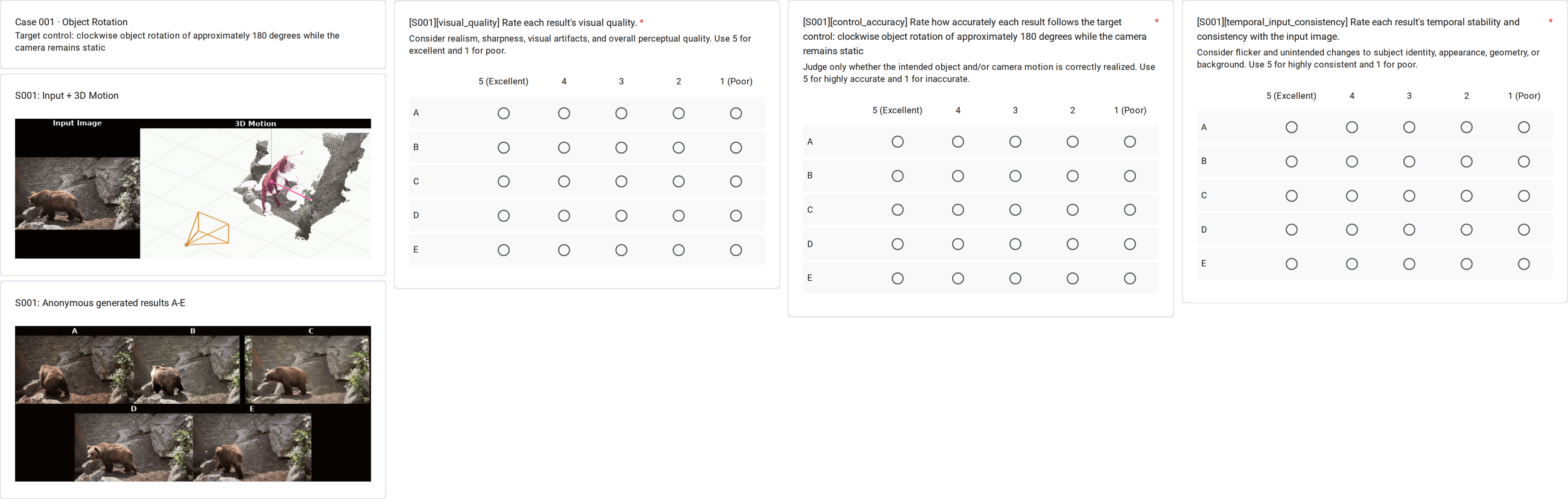}
    \caption{\textbf{One page of the user study.} \emph{Left:} the case and the
    five anonymized results. \emph{Right:} the three rating questions.}
    \label{fig:user_study_form}
\end{figure}

The study of Sec.~\ref{sec:user_study} is a form of nine cases, each on its own
page. A case states the target control in words, shows the input image beside a 3D visualization of the target
motion, and shows the five results as an anonymized grid labeled A--E, in
which our model occupies a different letter in each case. Three matrix
questions then ask for a rating from 5 (excellent) to 1 (poor) for every
result: \emph{visual quality} (realism, sharpness, artifacts, overall
perceptual quality), \emph{control accuracy} (whether the intended object and
camera motion is realized, judged on motion alone), and \emph{temporal and
input consistency} (flicker and unintended changes of subject identity,
appearance, geometry, or background). The nine cases cover object rotation
(five), camera yaw (one), object translation with a following camera
(two), and non-rigid dynamics (one). Fig.~\ref{fig:user_study_form} shows one page.

\section{Authored Trajectories and Insertions}
\label{sec:authored_section}

\label{sec:authored_controls}
\begin{figure}[tp]
    \centering
    \includegraphics[width=\linewidth]{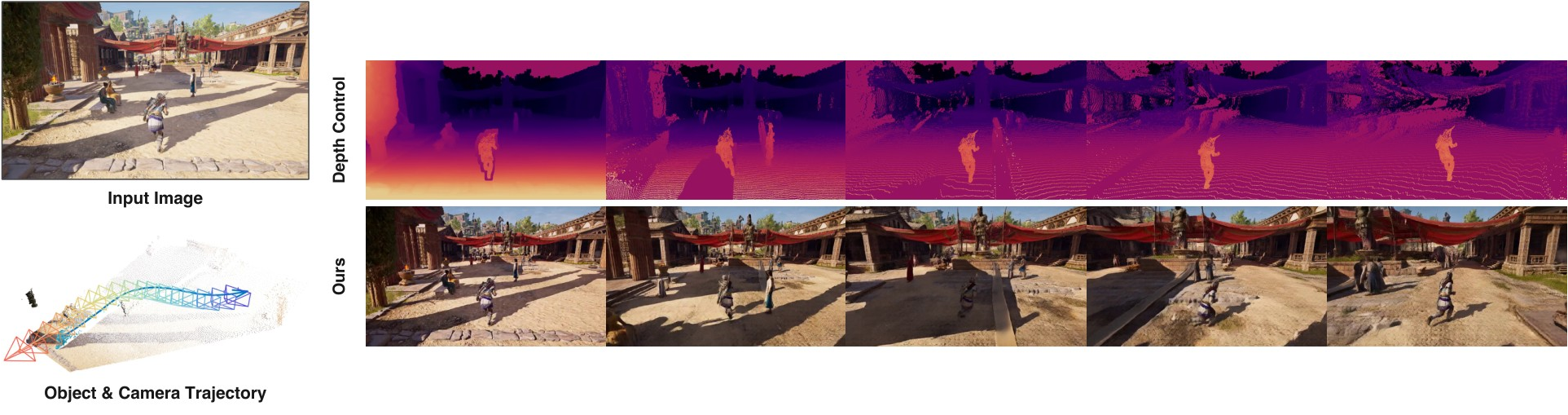}\\[4pt]
    \includegraphics[width=\linewidth]{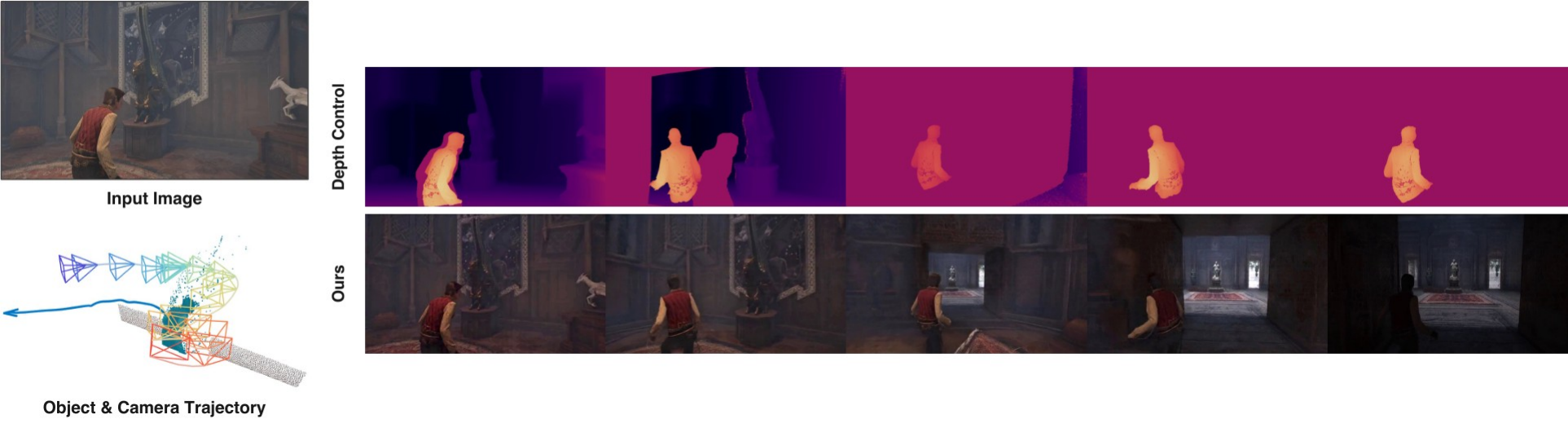}\\[4pt]
    \includegraphics[width=\linewidth]{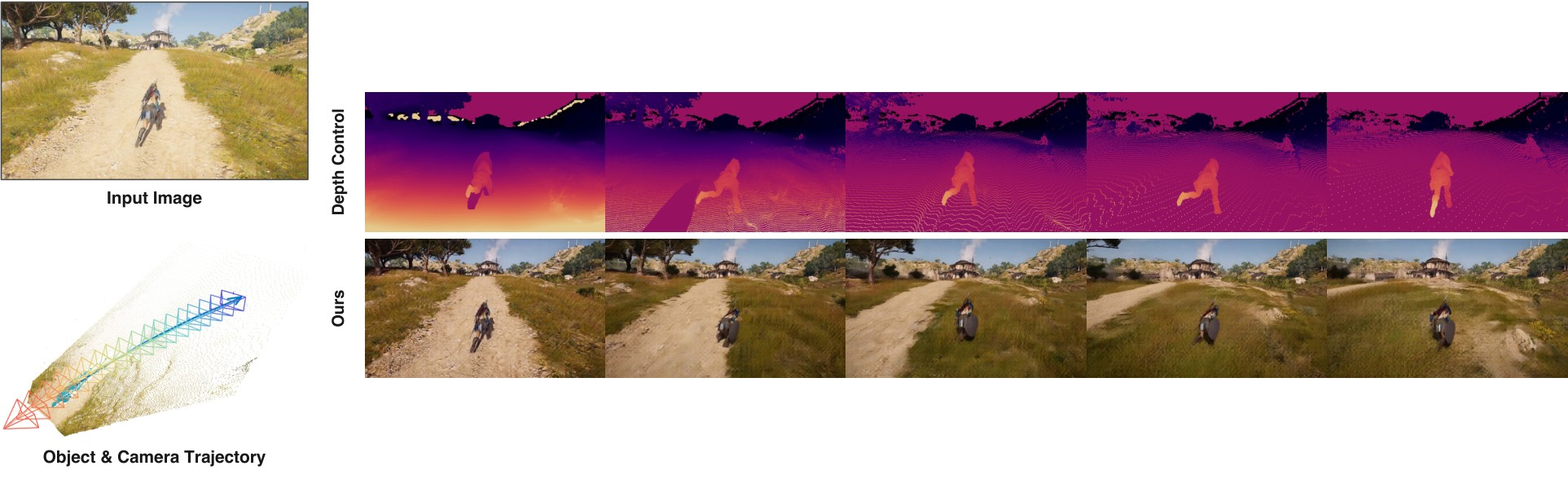}\\[4pt]
    \includegraphics[width=\linewidth]{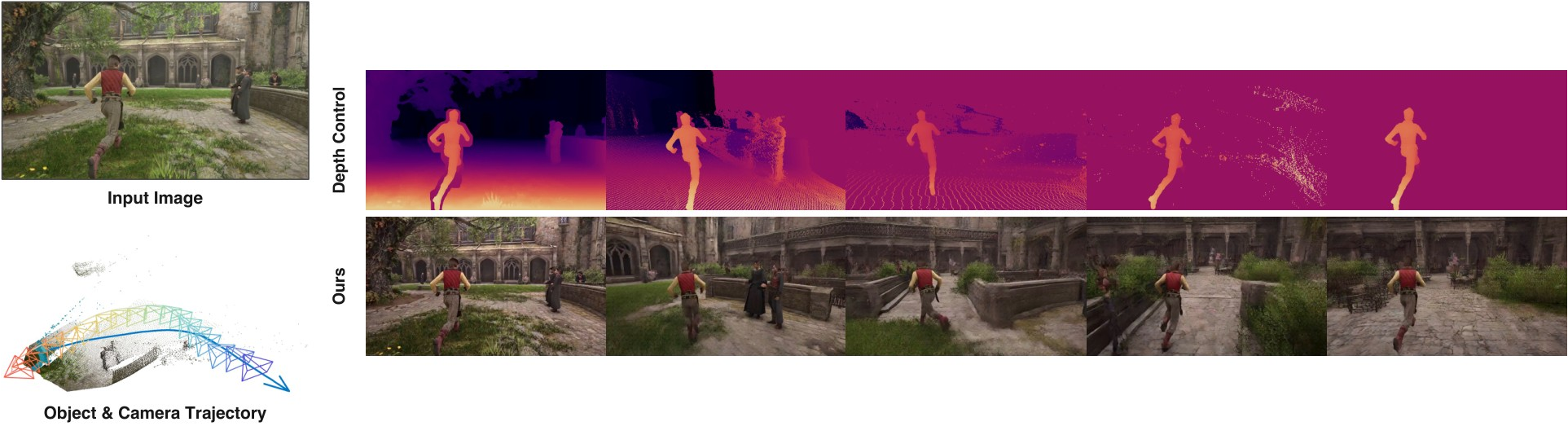}
    \caption{\textbf{Authored camera and object trajectories (1/2).} Each
    panel shows the input image and the authored trajectories on the left, the
    depth control rendered from the rigid 3D scene in the upper row, and the
    generated video below it. None of these controls comes from a source
    video.}
    \label{fig:authored_a}
\end{figure}

\begin{figure}[tp]
    \centering
    \includegraphics[width=\linewidth]{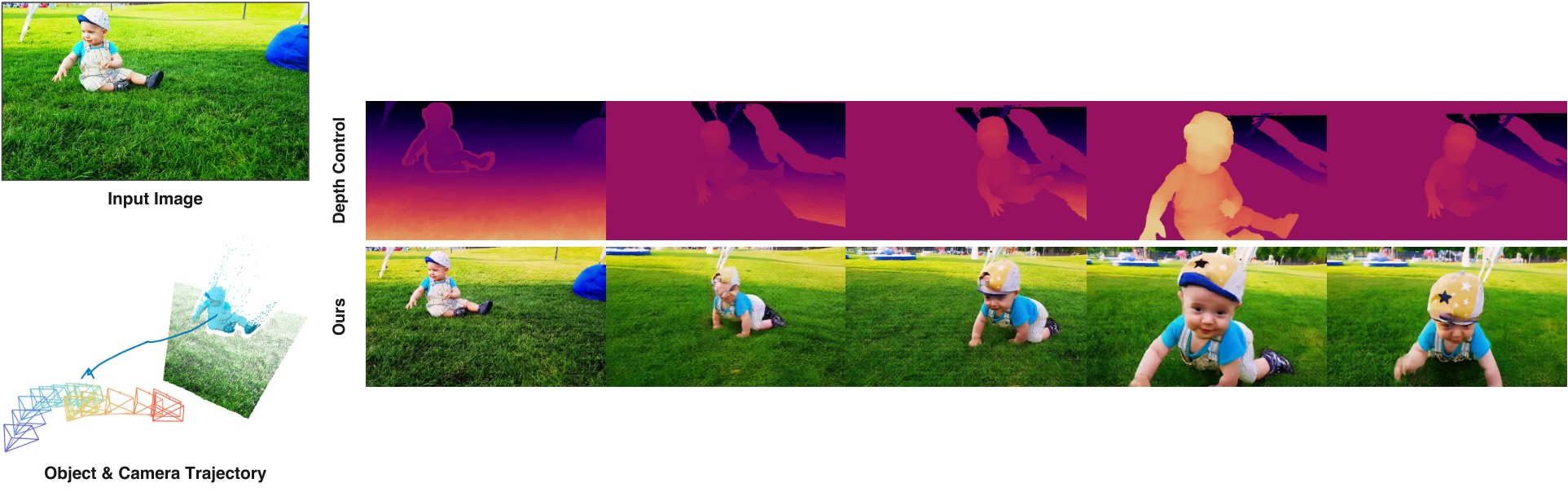}\\[4pt]
    \includegraphics[width=\linewidth]{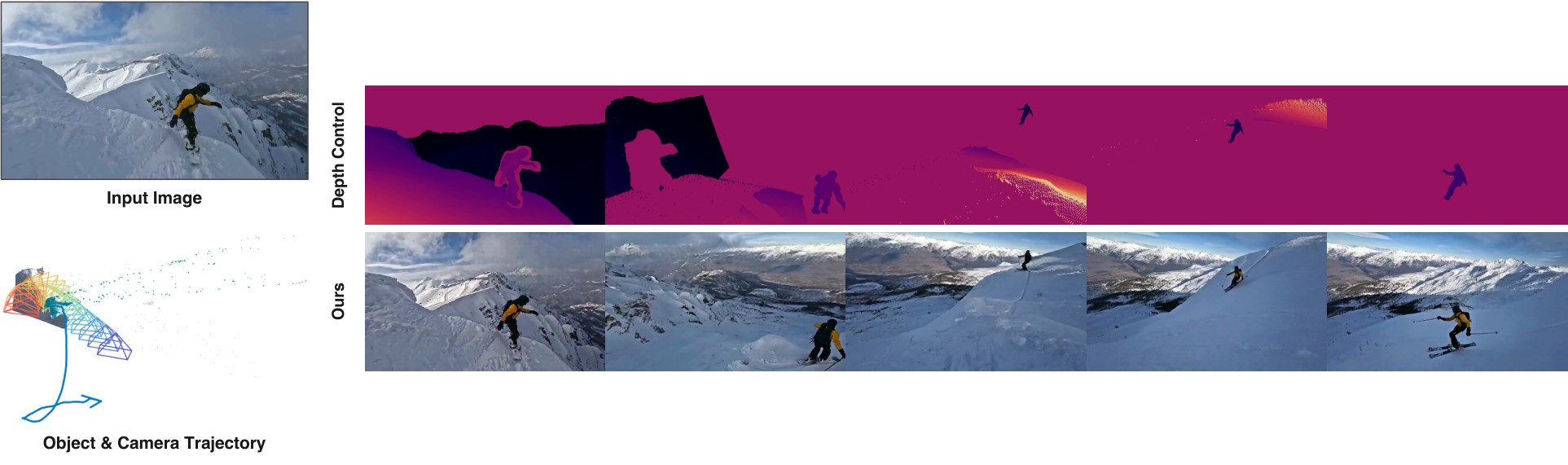}\\[4pt]
    \includegraphics[width=\linewidth]{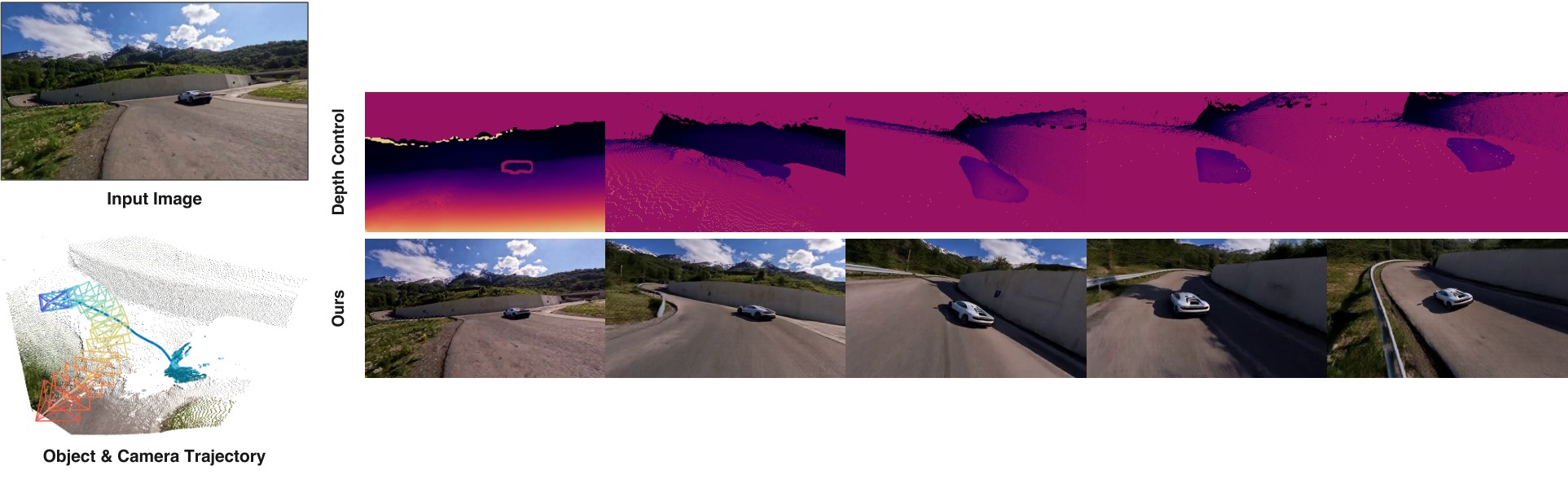}
    \caption{\textbf{Authored camera and object trajectories (2/2).} As in
    Fig.~\ref{fig:authored_a}. The generator supplies the non-rigid dynamics,
    illumination and background that the rigid control leaves unspecified.}
    \label{fig:authored_b}
\end{figure}

The 61 authored cases are public clips under re-authored camera and rigid
trajectories, single-image object insertions and user-supplied cases
(Fig.~\ref{fig:teaser}). They have no source video under the authored trajectories to compare
against, and the training data contain no insertion pairs, so we report them
qualitatively; Figs.~\ref{fig:authored_a} and~\ref{fig:authored_b} show
seven.

\section{Additional Qualitative Results}
\label{sec:more_qualitative}

\begin{figure}[p]
    \centering
    \includegraphics[width=\linewidth]{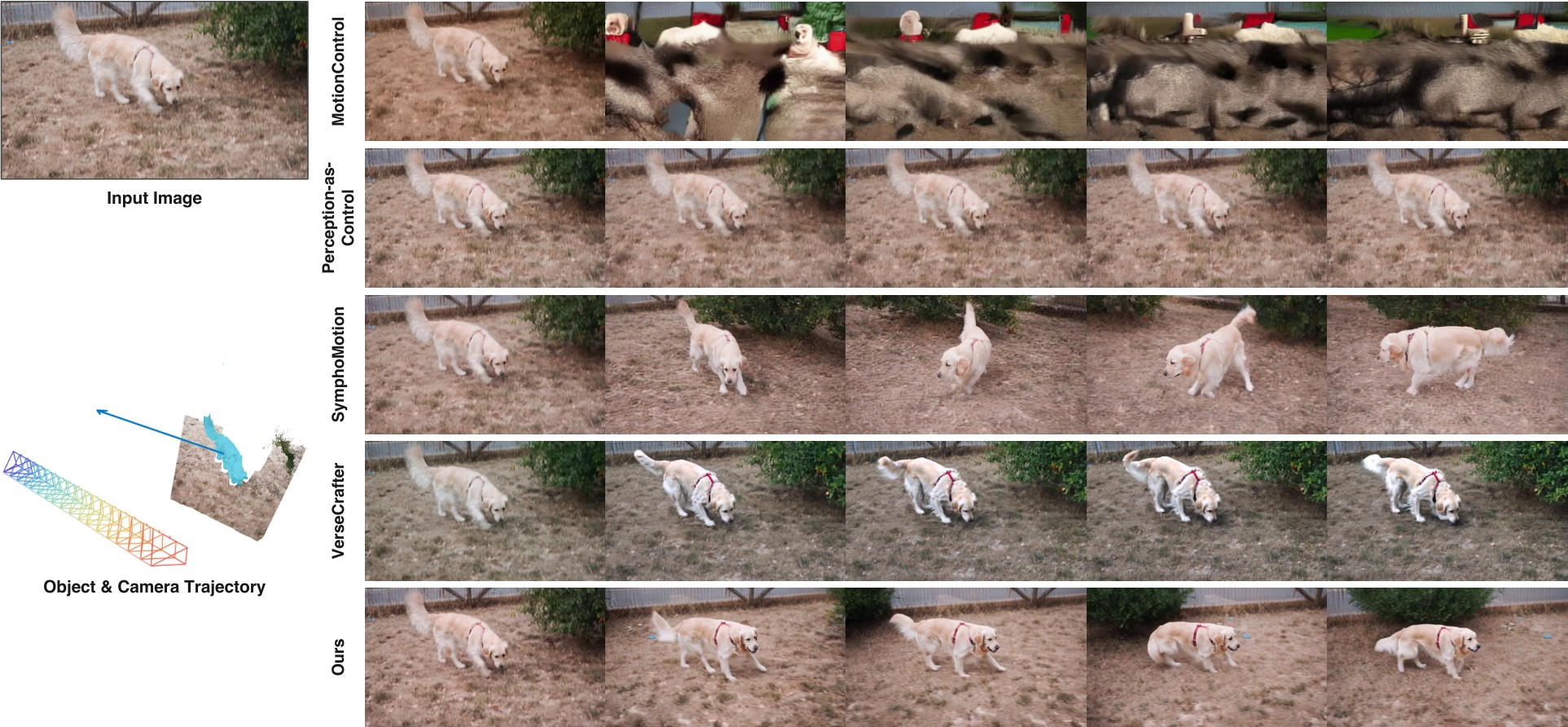}\\[3pt]
    \includegraphics[width=\linewidth]{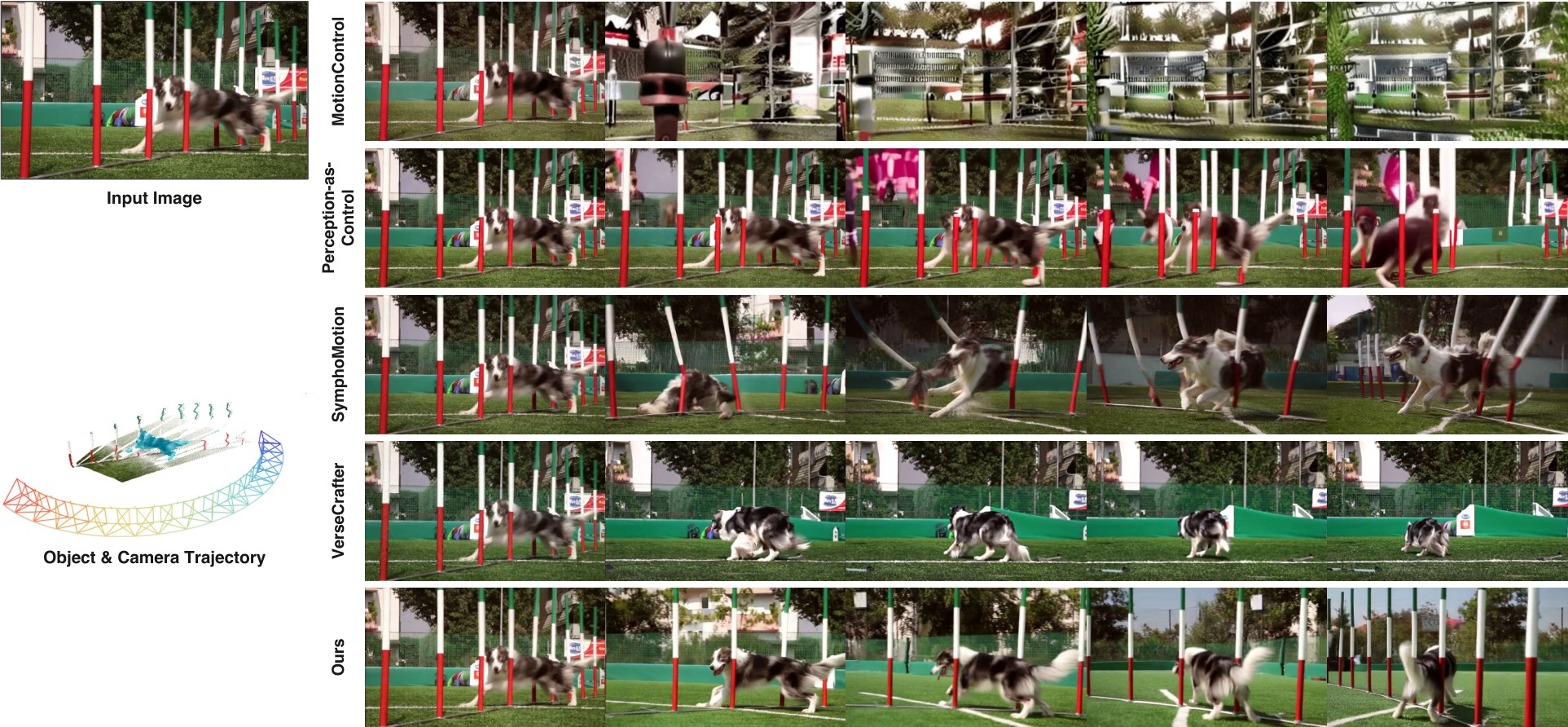}\\[3pt]
    \includegraphics[width=\linewidth]{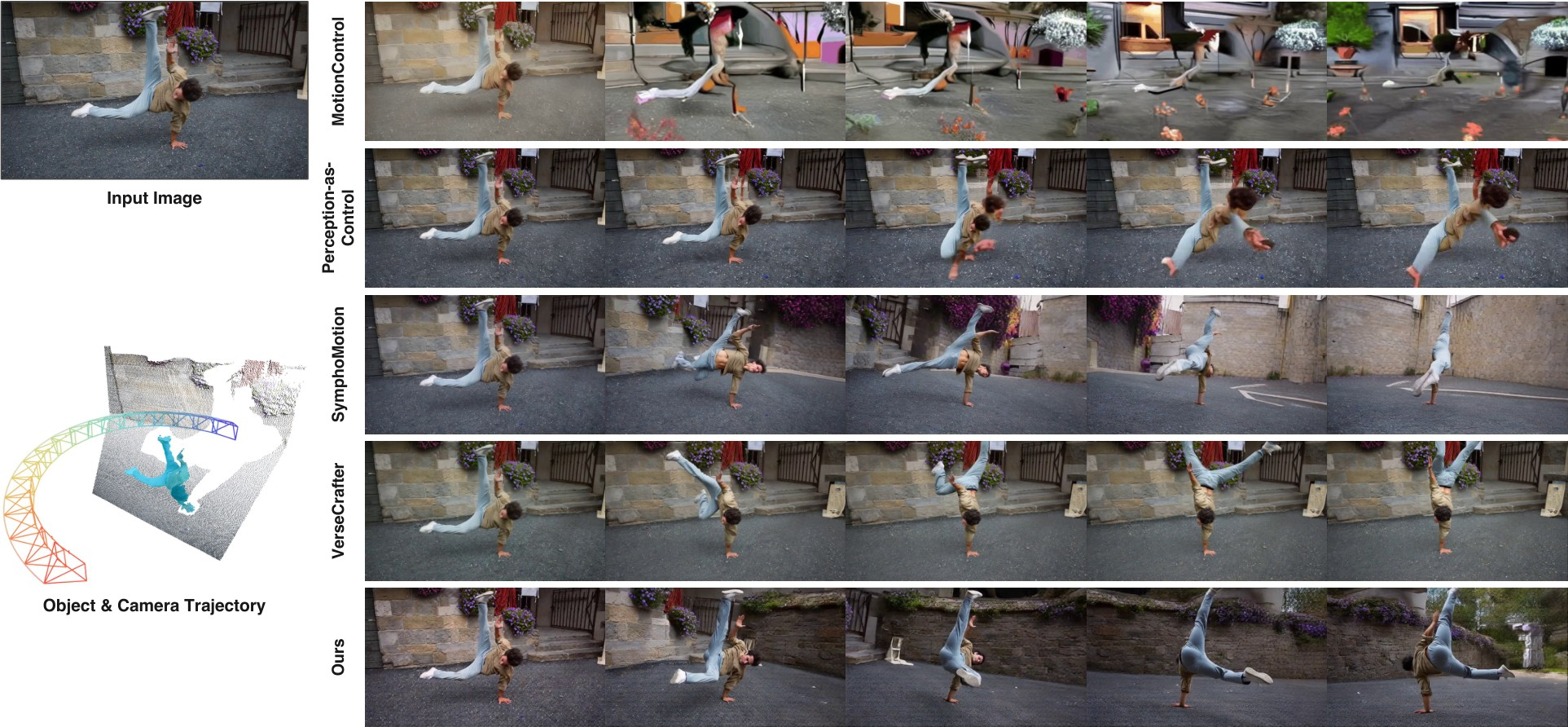}
    \caption{\textbf{Camera and object motion.} In the first two clips, the
    camera follows a trajectory (the coloured frusta) while the object is
    moved by its own rigid trajectory. In the third, the camera follows an arc
    around the subject and the object keeps its place, so the non-rigid dynamics come from the generator alone.}
    \label{fig:more_joint}
\end{figure}

\begin{figure}[p]
    \centering
    \includegraphics[width=\linewidth]{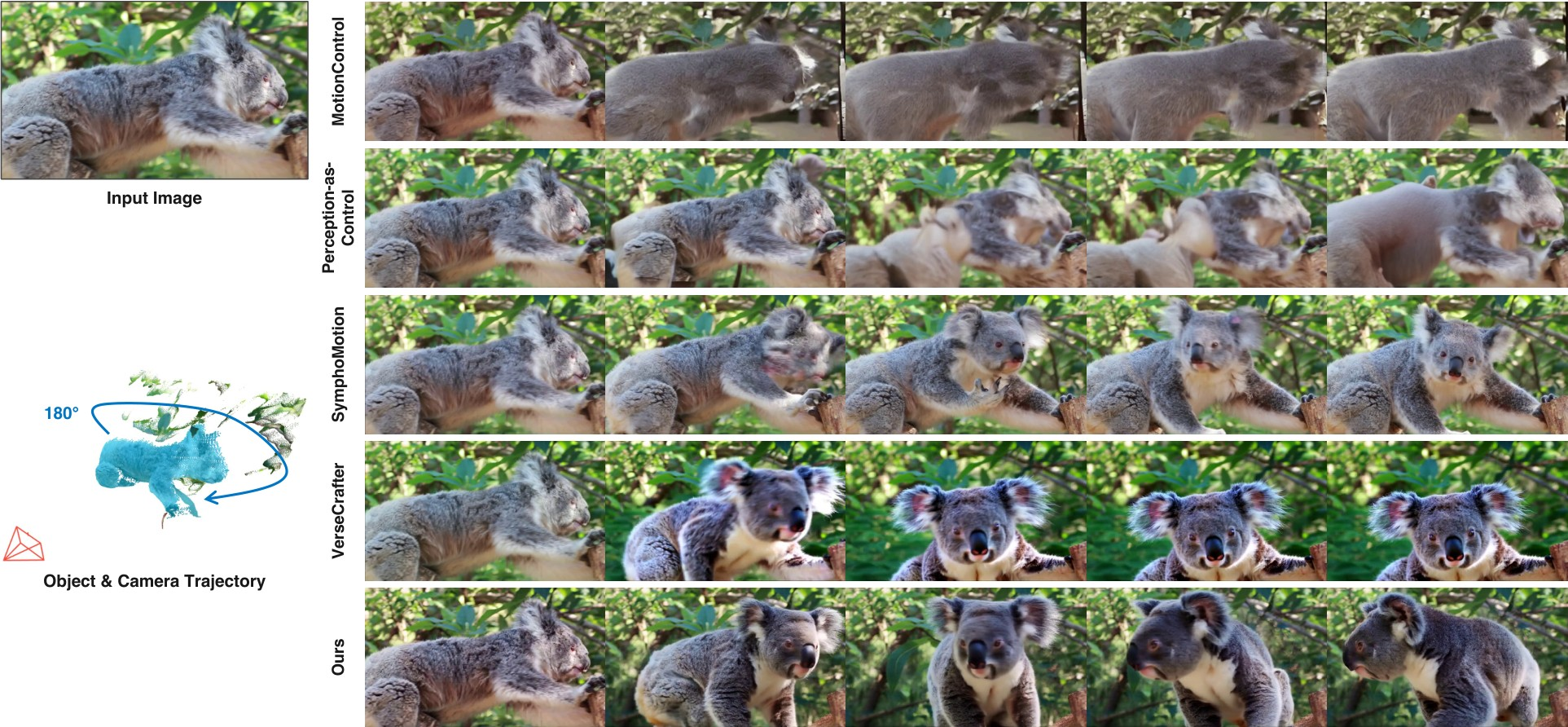}\\[3pt]
    \includegraphics[width=\linewidth]{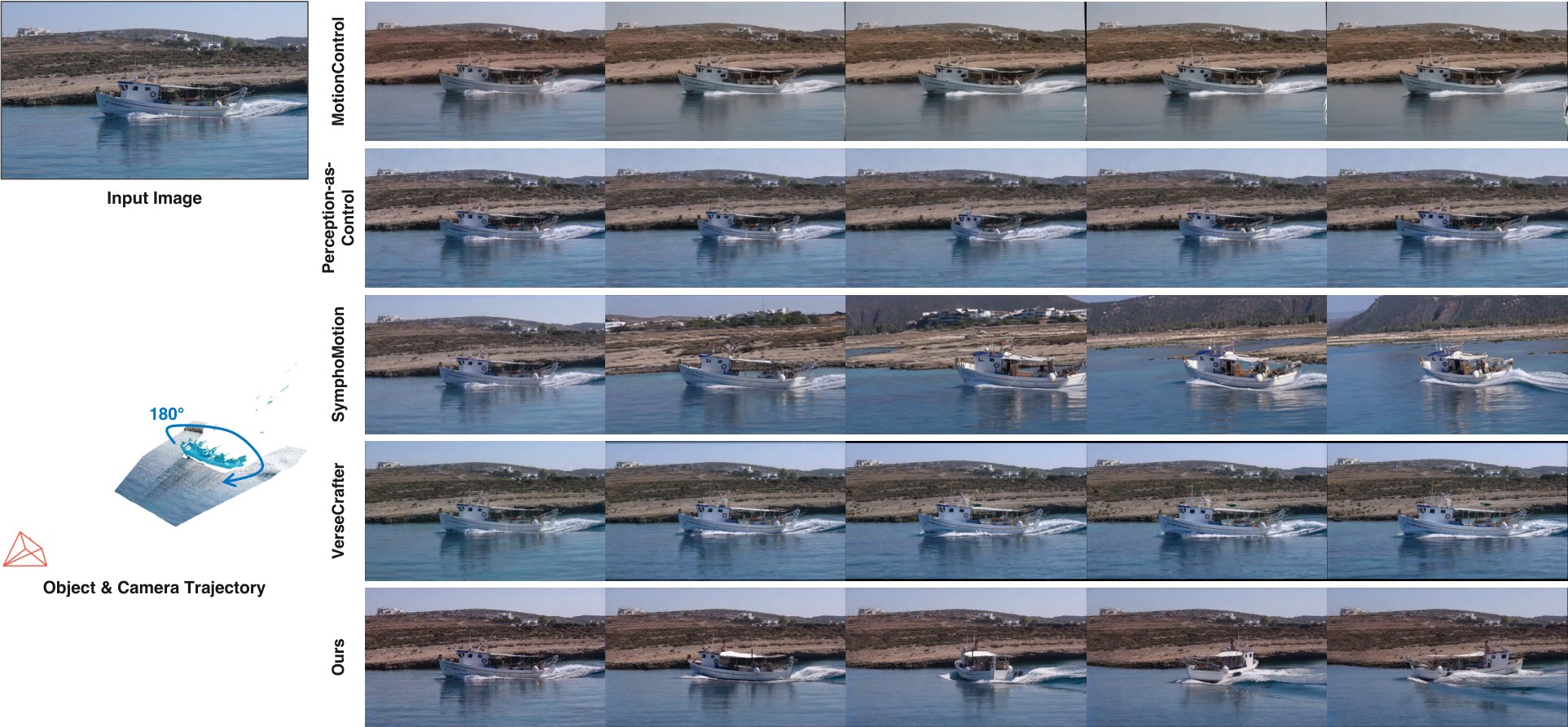}\\[3pt]
    \includegraphics[width=\linewidth]{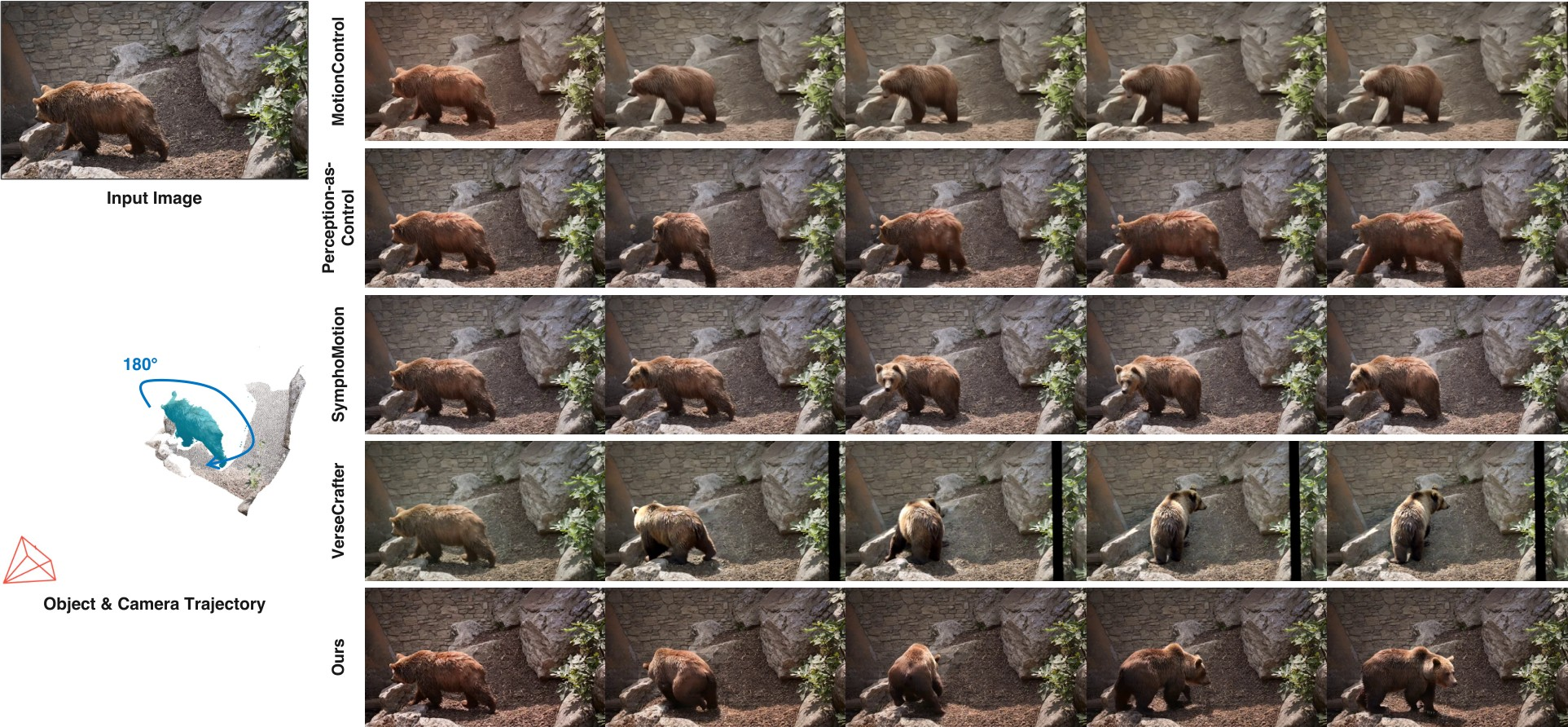}
    \caption{\textbf{Object rotation with a static camera.} The object is
    turned by $180^\circ$ in place, with the camera held at the single frustum
    on the left. The baselines keep the original heading or lose the subject,
    whereas \method{} completes the turn and shows the side that turns into view.}
    \label{fig:more_rotation}
\end{figure}

Figs.~\ref{fig:more_joint} and~\ref{fig:more_rotation} compare \method{} with
the four baselines on six further clips, grouped by the control they
receive. In every panel, the input image and the target trajectories are on the
left; the rows are MotionCtrl (labeled MotionControl),
Perception-as-Control, SymphoMotion, VerseCrafter and \method{} (Ours).

\section{Future Work Direction}
\label{sec:more_future_work_direction}

A promising direction is to enable continual adaptation of video world models as new domains, objects, or control distributions arrive, while preserving previously acquired capabilities. Prior work on continual learning has explored online hyperparameter adaptation and external-data-assisted knowledge preservation, which may provide useful principles for this setting~\citep{liu2021adaptive,liu2021rmm,liu2023online,liu2023continual,liu2024wakening}.

\end{document}